\documentclass{article} % For LaTeX2e
\usepackage{iclr2027_conference,times}

\usepackage{amsmath,amsfonts,bm}

\def\eqref#1{\textup{(\ref{#1})}}

\def\1{\bm{1}}

\definecolor{better}{HTML}{2E7D32}   % dark green

\definecolor{worse}{HTML}{C62828}    % dark red

\definecolor{basegray}{HTML}{F5F5F5} % light gray for baseline row

\DeclareMathAlphabet{\mathsfit}{\encodingdefault}{\sfdefault}{m}{sl}

\SetMathAlphabet{\mathsfit}{bold}{\encodingdefault}{\sfdefault}{bx}{n}

\DeclareMathOperator*{\argmin}{arg\,min}

\usepackage{wrapfig}
\usepackage{hyperref}
\usepackage{url}
\usepackage{graphicx}
\usepackage{caption}
\DeclareCaptionLabelFormat{continued}{#1~#2~(continued)}
\usepackage{booktabs}
\usepackage{amsmath,amssymb}
\usepackage{multirow}
\usepackage{xcolor}
\usepackage{colortbl}
\usepackage{enumitem}
\definecolor{appAccent}{HTML}{245C66}
\definecolor{appHeader}{HTML}{FFFFFF}
\definecolor{appStripe}{HTML}{FFFFFF}
\definecolor{appHighlight}{HTML}{E8F3FF}
\definecolor{appCodeBg}{HTML}{F7FAFA}
\definecolor{appMuted}{HTML}{65777C}
\definecolor{appSecondary}{HTML}{476C91}
\newcommand{\appHeadFont}{\color{black}\bfseries\boldmath}
\newcommand{\appTableStyle}{%
  \arrayrulecolor{black!70}%
  \setlength{\heavyrulewidth}{0.7pt}%
  \setlength{\lightrulewidth}{0.35pt}%
  \setlength{\cmidrulewidth}{0.3pt}%
  \renewcommand{\arraystretch}{1.10}%
}
\newcommand{\appEmph}[1]{\cellcolor{appHighlight}\bfseries\boldmath #1}
\usepackage[most]{tcolorbox}
\usepackage{listings}
\lstdefinelanguage{LeanFour}{
  morekeywords={import,noncomputable,section,open,namespace,end,structure,where,
  def,lemma,theorem,by,have},
  morekeywords=[2]{unfold,positivity,field_simp,ring,nlinarith,constructor},
  sensitive=true,
  morecomment=[l]{--},
  morecomment=[s]{/-}{-/},
  morestring=[b]",
}
\newtcblisting{leancertificate}{
  enhanced, breakable, listing only, listing engine=listings,
  colback=appCodeBg, colframe=appAccent,
  colbacktitle=appHeader, coltitle=appAccent,
  title={Lean 4 \enspace\textnormal{---\enspace VPE certificate}},
  title after break={Lean 4 \enspace\textnormal{---\enspace VPE certificate (continued)}},
  fonttitle=\small\bfseries,
  boxrule=0.4pt, arc=1mm,
  left=5pt, right=5pt, top=5pt, bottom=5pt,
  before skip=10pt, after skip=10pt,
  listing options={
    language=LeanFour,
    basicstyle=\ttfamily\fontsize{8.5}{10}\selectfont,
    keywordstyle=\color{appAccent}\bfseries,
    keywordstyle={[2]\color{appSecondary}},
    commentstyle=\color{appMuted}\itshape,
    stringstyle=\color{appSecondary},
    numbers=left, numberstyle=\tiny\color{appMuted},
    numbersep=8pt, xleftmargin=2em,
    columns=fullflexible, keepspaces=true,
    showstringspaces=false,
    breaklines=true, breakatwhitespace=true,
    tabsize=2, aboveskip=0pt, belowskip=0pt,
    literate={·}{{\ensuremath{\cdot}}}1
    {≠}{{\ensuremath{\ne}}}1,
  },
}
\newtcolorbox{finding}{
  colback=teal!5, colframe=teal!60!black,
  arc=2mm, boxrule=0.6pt,
  left=8pt, right=8pt, top=4pt, bottom=4pt,
}
\usepackage{titletoc}
\usepackage[makeroom]{cancel}
\usepackage{makecell}
\usepackage{tabularx}
\usepackage{array}
\usepackage{amsthm}

\title{Not All Error Yields to Scale: Where Scaling Stops in
Vision-Language Inference}

\author{
Xinye Zhao\thanks{Equal contribution.} \quad
Yunkai Dang\footnotemark[1] \quad
Yunchen Wu \quad
Wenbin Li\thanks{Corresponding author: \texttt{liwenbin@nju.edu.cn}, \texttt{xinye\_zhao@smail.nju.edu.cn}.} \\
\normalfont School of Intelligence Science and Technology, Nanjing University
}

\newcommand{\einf}{e_\infty}
\newcommand{\nlin}{N}
\newcommand{\rlin}{R}

\iclrfinalcopy
\begin{document}
\raggedbottom
\maketitle
\fancyhead[L]{Preprint}
\begin{abstract}
  Vision-language models (VLMs) face a fixed-budget trade-off between processing more visual information for fine-grained perception and using a larger language backbone for complex reasoning.
  Existing studies do not tell us which combination of backbone size and input resolution to deploy, especially in high-resolution deployments.
  To address this gap, we propose the Separable Law that describes how VLM performance changes with language backbone size and visual token count.
  We fit the law to measurements from 26 InternVL and QwenVL models, with language backbone sizes from 1B to 72B, on four high-resolution benchmarks with image sizes from 224 pixels to 8K.
  We find that the questions responding to scaling can be predicted from the skill they require, while a substantial fraction never responds at all.
  We also find that the two model families gain similarly from a larger backbone, while their gains from more visual tokens differ sharply.
  Combined with a cost law, the Separable Law gives a closed-form rule for allocating compute between backbone size and visual tokens.
  When deployment is limited to available configurations, the law identifies model and image sizes that perform close to the best feasible choice under the same budget.
  We hope our work offers a principled way to decide how much a model should be allowed to see at high resolution, given what it must reason about.
\end{abstract}

\section{Introduction}
\label{sec:intro}
\begingroup
\setlength{\intextsep}{2pt}
\begin{wrapfigure}{r}{0.36\textwidth}
  \centering
  \captionsetup{font=footnotesize,skip=3pt,justification=raggedright,singlelinecheck=false}
  \includegraphics[width=\linewidth]
  {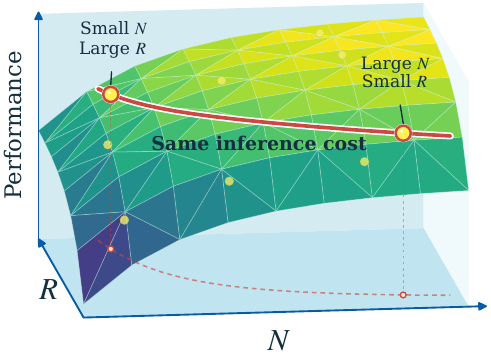}
  \vspace{-1.2em}
  \caption{{{Fixed-budget VLM inference.}}}
  \vspace{-0.2em}
  \label{fig:intro_scaling}
\end{wrapfigure}

Vision-language models (VLMs) have made substantial progress on tasks that require visual understanding and reasoning~\citep{yin2024survey,jin2025efficient,deng2026openvlthinker}.
Many real-world applications require these models to identify fine-grained details in high-resolution images containing large amounts of visual information~\citep{wu2024v,fu2024blink,zhang2025hrscene}.
To preserve these details, VLMs commonly process high-resolution inputs through {dynamic resizing}~\citep{wang2024qwen2,kimhi2026cares,liu2026one} or {image tiling}~\citep{chen2024far,liu2025nvila,luo2025large}, and the two strategies produce different visual-token counts as image resolution increases.
At the same time, complex reasoning tasks such as cross-region comparison~\citep{wang2025divide,li2026dicobench} and multi-step deduction~\citep{wang2025traceable,he2026vistahop} require greater language backbone capacity.
Meeting these perception and reasoning demands makes the VLM inference cost depend on the visual token count $\rlin$ and the backbone size $\nlin$, so a fixed inference budget requires balancing compute between the two.
This raises an important question, illustrated in Figure~\ref{fig:intro_scaling}: \textbf{Given a fixed inference budget, how should compute be allocated between the language backbone and the visual tokens?}

\par\endgroup

\begin{figure}[!t]
  \centering
  \includegraphics[width=0.9\linewidth]{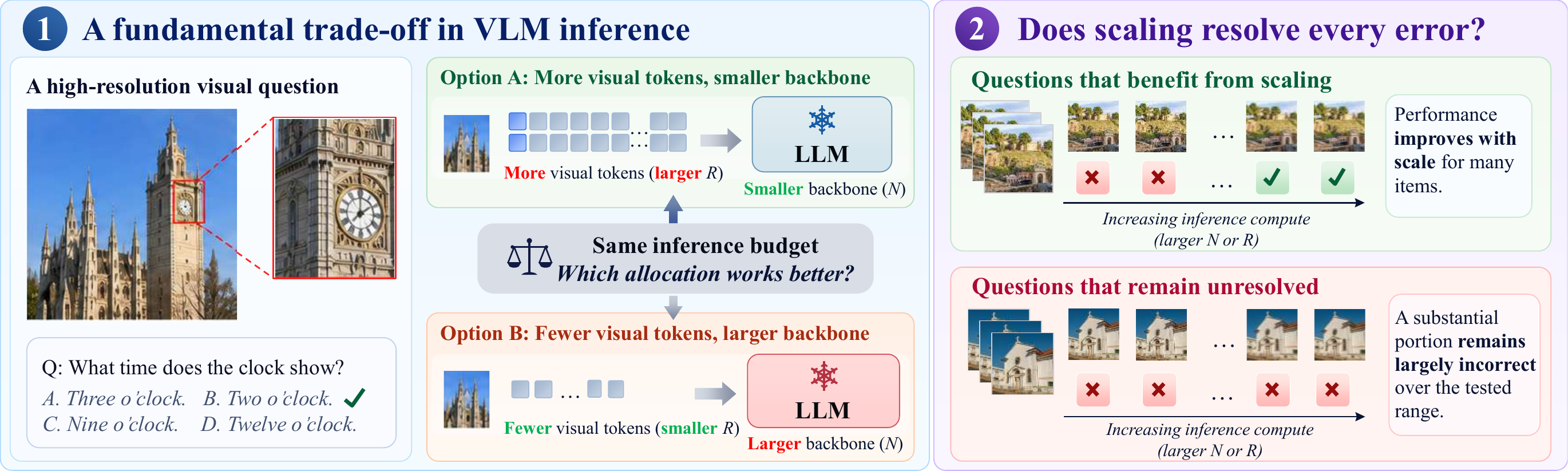}
  %\vspace{-0.95em}
  \caption{
    \textbf{Motivation.}
    \textbf{Left:} Language backbone size ($\nlin$) and visual tokens ($\rlin$) both consume inference compute, which raises the question of how to split a fixed budget between them.
    \textbf{Right:} Scaling helps many questions but leaves others wrong at every scale we test.
  }
  %\vspace{-1em}
  \label{fig:motivation}
\end{figure}

Existing studies have improved VLM inference by reducing the cost of individual components.
Token pruning and merging reduce the number of visual tokens~\citep{chen2024image,shang2025llava,zhang2024sparsevlm}, while efficient vision encoders~\citep{vasu2025fastvlm} and lightweight projectors~\citep{cha2024honeybee} reduce visual processing cost.
But these efficiency gains do not determine how a fixed inference budget should be divided between a larger backbone and more visual tokens.
Recent work models performance jointly over backbone size and visual-token count, but only at a low image resolution of $336$ pixels~\citep{li2024inference}.
Other work jointly adapts visual-token retention and active LLM computation within a given backbone~\citep{wang2026look}.
A complementary line of work evaluates test-time scaling strategies, including chain-of-thought prompting and self-consistency, across VLMs and tasks~\citep{sammani2026test}.
However, these studies remain confined to low-resolution inputs and do not characterize scaling behavior under high-resolution inputs, where how a model turns an image into visual tokens has more room to change the token count.
They also leave open how different image-processing strategies affect scaling behavior.

% However, these methods do not characterize how performance varies jointly with $\nlin$ and $\rlin$, so they cannot determine whether a fixed budget is better spent on more visual tokens with a smaller backbone or on fewer visual tokens with a larger backbone (Figure~\ref{fig:motivation}, left).
% Although the Chinchilla scaling law~\citep{hoffmann2022training} guides how training compute is split between parameters and data for language models, it does not address how inference compute should be allocated in VLMs.
% Broader multimodal scaling studies vary model size, training data, and test-time reasoning strategies but do not model visual token count as an independent variable~\citep{chen2024expanding,li2024inference}.
% Compute-optimal analysis of video VLMs also varies language model size, input frames, and visual tokens per frame~\citep{wang2025inference}.
% However, existing work does not examine how image resolution affects token counts across models or which questions remain incorrect at larger scales.
To address this gap, we propose the Separable Law that describes how VLM performance changes with language backbone size and visual token count.
Its form is $L(\nlin,\rlin)=e_{\infty}+A/\nlin^{\alpha}+B/\rlin^{\beta}$, where $\nlin$ is the language backbone size, $\rlin$ is the visual token count, and $L$ is the fraction of questions answered incorrectly.
The term $e_{\infty}$ is the error floor that remains even as both the language backbone size and the visual token count grow.
Each of the two reducible-error terms depends on only one variable: changing $\nlin$ with $\rlin$ held fixed affects only $A/\nlin^{\alpha}$, and changing $\rlin$ with $\nlin$ held fixed affects only $B/\rlin^{\beta}$, so the two components can be estimated separately.
To test the law, we evaluate $26$ models from the InternVL and QwenVL families, with model sizes from $1$B to $72$B, on four high-resolution benchmarks covering input resolutions from $224$ pixels to 8K.
For each combination of model series and benchmark, we fit the law separately and find that it fits the observed errors better than the alternative formulations we evaluate, so the two components can be estimated separately, one for the language backbone and the other for the visual tokens.

With the Separable Law established, we examine its fitted parameters to characterize how VLM performance varies with $\nlin$ and $\rlin$.
\textbf{(i) We begin with the fitted error floor $e_{\infty}$}, and find that some questions remain incorrectly answered across the evaluated range, even as the backbone grows larger and the visual-token count increases (Figure~\ref{fig:motivation}, right).
Further analysis shows that reasoning tasks have higher fitted error floors than perception tasks, though the gap comes from a minority of task subsets whose error never falls much.
Decomposing these floors more finely, we see that the required skill explains more variation than the visual domain does, with the most severe cases at particular pairings of skill and scene.
We then turn to \textbf{(ii) the two scaling exponents, $\alpha$ for how fast error falls with backbone capacity and $\beta$ for how fast it falls with visual tokens}.
We find that the two model families agree on capacity ($\alpha$) but differ sharply on visual tokens ($\beta$).
This difference arises because, as input image size increases, the visual-token count changes little for InternVL but grows substantially for QwenVL.
We also find that the typical ordering of the two exponents reverses between the model families, with $\alpha>\beta$ for QwenVL and $\beta>\alpha$ for InternVL.
Thus, a compute-allocation strategy optimized for one architecture may not transfer to another.

To put the Separable Law into practice, we turn it into a framework for allocating compute between the language backbone and the image size under a limited budget.
First, we ask how a fixed inference budget should be spent when input resolutions reach 8K. Combining the law with a cost law gives a closed-form answer: larger backbones for InternVL and more visual tokens for QwenVL.
Second, we ask which configuration to choose from those available in deployment. Selection by predicted error identifies the best one on average, even when the model and image sizes are held out from the fit.
Third, we ask whether a benchmark is difficult in a way that additional scale can resolve. Question-level results show how well a benchmark distinguishes models and whether its questions are more sensitive to backbone size or input image size.
We hope this turns visual scaling from something measured case by case into something that can be reasoned about and predicted.
% We hope our work offers a general way to reason about visual scaling and to predict how a VLM will respond to scale before training or evaluating it.

\section{Method}
\label{sec:setup}

\subsection{Preliminaries}
\label{sec:prelim}

\textbf{Scaling laws for language models.}
Scaling laws~\citep{kaplan2020scaling} describe how model performance changes with controllable resources.
For language model pretraining, the Chinchilla formulation~\citep{hoffmann2022training} provides a widely used parametric form.
It expresses the loss after training as the sum of an irreducible loss floor and two power-law terms, capturing the effects of model capacity and training data size, respectively:
\begin{equation}
  \label{eq:chinchilla}
  L(N,D)
  =
  e_{\infty}
  +
  \frac{A}{N^{\alpha}}
  +
  \frac{B}{D^{\beta}}.
\end{equation}
Here, $N$ denotes the number of model parameters and $D$ denotes the number of training tokens.
The term $e_{\infty}$ represents the asymptotic loss floor, while the coefficients $A$ and $B$, together with the exponents $\alpha$ and $\beta$, are fitted constants that characterize the decay rates along the two scaling axes.

\textbf{Inference-time scaling laws for VLMs.}
Such scaling laws have recently been extended to VLM inference.
One such law models performance as a multiplicative function of language backbone size $\nlin$ and the total number of inference tokens $T$~\citep{li2024inference},
\begin{equation}
  \label{eq:li-law}
  L(N,T)
  =
  \frac{A}{N^{\alpha}}\cdot\frac{B}{T^{\beta}}
  +
  D,
\end{equation}
where $T$ counts visual tokens retained by a learned compression module at a fixed input resolution.
An analogous law has been fit for video VLMs, where the sampled frame count determines the token budget and the model is fine-tuned for each configuration before evaluation~\citep{wang2025inference}.

\subsection{Experimental Setup}
\label{sec:exp-setup}

\textbf{Models and benchmarks.}
We evaluate five model series from the QwenVL and InternVL families, with backbone sizes ranging from $1$B to $72$B.
These include Qwen2.5-VL (3B, 7B, 32B, 72B)~\citep{bai2025qwen25vltechnicalreport}, Qwen3-VL (2B, 4B, 8B, 32B)~\citep{bai2025qwen3}, InternVL2.5 (1B, 2B, 4B, 8B, 26B, 38B)~\citep{chen2024expanding}, InternVL3 (1B, 2B, 8B, 9B, 14B, 38B)~\citep{zhu2025internvl3}, and InternVL3.5 (1B, 2B, 4B, 8B, 14B, 38B)~\citep{wang2025internvl3}.
We count only the language backbone in $\nlin$ and leave out the vision encoder and projector, although larger models within a series often differ in those components as well.
We test these models on four benchmarks built from high-resolution images, namely HR-Bench~\citep{wang2025divide}, MME-RealWorld~\citep{zhang2024mme}, TreeBench~\citep{wang2025traceable}, and V\textsuperscript{*}\,Bench~\citep{wu2024v}.
To vary the visual input, we resize each image to one of several preset sizes and leave the model's own visual processor unchanged.

\textbf{Units of analysis.}
Each evaluation gives a pair $(\nlin,\rlin)$ on one benchmark, which we call a cell.
We set the input image size and measure $\rlin$, the number of visual tokens that enter the language backbone.
All evaluations use greedy decoding, and the error $L$ of a cell is one minus its mean multiple-choice accuracy.
Each of the 26 models is measured under 25 combinations of benchmark and image size, giving 650 cells and 536{,}781 valid evaluations.
Pairing each model series with each benchmark gives 20 groups, which serve as the units for fitting our Separable Law.

\subsection{Our Separable Inference Scaling Law}
\label{sec:sep-law}

Existing inference-time laws~\citep{li2024inference,wang2025inference} establish that downstream error responds to both model scale and the visual token budget, yet their functional form is assumed rather than tested.
Eq.\eqref{eq:li-law} couples the two axes multiplicatively by construction, and a separable alternative is never fitted or compared against it.
Neither law is tested on pretrained image VLMs when the input image size itself changes.
Moreover, these studies change the token count by adding a learned compression module or retraining the model for each setting, so their fits mix the effect of more visual tokens with the effect of changing the model itself.

Our formulation instead starts from an accounting observation.
For a VLM answering a visual query, the token stream consumed by the language backbone is the inference-time counterpart of the training data $D$ in Eq.\eqref{eq:chinchilla}, and it is overwhelmingly visual.
A single high-resolution image expands into thousands of visual tokens and dwarfs the accompanying textual query.
We therefore replace $D$ directly with the number of visual tokens visible to the backbone and propose a law of the same form, with capacity and visual evidence contributing separately as they did for parameters and data:
\begin{equation}
  \label{eq:sep-law}
  L(\nlin,\rlin)
  =
  \einf
  +
  \frac{A}{\nlin^{\alpha}}
  +
  \frac{B}{\rlin^{\beta}}.
\end{equation}
Here, $L$ denotes downstream error, $\nlin$ is the number of parameters in the language backbone, and $\rlin$ is the number of visual tokens.
The parameters $\alpha$ and $\beta$ are the capacity and visual-token scaling exponents, respectively, governing how the corresponding error terms decay as $\nlin$ and $\rlin$ increase.
We call Eq.\eqref{eq:sep-law} the Separable Law, since the contributions of $\nlin$ and $\rlin$ enter as two terms that add rather than multiply, and we use this name for it throughout the paper.

\subsection{Validating the Separable Form}
\label{sec:fit-law}

Before interpreting the fitted components of Eq.\eqref{eq:sep-law}, we check how well it fits and whether the data favor the separable form over coupled alternatives.

\begin{figure}[t]
  \centering
  \includegraphics[width=0.95\linewidth]{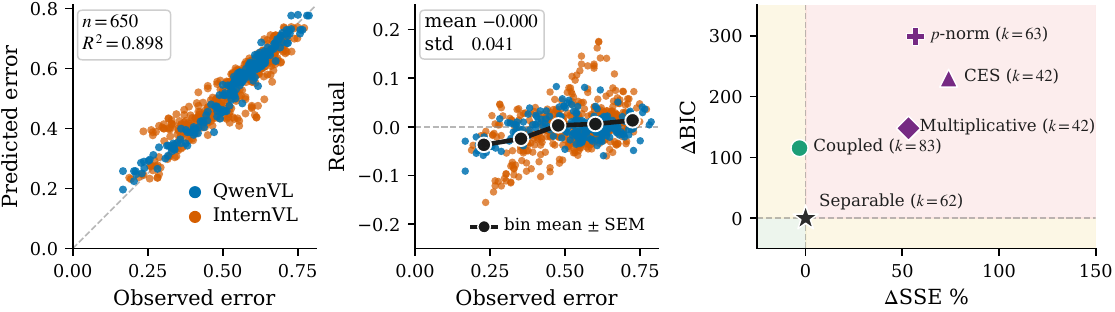}
  \caption{\textbf{Fit quality and form comparison for the Separable Law.}
  \textbf{Left:} predicted vs.\ observed error, one dot per $(\nlin,\rlin)$ cell, with QwenVL in blue, InternVL in orange, and the line $y{=}x$.
  \textbf{Middle:} residual vs.\ observed error, with mean $\pm$ standard error of the mean (SEM) in each of five error bins.
  \textbf{Right:} candidate forms in $\Delta$-space relative to the separable form, marked by a charcoal star at the origin.
  The axes show $\Delta$SSE\,\% and $\Delta$BIC, and $k$ is the nominal number of free parameters.}
  \label{fig:validation-fit-quality}
\end{figure}

\textbf{The law fits.}
We fit the five parameters of Eq.\eqref{eq:sep-law} separately for each of the 20 $(\text{series},\text{benchmark})$ groups and pool the predictions across 650 cells.
The pooled coefficient of determination ($R^2$) of $0.898$ indicates that the law captures the observed error surface well (Figure~\ref{fig:validation-fit-quality}, left).
Residuals are centered at zero with a standard deviation of $0.041$, showing little overall bias but some variation across cells.
The near-zero mean nevertheless masks a systematic trend, as mean residuals rise from $-0.037$ in the lowest observed-error bin to $+0.014$ in the highest (Figure~\ref{fig:validation-fit-quality}, middle).
Fit quality also varies across groups and is associated with the observed token span within each group.
Overall, the fit remains good across the whole grid, with no region where the law clearly breaks down.

\textbf{The data favor the separable form among the tested candidates.}
A good fit alone does not establish separability, so we compare Eq.\eqref{eq:sep-law} with alternative forms.
These include the multiplicative law (Eq.\eqref{eq:li-law}), written for group $d$ as $L_d = D_d + K_d\,\nlin^{-\alpha}\rlin^{-\beta}$, and a coupled form that adds the term $C_d(\nlin\rlin)^{-\gamma}$ to Eq.\eqref{eq:sep-law}, with group-specific $C_d\ge0$.
We also test two standard ways of letting the two resources interact.
A constant-elasticity-of-substitution (CES) form~\citep{arrow1961capital} aggregates $\nlin$ and $\rlin$ through a single fitted parameter that controls how far one resource can substitute for the other.
A $p$-norm form aggregates the two error terms and the floor as a generalized mean, so that as $p$ grows error is increasingly set by whichever resource is scarcest.
All candidates are fit on the same 650 cells and share their exponents across the 20 groups.
We score each candidate by the sum of squared errors (SSE) over those cells, which measures how closely the fitted surface tracks the observed error, and by the Bayesian information criterion (BIC), which adds a penalty for the number of fitted parameters so that a richer form is not favored merely for having more of them.
Lower values are better for both scores.
Separable and coupled forms use matched unconstrained group intercepts and nonnegative amplitudes.
The multiplicative form has $53.2\%$ higher SSE than the separable form and a higher BIC despite fewer parameters (Figure~\ref{fig:validation-fit-quality}, right).
Only the coupled form reduces SSE, from $1.221$ to $1.182$, but raises BIC by $115.2$ due to its additional parameters.
None of the other candidates earns its extra parameters, so we use the Separable Law throughout.

\section{Finding 1: The Cognitive Ceiling}
\label{sec:ceiling}

\begin{finding}
\textbf{Finding 1:}
Scaling helps most questions but not all.
About a third of them stay wrong at every model size and visual-token count we evaluate, and scaling shifts them in both directions without settling them.
Which questions these are depends mainly on the visual skill a question demands, with the worst cases confined to particular pairings of skill and scene.
\end{finding}

We study scaling limits through the fitted floor $\einf$.
Because it is fitted to the average error over many questions, a task can end up with a floor close to zero while a few of its questions stay wrong at the largest model size and highest visual-token count we evaluate.
We therefore read this floor at three levels, from the perception and reasoning subsets these benchmarks provide (\S\ref{sec:einf-task}) to individual questions (\S\ref{sec:regime}) and to groups defined by visual scene and required skills (\S\ref{sec:DS-granularity}).

\subsection{\texorpdfstring{$\einf$ is Higher for Reasoning Tasks Than for Perception Tasks}{E-inf is Higher for Reasoning Tasks Than for Perception Tasks}}
\label{sec:einf-task}

\begingroup
\setlength{\intextsep}{2pt}
\begin{wraptable}{r}{0.48\textwidth}
\centering
\captionsetup{font=footnotesize,position=top,skip=3pt,justification=raggedright,singlelinecheck=false}
\caption{Results of per-task $\einf$ across 40 fits.}
\label{tab:einf-task}

\footnotesize
\renewcommand{\arraystretch}{1.12}
\setlength{\tabcolsep}{3pt}

\begin{tabular}{@{}lcccc@{}}
\toprule
\rowcolor{gray!12}
\textbf{Task type}
& \makecell{\textbf{Mean}\\$\einf$}
& \makecell{\textbf{Median}\\$\einf$}
& \textbf{Floor}
& \textbf{Saturation} \\
\midrule

\textbf{Perception}
& 0.073
& 0.001
& \textbf{72\%}
& 20\% \\

\textbf{Reasoning}
& \textbf{0.155}
& 0.001
& 53\%
& \textbf{33\%} \\

\bottomrule
\end{tabular}
\end{wraptable}

% \textbf{A saturated minority, not a uniform shift.}
% Using the benchmarks' native perception and reasoning subsets, we fit Eq.\eqref{eq:sep-law} separately for each of the five model series, obtaining 40 $(\text{series}, \text{benchmark}, \text{task})$ fits.
% Reasoning has a higher mean $\einf$ than perception ($0.155$ vs.\ $0.073$), yet both medians lie at the lower search bound of $0.001$.
% We define \emph{Floor} as $\einf \le 0.01$ and \emph{Saturation} as $\einf \ge 0.20$, distinguishing near-zero floors from substantial residual error.
% In Table~\ref{tab:einf-task}, $72\%$ of perception fits and $53\%$ of reasoning fits meet the Floor criterion, while $20\%$ and $33\%$, respectively, meet Saturation.
% These fits suggest that saturation is concentrated in a minority of task subsets, with reasoning more often affected.
% This pattern suggests a \emph{cognitive ceiling}: persistent error that joint increases in $N$ and $R$ do little to reduce over the tested range.

\textbf{A minority of task subsets keeps a high floor.}
We start from the split these benchmarks already provide, labeling each question as perception or reasoning, and ask whether the fitted floor $\einf$ differs between the two.
We fit Eq.\eqref{eq:sep-law} separately for each of the five model series on these subsets, obtaining 40 $(\text{series}, \text{benchmark}, \text{task})$ fits.
Here $\einf$ is the error Eq.\eqref{eq:sep-law} approaches as $\nlin$ and $\rlin$ both grow without bound, so a high fitted $\einf$ means that neither a larger backbone nor more visual tokens brings that task subset near zero error.
We call such persistent error a \emph{cognitive ceiling} and flag it with \emph{Saturation}, $\einf \ge 0.20$, as opposed to \emph{Floor}, $\einf \le 0.01$, where the fitted error instead decays to essentially zero.
In Table~\ref{tab:einf-task}, $20\%$ of perception fits and $33\%$ of reasoning fits meet Saturation, while $72\%$ and $53\%$ respectively meet Floor.
The saturated fits also account for the higher mean $\einf$ on reasoning, $0.155$ against $0.073$, since both medians sit at the lower search bound of $0.001$.
These fits are aggregates over whole task subsets and hide individual questions that never improve, so we next ask which questions sit behind them.

\par\endgroup
\begin{figure}[t]
  \centering
  \includegraphics[width=\linewidth]{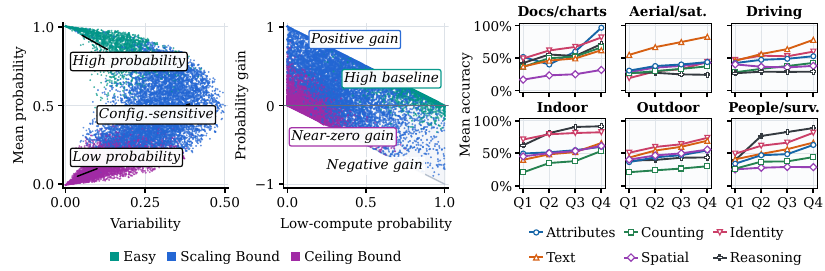}
  \vspace{-1em}
  \caption{
\textbf{Question-level scaling heterogeneity and responses by domain and skill.}
\textbf{Left:} Mean probability of the correct option versus its variability across $(N,R)$ configurations.
\textbf{Middle:} Probability gain $\bar p_{Q_4}-\bar p_{Q_1}$ versus its low-compute value $\bar p_{Q_1}$.
\textbf{Right:} Accuracy across compute quartiles, with one panel per visual domain and one curve per skill.
}
\vspace{-1em}
  \label{fig:DS-einf}
\end{figure}

\subsection{Sample-Level Regime Discovery}
\label{sec:regime}
% \textbf{Scaling regimes cut across native labels.}
% We analyze 19{,}890 $(\text{series}, \text{benchmark}, \text{item ID})$ trajectories across the available $(N,R)$ configurations.
% Based on mean accuracy across equally weighted configurations, $20.0\%$ are \textsc{Easy} ($>0.90$), $49.4\%$ are \textsc{Scaling Bound} ($0.20$--$0.90$), and $30.7\%$ are \textsc{Cognitive Ceiling} ($<0.20$).
% The probability maps reveal contrasting scaling responses (Figure~\ref{fig:DS-einf}): \textsc{Cognitive Ceiling} trajectories cluster at low gold-answer probabilities (left), with a mean probability gain of only $0.014$ from the lowest to the highest within-trajectory compute quartile, compared with $0.239$ for \textsc{Scaling Bound} trajectories (middle).
% Thus, scaling gains concentrate in a responsive subset, while a substantial minority remains largely unresolved over the tested range.
% These regimes cut across native perception and reasoning labels: reasoning has a higher ceiling rate ($42.2\%$ vs.\ $26.5\%$), yet most reasoning trajectories remain \textsc{Scaling Bound} or \textsc{Easy}.
% Ceiling rates also vary across benchmarks, from $18\%$--$20\%$ in HR-Bench and V\textsuperscript{*}\,Bench to $39\%$--$41\%$ in MME-RealWorld and TreeBench.
% These differences motivate a finer-grained domain--skill analysis.
\textbf{Scaling response separates questions that the perception and reasoning labels group together.}
The fits above describe whole task subsets, so we now move to individual questions and ask which constitute the ceiling.
We analyze 19{,}890 trajectories, one per question and model series.
Using mean accuracy across configurations, we sort each trajectory into three regimes: $20.0\%$ are \textsc{Easy}, above $0.90$ accuracy; $49.4\%$ are \textsc{Scaling Bound}, from $0.20$ to $0.90$; and $30.7\%$ are \textsc{Ceiling Bound}, below $0.20$.
For each trajectory we record the probability the model assigns to the correct option, comparing its mean over the lowest and highest compute quartiles.
\textsc{Ceiling Bound} trajectories cluster at low probabilities (Figure~\ref{fig:DS-einf}, left) and gain only $0.014$ between the two quartiles, against $0.239$ for \textsc{Scaling Bound} trajectories (Figure~\ref{fig:DS-einf}, middle).
Gains therefore concentrate in the responsive \textsc{Scaling Bound} subset, whereas \textsc{Ceiling Bound} trajectories move in both directions, $32.1\%$ gaining more than $0.05$ and $27.8\%$ losing more than $0.05$.
These regimes also cut across the perception and reasoning labels, since reasoning has a higher \textsc{Ceiling Bound} rate, $42.2\%$ against $26.5\%$, yet most reasoning trajectories still respond to scaling.

\subsection{Domain and Skill Granularity}
\label{sec:DS-granularity}

\definecolor{domainbg}{RGB}{235,243,250}
\definecolor{domainfg}{RGB}{50,95,135}
\definecolor{skillbg}{RGB}{250,240,230}
\definecolor{skillfg}{RGB}{155,90,45}

\begingroup
\setlength{\intextsep}{2pt}
\begin{wraptable}{r}{0.60\textwidth}
  \centering
  \captionsetup{font=footnotesize,position=top,skip=3pt,justification=raggedright,singlelinecheck=false}
  \caption{Unified $6\times6$ domain and skill taxonomy.}
  \label{tab:taxonomy}

  \footnotesize
  \setlength{\tabcolsep}{2pt}
  \renewcommand{\arraystretch}{1.0}

    \begin{tabular*}{\linewidth}{@{\extracolsep{\fill}}llll@{}}
      \toprule

      \multicolumn{2}{c}{
        \cellcolor{domainbg}
        \textcolor{domainfg}{\textbf{Domain $D$}}
      }
      &
      \multicolumn{2}{c}{
        \cellcolor{skillbg}
        \textcolor{skillfg}{\textbf{Skill $S$}}
      }
      \\

      \midrule

      \textcolor{domainfg}{\textbf{D1}}
      & Documents, Charts, Slides
      & \textcolor{skillfg}{\textbf{S1}}
      & Attribute Recognition \\

      \textcolor{domainfg}{\textbf{D2}}
      & Aerial, Satellite
      & \textcolor{skillfg}{\textbf{S2}}
      & Text Reading \\

      \textcolor{domainfg}{\textbf{D3}}
      & Vehicles, Driving
      & \textcolor{skillfg}{\textbf{S3}}
      & Counting \\

      \textcolor{domainfg}{\textbf{D4}}
      & Indoor
      & \textcolor{skillfg}{\textbf{S4}}
      & Spatial Relation \\

      \textcolor{domainfg}{\textbf{D5}}
      & Outdoor
      & \textcolor{skillfg}{\textbf{S5}}
      & Object Identification \\

      \textcolor{domainfg}{\textbf{D6}}
      & People, Surveillance
      & \textcolor{skillfg}{\textbf{S6}}
      & Scene Reasoning \\

      \bottomrule
    \end{tabular*}
\end{wraptable}

\textbf{Skill captures the broad pattern, with extremes in specific domains.}
Locating the ceiling requires question labels that are comparable across the four benchmarks, so we re-label each unique question under a unified $6\times6$ taxonomy of visual domain $D$ and required skill $S$ (Table~\ref{tab:taxonomy}).
Domain describes the visual context, while skill specifies the operation needed to answer the question.
For each supported $(D,S)$ pairing, we pool observations across the five model series and fit a separate instance of Eq.\eqref{eq:sep-law}, obtaining 33 fitted floors.
The domain panels in Figure~\ref{fig:DS-einf} (right) show accuracy across four compute quartiles, from the lowest quartile $Q_1$ to the highest quartile $Q_4$, and the skill holding the lowest curve changes from one domain to the next, so the same operation is not uniformly difficult.
A two-factor decomposition of the 33 floors attributes $36.67\%$ of their variance to skill and $7.84\%$ to domain, placing the broad pattern on the required operation.
Marginal means nevertheless miss the most severe pairings, since Aerial/Satellite with Scene Reasoning has a fitted floor of $0.780$, against domain and skill marginals of $0.227$ and $0.228$, and the same pairing also holds the lowest observed accuracy in that domain from the second quartile on.
At the other end, most Text Reading and Object Identification pairings reach the lower search bound of $0.001$, and the two highest point estimates in this group, $0.080$ and $0.180$, both stay below the Saturation threshold.
Skill therefore gives the broad picture, while the most severe ceilings can be identified only when both the domain and the skill are specified.

\par\endgroup

\section{Finding 2: Architectural Divergence}
\label{sec:arch-divergence}

{\begin{finding}
\textbf{Finding 2:}
The gains from a larger language backbone and from more visual tokens depend on the architecture.
The two families agree on capacity but differ by more than an order of magnitude on visual tokens, because their frontends turn the same pixel budget into very different token counts.
Which of the two axes pays off faster reverses between them, so an allocation tuned on one architecture need not carry over to another.
\end{finding}}

Above the floor studied in Finding~1, error falls at rates set by the two exponents of Eq.\eqref{eq:sep-law}, $\alpha$ for capacity and $\beta$ for visual tokens.
We compare them across the two families (\S\ref{sec:beta-arch}), trace their difference to the visual frontends (\S\ref{sec:beta-mech}), and examine the reversal in their ordering (\S\ref{sec:ab-ratio}).

\subsection{Capacity exponents agree across families, visual-token exponents do not}
\label{sec:beta-arch}

\textbf{Capacity exponents differ less than visual-token exponents.}
For the 20 $(\text{series}, \text{benchmark})$ groups, we search over $\alpha,\beta\in[0.05,10]$ to reduce fitting instability under limited data.
We compare only the fits whose exponents land inside this range, since an exponent stopping at a bound is a lower limit rather than an estimate.
For $\alpha$, twelve InternVL and five of eight QwenVL groups qualify, with medians of $0.19$ and $0.59$ (Figure~\ref{fig:sec7-combined}b).
The three QwenVL groups that do not qualify stop at the upper bound, and adding them back raises the QwenVL median to $0.78$ (Table~\ref{tab:per-group-fits}).
The architectural difference is therefore more pronounced in visual-token scaling than in capacity scaling.

\textbf{Visual-token exponents differ in both magnitude and boundary behavior.}
For $\beta$, the cross-family ordering reverses: the three interior InternVL estimates, all from HR-Bench, have a median of $4.66$, versus $0.17$ for QwenVL (Figure~\ref{fig:sec7-combined}b).
These medians cover all eight QwenVL groups but only three of twelve InternVL groups, as the remaining nine reach the upper bound of $10$ and carry the family's largest $\beta$ values, so the gap quoted here is conservative.
To understand this contrast, we next examine how far the token ranges of the two visual frontends actually extend.

\subsection{Discrete Tiling and Continuous Resizing Expose Different Token Ranges}
\label{sec:beta-mech}

\begin{figure}[t]
\centering
\includegraphics[width=0.95\linewidth]{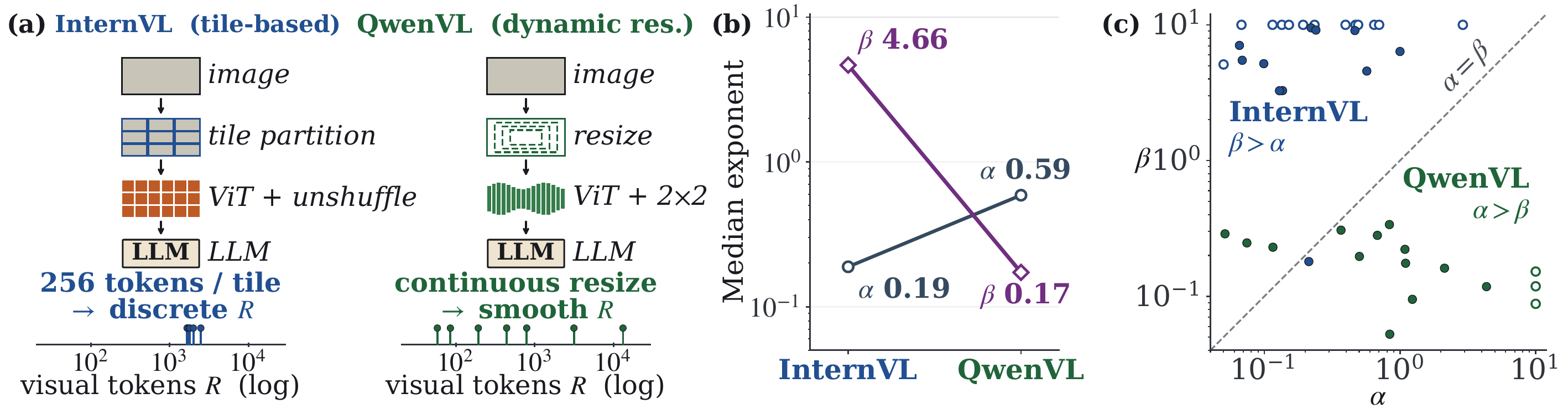}
% \vspace{-1em}
\caption{
\textbf{Visual frontends expose different token ranges and scaling exponents.}
\textbf{(a)} InternVL uses discrete tiling, whereas QwenVL uses continuous resizing.
\textbf{(b)} Median exponents reverse, with $\beta>\alpha$ for InternVL and $\alpha>\beta$ for QwenVL.
\textbf{(c)} Task-level fits in the $(\alpha,\beta)$ plane, where filled markers denote interior estimates and hollow markers denote boundary-clipped ones.
}

\vspace{-1em}
\label{fig:sec7-combined}
\end{figure}

\textbf{Discrete tiling exposes a short token range.}
InternVL picks a tile grid of at most twelve cells by matching the aspect ratio of the input, and uses area only to break ties~\citep{chen2024far}.
It then resizes the image to that grid and cuts it into $448\times448$ tiles, each yielding $256$ visual tokens that enter the LLM after pixel unshuffle.
Pixel count therefore barely enters the choice of grid, and raising the pixel budget from $224$ to $2048$\,px leaves the tile count unchanged for $70.2\%$ of questions, so the within-group token span has a median of only $1.22\times$ (Figure~\ref{fig:sec7-combined}a).
Such a narrow range cannot pin down how fast the error decays, which is why nine InternVL $\beta$ estimates stop at the upper bound.

\textbf{Continuous resizing exposes a much longer token range.}
QwenVL resizes the whole image to the target pixel count and reads it with 2D RoPE~\citep{wang2024qwen2}.
A $2\times2$ MLP merger compresses every four patches into one visual token, so the token count follows the pixel count continuously rather than in blocks.
Raising the pixel budget over the same $224$ to $2048$\,px range therefore moves the within-group token count by $34.09\times$ to $87.79\times$, and the full grid covers roughly 37 to 15,000 visual tokens (Figure~\ref{fig:sec7-combined}a).
Over such a wide range, the decay is visible, and all eight QwenVL groups return interior $\beta$ estimates with a median of $0.17$, far below the steep values InternVL reaches.

\subsection{\texorpdfstring{The $\alpha/\beta$ Ordering Reverses Across Architectures}{The alpha/beta Ordering Reverses Across Architectures}}
\label{sec:ab-ratio}
\textbf{On the evaluated grids, the exponent balance reverses across architectures.}
The ratio $\alpha/\beta$ compares the two axes within one architecture, with a value above $1$ meaning capacity decays faster and below $1$ meaning visual tokens do.
We compute it on the 40 task-level fits and keep those with both exponents in the search interior, which retains most QwenVL fits but fewer than half of the InternVL ones.
The median ratio is $2.55$ for QwenVL and $0.0395$ for InternVL, and the two families separate clearly in Figure~\ref{fig:sec7-combined}c.
Almost all of the dropped InternVL fits stop at the upper bound on $\beta$, so $0.0395$ is an upper bound on the true median.
The larger exponent therefore corresponds to capacity in QwenVL and to visual tokens in InternVL, which we next turn into an allocation rule.
\section{Finding 3: From Laws to Practice}
\label{sec:laws-practice}

\begin{finding}
\textbf{Finding 3:}
The best way to spend a fixed budget depends on how much error each axis removes per unit of compute.
Combined with a cost law, our law directs extra compute toward a larger backbone for InternVL and toward more visual tokens for QwenVL.
The option our law ranks first performs nearly as well as the best available option, even when it was held out from the fit.
\end{finding}

We first fit a cost law alongside the Separable Law and solve the two together for a continuous allocation (\S\ref{sec:vpe-decision}), then ask how to choose among the configurations a deployment actually offers (\S\ref{sec:vpe-validation}), and finally turn the same quantities around to describe the benchmarks themselves (\S\ref{sec:bench-diag}).

\subsection{From Laws to a Deployment Decision}
\label{sec:vpe-decision}

\textbf{A fitted cost law for the two axes.}
Since $\nlin$ and $\rlin$ are both defined where the visual tokens enter the language backbone, we follow \citet{li2024inference} in measuring allocation cost by the FLOPs the language backbone spends on the prompt, the visual tokens plus the question text.
This gives a common basis across the two frontends, and for one cell of the grid the cost is
\begin{equation}
    C_{\mathrm{cell}}
    = 2P_{\mathrm{LLM}}T_{\mathrm{prompt}}
    = 2P_{\mathrm{LLM}}(\rlin+T_0),
    \label{eq:cost-proxy}
\end{equation}
where $P_{\mathrm{LLM}}$ is the parameter count of the language backbone, $T_{\mathrm{prompt}}$ is the length of the prompt, and $T_0$ comprises the question, template, and special tokens.
The vision encoder is left out of this count, so the allocations are optimal under a language-side budget.
Cost is linear in $\nlin$ at fixed $\rlin$, but doubling $\rlin$ does not double the cost, since $T_0$ does not change with $\rlin$.
Eq.\eqref{eq:cost-proxy} is therefore not an exact power law in $(\nlin,\rlin)$, but a power law approximates it over the observed grid, so we take
\begin{equation}
    C(\nlin,\rlin)
    = \kappa \nlin^{p}\rlin^{q},
    \label{eq:cost}
\end{equation}
where $\kappa$ sets the scale and $p$ and $q$ are the fitted cost elasticities for capacity and visual tokens.
We use it as the budget constraint below.
Fitting it over the same 650 cells gives $p=1.00$, matching the linearity above, together with $q=0.77$ and log-space $R^2=0.995$.

\textbf{Visual--Parameter Exchange balances return against cost.}
Given the error and cost laws, we now ask how a fixed budget should be split between the two axes. This gives the optimization problem
\begin{equation}
    (\nlin^*,\rlin^*)
    = \argmin_{\nlin,\rlin>0} L(\nlin,\rlin)
    \quad \text{subject to} \quad
    \kappa\nlin^p\rlin^q\le C_0.
    \label{eq:vpe-problem}
\end{equation}
Its solution follows from two conditions.
The first places it on the budget boundary $\kappa\nlin^p\rlin^q=C_0$, since all fitted coefficients and exponents are positive, so error falls along both axes and an optimum never leaves budget unspent.
The second fixes where on that boundary it sits, because if one axis returned more error reduction per unit of spending than the other, shifting spending toward it would lower error further, so at the optimum the two returns match,
\begin{equation}
    \frac{\alpha A}{p}(\nlin^*)^{-\alpha}
    = \frac{\beta B}{q}(\rlin^*)^{-\beta}.
    \label{eq:vpe-balance}
\end{equation}
Both conditions are linear in $\log \nlin$ and $\log \rlin$, so the pair is a $2\times2$ system with determinant $p\beta+q\alpha$ and right-hand sides $\log \bar C$ and $\log K$, where $\bar C=C_0/\kappa$ and $K=\alpha A q/(\beta B p)$, and solving gives
\begin{equation}
    \nlin^*
    = \bar C^{\frac{\beta}{p\beta+q\alpha}}
      K^{\frac{q}{p\beta+q\alpha}},
    \qquad
    \rlin^*
    = \bar C^{\frac{\alpha}{p\beta+q\alpha}}
      K^{-\frac{p}{p\beta+q\alpha}}.
    \label{eq:vpe-closed}
\end{equation}
We call this the Visual--Parameter Exchange (VPE) allocation, named for the rate at which the two axes trade against each other at the optimum, and it is the unique optimum under the fitted cost law.
Since a deployment moves between budgets rather than sitting at one, we differentiate Eq.\eqref{eq:vpe-closed},
\begin{equation}
    \frac{\partial \log \nlin^*}{\partial \log C_0}
    = \frac{\beta}{p\beta+q\alpha},
    \qquad
    \frac{\partial \log \rlin^*}{\partial \log C_0}
    = \frac{\alpha}{p\beta+q\alpha}.
    \label{eq:vpe-elasticity}
\end{equation}
Which axis grows faster is decided by whether $\beta$ or $\alpha$ is larger, so the reversal of Finding~2 makes InternVL add capacity first and QwenVL add visual tokens first (Figure~\ref{fig:vpe}a).
Only the exponents enter these two rates, so the direction of expansion is more robust than the point it starts from.

\subsection{Choosing Among Available Configurations}
\label{sec:vpe-validation}

\begin{figure*}[t]
\centering
\includegraphics[width=0.95\linewidth]
{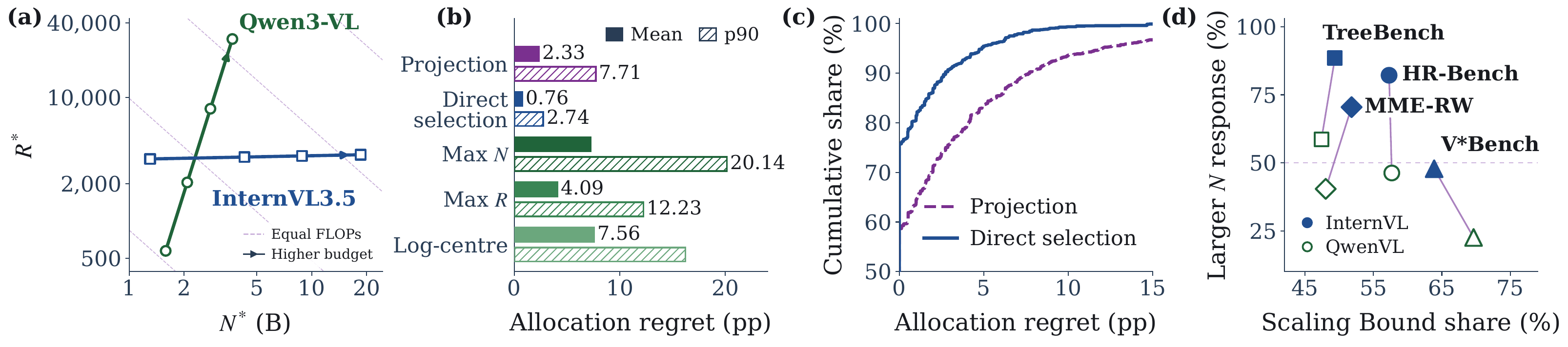}
\caption{\textbf{Allocation and benchmark diagnostics.}
\textbf{(a)} VPE trajectories $(\nlin^{*},\rlin^{*})$ for one series per family, arrows marking increasing budget.
\textbf{(b)} Mean (solid) and p90 (hatched) regret under joint holdout.
\textbf{(c)} Regret CDFs for projection and direct selection.
\textbf{(d)} Per benchmark and family, the \textsc{Scaling Bound} share against the share of those questions responding more to $\nlin$ than to $\rlin$.
}
\label{fig:vpe}
\label{fig:bench_diag}
\end{figure*}

\textbf{The fitted law transfers to configurations it never saw.}
A deployment picks from a finite catalogue, so the continuous optimum must become a choice among available configurations.
Projection takes the continuous optimum, pulls $\nlin^*$ and $\rlin^*$ back into the observed range, and selects the nearest feasible configuration in log coordinates.
Direct selection skips the continuous optimum and instead evaluates Eq.\eqref{eq:sep-law} at every feasible configuration, taking the lowest.
To keep training and evaluation apart, we drop one backbone size and one pixel budget from the performance fit and score allocation only on the configurations left out.
Across $2{,}379$ such decisions, projection carries a mean regret of $2.33$ pp and a $90$th percentile of $7.71$ pp.
Three static rules that make no use of Eq.\eqref{eq:sep-law}---taking the largest backbone, the largest token count, or the log-centre of the range allowed by the budget---have mean regrets from $4.09$ to $7.56$ pp and $90$th-percentile regrets from $12.23$ to $20.14$ pp (Figure~\ref{fig:vpe}b).
Direct selection is better still, at $0.76$ pp and $2.74$ pp, and the regret CDF shows the same ordering across thresholds (Figure~\ref{fig:vpe}c).
Both read the same law over the same candidates, so this gap comes from the projection step.
VPE still earns its place by describing how the balance shifts as the budget grows, whereas direct selection is what we would apply to a fixed catalogue.

\subsection{What a Benchmark Separates and Which Axis It Rewards}
\label{sec:bench-diag}

\textbf{The same analysis can also be used to examine a benchmark.}
Only the \textsc{Scaling Bound} questions of a benchmark separate models, since the other two regimes give the same answer for every model.
Among these questions, we call one $\nlin$-dominated when its accuracy moves further along the backbone axis than along the visual token axis.
The dominant axis, however, depends on the frontend, because InternVL's token count barely moves with image size, so InternVL shows a larger $\nlin$-dominated share than QwenVL on every benchmark (Figure~\ref{fig:vpe}d).
Even so, both frontends rank the four benchmarks the same way, so a benchmark can still be assessed by how many of its questions separate models and whether they call for a larger backbone or more visual tokens.

\section{Related Work}
\textbf{Inference scaling and compute allocation in VLMs.}
How VLMs should spend inference compute has drawn growing attention as their inputs grow larger.
Approaches range from making visual tokens cheaper through nested token representations, adaptive patching, and efficient encoders~\citep{cai2025matryoshka,liu2026one,vasu2025fastvlm} to choosing the input resolution per task~\citep{luo2025task,kimhi2026cares} and requesting a sharper view only when needed~\citep{yang2026visionthink,lee2026ergo}.
Test-time methods add a further axis, spending compute on search, verification, or renewed looking~\citep{snell2024scaling,wang2025scaling,bai2026multi,avogaro2026sparc}.
Closest to our goal, scaling analyses relate accuracy to model size and visual-token count~\citep{li2024inference,du2025exploring,wang2025inference,li2025scaling}, with the token count largely set by learned compression, frame sampling, or retraining.
Our work fits a law over backbone size and the visual-token count produced at each input size, across frontends that process images differently.

\textbf{Perceptual bottlenecks and uneven scaling across questions.}
Understanding where VLMs fail has moved from aggregate scores toward finer diagnosis.
Diagnostic benchmarks locate failures in fine-grained perception~\citep{tong2024eyes,fu2024blink,wu2024v,zhang2024mme,wang2025divide}, and adaptive perception methods address them by searching or zooming into the region that holds the answer~\citep{wu2024v,wang2025retrieval,shen2025zoomeye,liu2026perceptual}.
Scaling gains also differ across capabilities in contrastive VLMs~\citep{al2024unibench}.
Furthermore, question-level analyses track confidence over training~\citep{swayamdipta2020dataset}, separate model ability from question difficulty~\citep{truong2026item}, or show that per-question gains and losses offset as video compute grows~\citep{sun2026stable}.
Our work follows each question as the backbone and visual tokens grow together, and relates those that stay wrong to the skill they require.

\section{Conclusion}
\label{sec:conclusion}
In this work, we propose the Separable Law, which describes how VLM performance changes with language backbone size and visual token count.
We fit it on four high-resolution benchmarks with input sizes from $224$ pixels to 8K, where balancing the two matters most.
We find that about a third of questions remain out of reach at every scale we test, and that more visual tokens pay off far more for QwenVL than for InternVL.
We hope our work offers a principle for allocating limited inference compute in high-resolution settings, applicable beyond the specific models and benchmarks we test.
\begingroup

\section*{AI Use Statement}
In this work, we use generative AI tools for two purposes: to support qualitative data analysis and to edit this paper for grammar, wording, and readability.
For the first purpose, we use GPT-4o-mini to assign domain and required-skill labels to benchmark questions, and we then inspect these assignments against the taxonomy and decision rules of Appendix~\ref{app:taxonomy}.
For the second purpose, we check every AI-assisted edit against the underlying results.
In both cases, we review all AI-assisted work, and we take responsibility for the final content of this work, including text, claims, or artifacts produced with the aid of generative AI.

\section*{Reproducibility Statement}
We evaluate publicly released models without modifying them in any way.
For the evaluation itself, Section~\ref{sec:exp-setup} and Appendix~\ref{app:setup} describe the models, benchmarks, input sizes, and the evaluation protocol, while Appendix~\ref{app:setup-fitting} specifies how the law is fitted.
For the analyses built on it, Appendix~\ref{app:taxonomy} documents the domain and skill annotation, while Appendix~\ref{app:vpe} documents the allocation and regret evaluation.
Since all models and benchmarks are publicly available, the evaluation grid can be reconstructed from these descriptions.
\par
\endgroup
\bibliography{iclr2027_conference}
\bibliographystyle{iclr2027_conference}
\newpage
\appendix
% Avoid isolated paragraph openings and endings across appendix pages.
\clubpenalty=10000
\widowpenalty=10000
\displaywidowpenalty=10000

\section*{Appendix Contents}
\startcontents[appendix]
\begingroup
\small
\printcontents[appendix]{}{1}{}
\par\endgroup
\clearpage

\section{Extended Experimental Setup}
\label{app:setup}

This section traces each measurement from a model and image to a cell-level error rate.
We describe the visual frontends and evaluated models (Appendix~\ref{app:models-frontends}), the benchmarks (Appendix~\ref{app:benchmarks}), and how input size determines the visual token count (Appendix~\ref{app:setup-benchmarks-grid}).
We then define each experimental cell, corresponding to one model, input size, and benchmark, and report the scale of the study (Appendix~\ref{app:setup-units}), before detailing decoding and scoring (Appendix~\ref{app:setup-decoding-scoring}).

\subsection{Models and Visual Frontends}
\label{app:models-frontends}

Our experiments cover 26 models in five InternVL and QwenVL series.
InternVL produces a fixed number of tokens per tile, whereas QwenVL varies the token count with input size.
This distinction motivates the comparison of $\beta$ in \S\ref{sec:beta-arch}. Appendices~\ref{app:family-internvl} and~\ref{app:family-qwen} describe the frontends, and Appendix~\ref{app:eval-models} lists the models and defines $\nlin$ as the parameter count of the language backbone.

\subsubsection{InternVL Family}
\label{app:family-internvl}
Since InternVL 1.5, the InternVL series has turned an image into visual tokens in the same way, by cutting it into tiles of a fixed size.
Later versions mainly changed the training recipe and the efficiency of inference.
Table~\ref{tab:internvl-evo} summarizes these changes from InternVL 1.0 to InternVL3.5.

\begin{table}[htbp]
\appTableStyle
\centering
\small
\caption{Evolution of the InternVL series, focusing on how each version turns an image into tokens.}
\label{tab:internvl-evo}
\resizebox{\linewidth}{!}{%
\begin{tabular}{llll}
\toprule
\rowcolor{appHeader}
\appHeadFont Version & \appHeadFont Input strategy & \appHeadFont Visual token mechanism & \appHeadFont Effect on visual token count \\
\midrule
InternVL 1.0 & Fixed input size & ViT features projected to the language backbone & Fixed count \\
\rowcolor{appStripe}
InternVL 1.5 & Dynamic tiling & Up to 40 tiles of $448{\times}448$ & Changes in whole tiles \\
InternVL2.5 & Dynamic tiling & Pixel unshuffle to 256 tokens per tile & Changes in whole tiles \\
\rowcolor{appStripe}
InternVL3 & Dynamic tiling & 256 tokens per tile, V2PE positions & Changes in whole tiles \\
InternVL3.5 & Dynamic tiling & 256 tokens per tile, ViR only in Flash variants & Changes in whole tiles \\
\bottomrule
\end{tabular}
}
\end{table}

\paragraph{Establishing the tiling mechanism (InternVL 1.0$\to$InternVL 1.5).}
InternVL 1.0 used inputs of a fixed size and encoded them with a vision transformer (ViT)~\citep{chen2024internvl}.
InternVL 1.5 introduced dynamic tiling, which splits an image into as many as 40 tiles of $448{\times}448$ pixels~\citep{chen2024far}.
The ViT encodes each tile separately, and the resulting features are concatenated before they enter the language backbone.
InternVL 1.5 also introduced pixel unshuffle, which compresses the 1,024 ViT output tokens of each tile into 256 visual tokens.
All later versions keep this design.

\paragraph{Scaling up and changing the training recipe (InternVL2.5$\to$InternVL3).}
InternVL2.5 offered a systematic range of model sizes and improved performance through stronger data filtering and more training data~\citep{chen2024expanding}.
InternVL3 replaced the usual two-stage recipe, in which a ViT is attached to a pretrained language model, with native multimodal pre-training~\citep{zhu2025internvl3}.
The language backbone and InternViT start from pretrained weights, and native multimodal pre-training then trains them jointly on text-only and multimodal data.
InternVL3 also introduced Variable Visual Position Encoding (V2PE) to handle long multimodal contexts.

\paragraph{Efficiency features (InternVL3.5).}
InternVL3.5 adds two features for efficient inference~\citep{wang2025internvl3}.
The Visual Resolution Router (ViR), used in the InternVL3.5-Flash variants, chooses a compression rate for the visual tokens of each image.
Decoupled Vision-Language Deployment (DvD) places the ViT and the language backbone on separate GPUs.
We evaluate the standard InternVL3.5 models without ViR, so every tile yields 256 visual tokens, as in InternVL2.5 and InternVL3.

\paragraph{Tiling setting in our evaluation.}
We allow at most 12 tiles per image in all three InternVL series.
When an image is split into more than one tile, the model also receives a smaller copy of the whole image as one extra tile.
An image therefore uses at most 13 tiles, or 3,328 visual tokens.
The tile grid is chosen to match the aspect ratio of the input, so for most questions the number of tiles stays the same across our input sizes (\S\ref{sec:beta-mech}).

\subsubsection{QwenVL Family}
\label{app:family-qwen}
The QwenVL series moved from a fixed input size to dynamic resolution in Qwen2-VL, which resizes the whole image and produces a visual token count that changes with the image size.
Later versions kept this frontend and changed the vision encoder and the training.
Table~\ref{tab:qwenvl-evo} summarizes these changes from Qwen-VL to Qwen3-VL.

\begin{table}[htbp]
\appTableStyle
\centering
\small
\caption{Evolution of the QwenVL series, focusing on how each version turns an image into tokens.}
\label{tab:qwenvl-evo}
\resizebox{\linewidth}{!}{%
\begin{tabular}{llll}
\toprule
\rowcolor{appHeader}
\appHeadFont Version & \appHeadFont Input strategy & \appHeadFont Visual token mechanism & \appHeadFont Effect on visual token count \\
\midrule
Qwen-VL & Fixed $448{\times}448$ input & Cross-attention adapter to a fixed number of tokens & Fixed count \\
\rowcolor{appStripe}
Qwen2-VL & Dynamic resolution & 2D-RoPE and $2{\times}2$ MLP patch merger & Changes with input size in small steps \\
Qwen2.5-VL & Dynamic resolution & Window attention, same patch merger & Changes with input size in small steps \\
\rowcolor{appStripe}
Qwen3-VL & Dynamic resolution & SigLIP-2, DeepStack, same patch merger & Changes with input size in small steps \\
\bottomrule
\end{tabular}
}
\end{table}

\paragraph{From fixed to dynamic resolution (Qwen-VL$\to$Qwen2-VL).}
The original Qwen-VL resized every image to $448{\times}448$ and compressed the ViT output into a fixed number of visual tokens through a cross-attention adapter~\citep{bai2023qwen}.
Input size therefore had little effect on what reached the language backbone.
Qwen2-VL redesigned this frontend in three ways~\citep{wang2024qwen2}.
It replaced absolute position embeddings with two-dimensional rotary position embedding (2D-RoPE), which lets the ViT accept images of any size.
It compressed tokens with a $2{\times}2$ MLP patch merger instead of cross-attention.
It introduced Multimodal Rotary Position Embedding (M-RoPE), which splits position information into time, height, and width so that one model handles both images and video.
As a result, the visual token count changes with the input size in small steps set by the patch grid, much finer than the whole-tile steps of InternVL.

\paragraph{Balancing efficiency and scale (Qwen2-VL$\to$Qwen2.5-VL).}
Dynamic resolution makes the ViT expensive on large inputs.
Qwen2.5-VL reduced this cost with window attention and kept the dynamic resolution frontend and the patch merger~\citep{bai2025qwen25vltechnicalreport}.
It also trained the ViT from scratch and expanded the pre-training data.
The way input size maps to visual token count stays the same as in Qwen2-VL.

\paragraph{Deeper fusion of vision and language (Qwen2.5-VL$\to$Qwen3-VL).}
Qwen3-VL replaced the custom ViT with SigLIP-2 and added three architectural changes~\citep{bai2025qwen3}.
DeepStack feeds ViT features from several depths into the matching layers of the language backbone through light residual connections.
An interleaved version of M-RoPE spreads the time, horizontal, and vertical position components across frequency bands.
Explicit text timestamp tokens mark time in video.
The dynamic resolution frontend and the $2{\times}2$ MLP merger stay in place, so the visual token count still changes with the input size in small steps.

\subsubsection{Evaluated Models}
\label{app:eval-models}

Table~\ref{tab:model-summary} lists the 26 models we evaluate, and the same definition of $\nlin$ applies to both families.

\paragraph{Definition of $\nlin$.}
$\nlin$ is the exact parameter count of the language backbone of each model.
We compute it from the architecture specification, exclude the vision encoder and projector, and do not rely on the size in the model name.
Within a series, larger models may also change the vision encoder, the training data, and the training recipe, so differences in $\nlin$ carry these changes as well.

\begin{table}[htbp]
\appTableStyle
\centering
\small
\caption{Evaluated model series and their visual frontends, 26 models in total. Sizes follow the model names, and $\nlin$ is computed as described above.}
\label{tab:model-summary} % renamed from tab:checkpoint-summary
\begin{tabular}{lllc}
\toprule
\rowcolor{appHeader}
\appHeadFont Family & \appHeadFont Series & \appHeadFont Evaluated sizes & \appHeadFont Visual frontend \\
\midrule
InternVL & InternVL2.5 & 1B, 2B, 4B, 8B, 26B, 38B & Dynamic tiling \\
\rowcolor{appStripe}
InternVL & InternVL3 & 1B, 2B, 8B, 9B, 14B, 38B & Dynamic tiling \\
InternVL & InternVL3.5 & 1B, 2B, 4B, 8B, 14B, 38B & Dynamic tiling \\
\rowcolor{appStripe}
QwenVL & Qwen2.5-VL & 3B, 7B, 32B, 72B & Dynamic resolution \\
QwenVL & Qwen3-VL & 2B, 4B, 8B, 32B & Dynamic resolution \\
\bottomrule
\end{tabular}
\end{table}

\subsection{Benchmarks}
\label{app:benchmarks}

We evaluate on four benchmarks built from high-resolution images.

\paragraph{HR-Bench.}
HR-Bench tests fine-grained perception on images at 4K and 8K resolution~\citep{wang2025divide}.
It targets details that are often lost when an image is downsampled, and it has two subtasks, perception of a single instance and perception across instances.
The 4K and 8K releases contain the same 800 questions.

\paragraph{MME-RealWorld.}
MME-RealWorld is a large benchmark, annotated by humans, that tests multimodal models on high-resolution images of real-world scenes~\citep{zhang2024mme}.
It contains 13,366 images and more than 29,000 multiple-choice questions.
The full benchmark has 43 subtasks in five fields, namely OCR, remote sensing, charts and tables, monitoring, and autonomous driving.
We use the MME-RealWorld Lite split, which contains 1,782 unique question IDs in our evaluation pipeline.

\paragraph{TreeBench.}
TreeBench is a diagnostic benchmark for visual reasoning grounded in the image~\citep{wang2025traceable}.
Its questions focus on small targets in cluttered scenes and ask about spatial hierarchy, interactions between objects, and visual evidence that can be traced in the image.
It contains 405 questions on high-resolution images from the SA-1B collection.

\paragraph{V\textsuperscript{*}\,Bench.}
V\textsuperscript{*}\,Bench tests whether a model can locate and reason about tiny details in crowded scenes~\citep{wu2024v}.
It has two question types, recognizing object attributes and judging spatial relations.
It contributes 191 questions in our evaluation pipeline.

\begin{table}[htbp]
\appTableStyle
\caption{Native image sizes and evaluated input sizes.}
\label{tab:bench-res}
\centering
\small
\begingroup
\renewcommand{\arraystretch}{1.22}
\setlength{\tabcolsep}{6pt}
\arrayrulecolor{black!70}
\begin{tabular}{@{}lcc@{}}
\toprule
\rowcolor{appHeader}
\appHeadFont Benchmark & \appHeadFont Native size ($W \times H$) & \appHeadFont Input sizes (long edge, px) \\
\midrule
\textbf{HR-Bench-4K} & $\approx 4023 \times 3503$ & $224, 336, 512, 768, 1024, 2048$ \\
\rowcolor{appStripe}
\textbf{HR-Bench-8K} & $\approx 5727 \times 4430$ & $4096$ \\
\textbf{MME-RealWorld} & $\approx 2000 \times 1500$ & $224, 336, 512, 768, 1024, 2048$ \\
\rowcolor{appStripe}
\textbf{TreeBench} & $\approx 2152 \times 1615$ & $224, 336, 512, 768, 1024, 2048$ \\
\textbf{V\textsuperscript{*}\,Bench} & $\approx 2246 \times 1582$ & $224, 336, 512, 768, 1024, 2048$ \\
\bottomrule
\end{tabular}
\endgroup
\end{table}

\subsection{Input Sizes and Visual Token Counting}
\label{app:setup-benchmarks-grid}

\paragraph{Input sizes.}
We resize every image so that its longer edge equals one of a fixed set of input sizes, keeping the aspect ratio.
For an input size $b$ and a native size $(H,W)$, the scale factor is $s=b/\max(H,W)$ and the resized image has size $(H',W')=\mathrm{round}(sH,sW)$.
All four benchmarks share six input sizes, namely $224$, $336$, $512$, $768$, $1024$, and $2048$\,px.
For HR-Bench, the six shared sizes use the 4K release, and an additional $4096$\,px setting uses the separately released 8K split.
The two releases contain the same $800$ questions but assign them different question IDs.
We keep their observations separate for scoring and for building trajectories, and pair matching questions across the two releases when resampling images.
The input sizes cover three ranges.
The sizes $224$ and $336$\,px match standard ViT inputs, $512$ to $1024$\,px are common in deployment, and $2048$ and $4096$\,px test very high resolution.
At $1024$\,px and below almost every image is shrunk, while at $2048$\,px many MME-RealWorld images are smaller than the target and are enlarged.
We also ran each benchmark at native size but leave this setting out of all fits.
At native size the longer edge differs from image to image, so the input size is not fixed and $\rlin$ cannot be compared across questions.
Without it, the grid has $650$ cells across the five model series.

\paragraph{Measured visual token counts.}
The same input size produces different numbers of visual tokens in different models, so we measure $\rlin$ instead of computing it from the input size.
Every record logs the number of visual tokens that enter the language backbone, and the $\rlin$ of a cell is the mean over its records.
For QwenVL we read this count from the image tokens in the processed input, after the patch merger.
For InternVL3 and InternVL3.5 we count the image placeholder tokens that the model inserts into the prompt, one for each visual token.
For InternVL2.5 we multiply the logged tile count, including the extra tile for the whole image, by 256 tokens per tile.
These measurements also reveal how strongly each frontend changes its visual token count with input size.

\subsection{Units of Analysis and Scale of the Study}
\label{app:setup-units}

\paragraph{Cells and groups.}
A cell is one model evaluated at one input size on one benchmark, and it is the unit of fitting.
For bootstrap resampling, the unit is an image together with all its questions across every evaluated configuration, since different cells share the same images.
Each benchmark is run at six input sizes, and HR-Bench adds the $4096$\,px setting, which gives $4\times6+1=25$ combinations of benchmark and input size.
The $26$ models therefore give $26\times25=650$ cells.
The cells form $20$ series--benchmark groups.
A QwenVL group contains $24$ cells, or $28$ on HR-Bench, and an InternVL group contains $36$, or $42$ on HR-Bench.
Each group varies along two axes.
Along $\nlin$ it covers the backbone sizes of the series, six for each InternVL series and four for each QwenVL series, and along $\rlin$ it covers six or seven input sizes.

\paragraph{Scale of the study.}
We recorded $537{,}940$ generations, of which $691$ failed during inference.
Of the remaining $537{,}249$ scored records, $468$ lack usable probabilities for the correct option, which leaves $536{,}781$ records with both signals across the $650$ cells.
The question-level analysis in \S\ref{sec:regime} uses all scored records and keeps the last record when a question ID repeats within a configuration.
This gives $515{,}938$ observations that form $19{,}890$ trajectories, where a trajectory is one question evaluated by one model series across its configurations.
Each series has $3{,}978$ trajectories, from $1{,}600$ HR-Bench question IDs, $1{,}782$ MME-RealWorld questions, $405$ TreeBench questions, and $191$ V\textsuperscript{*}\,Bench questions.
The $1{,}600$ HR-Bench IDs consist of $800$ from the 4K release, evaluated at the six shared sizes, and $800$ from the 8K release, evaluated only at $4096$\,px.

\subsection{Decoding and Scoring}
\label{app:setup-decoding-scoring}

\paragraph{Decoding.}
We run the public instruction-tuned models in BF16 without modifying them.
Prompting is zero-shot, and the system and user prompts ask the model to answer with a single capital option letter.
Decoding is greedy, meaning the model always takes the most probable next token without sampling or beam search, so the temperature setting has no effect.
Of the $104$ runs, one for each model and benchmark, $96$ allow at most $48$ new tokens with a repetition penalty of $1.05$.
The other $8$ runs, Qwen2.5-VL 3B, 7B, and 32B and InternVL2.5-38B on HR-Bench and MME-RealWorld, allow at most $32$ new tokens with no repetition penalty.
In total, $429$ generations reach the token limit, which is about $0.08\%$ of all generations.
Of these, $304$ reach $48$ tokens and $125$ reach $32$ tokens, and $51$ contain no option letter that the parser can extract.

\paragraph{Scoring.}
Each scored record carries two predictions.
The text prediction is parsed from the generated answer, and the logit prediction is the option with the highest logged confidence.
The parser first looks for an answer in parentheses or after an explicit answer phrase, and otherwise takes the last valid option letter that stands alone.
Option confidences are normalized over the scored answer labels, and the decoding step at which they are read follows each model's implementation.
The two predictions agree on $518{,}644$ of the $536{,}781$ records with both signals, or $96.6\%$, so we use text accuracy as the main signal.
The error in Eq.\eqref{eq:sep-law} is $L=1-\text{text accuracy}$, computed for each cell over its scored records.
The question-level robustness analysis also uses the logged probability of the correct option, $p_{\rm gold}$.
We use this normalized score as a relative measure and do not assume that it is calibrated.
For chance normalization we use the actual number of options of each question, after checking that the prompt and the log agree.
This normalization leaves the error used in the Separable fits unchanged.

\section{Fitting and Validating the Separable Law}
\label{app:validation}

{We describe the fitting procedure and group-level diagnostics (Appendices~\ref{app:setup-fitting} and~\ref{app:val-fit-diagnostics}), then compare the Separable law with coupled alternatives (Appendix~\ref{app:val-formrace}).
We test robustness to image resampling, precision weighting, and held-out configurations (Appendices~\ref{app:separability-uncertainty} and~\ref{app:val-cv}).
Because some edge predictions fall outside $[0,1]$, we also fit a bounded law and verify that the decisions and exponent ordering are unchanged (Appendix~\ref{app:bounded-error-law}).
Finally, we quantify parameter uncertainty (Appendix~\ref{app:val-boxsens}) and test sensitivity to the HR-Bench $4096$\,px setting and cell-level token aggregation (Appendix~\ref{app:val-without-hr4096}).}

\subsection{Fitting Procedure}
\label{app:setup-fitting}

\paragraph{Normalization.}
{We divide both axes by a common reference point, $n=\nlin/\widetilde N$ and $r=\rlin/\widetilde R$, where $\widetilde N$ and $\widetilde R$ are the geometric means of $\nlin$ and $\rlin$ over all $650$ cells.
We write the fitted amplitudes at this reference point as $A_n$ and $B_r$, so they give the error contributed by each axis at the same point and can be compared across groups.
The coefficients $A$ and $B$ of Eq.\eqref{eq:sep-law}, which the VPE allocation in \S\ref{sec:vpe-decision} uses, follow as $A=A_n\widetilde N^{\alpha}$ and $B=B_r\widetilde R^{\beta}$, with $\nlin$ in billions of parameters as in the cost law.
The choice of reference point changes only $A_n$ and $B_r$, and it leaves the exponents, the floor, every predicted error, and the VPE allocation unchanged.}

\paragraph{{Objective.}}
{We fit each group separately by minimizing the unweighted squared error over its cells, so that each configuration counts once,}
\begin{equation}
\min_{\alpha,\beta,e_\infty,A_n,B_r}\;
\sum_{i\in g}\Bigl[L_i-e_\infty-A_n\,n_i^{-\alpha}-B_r\,r_i^{-\beta}\Bigr]^2,
\qquad A_n,B_r\ge 0.
\end{equation}

\paragraph{{Optimization.}}
{We optimize the exponents over $(\log\alpha,\log\beta)$ with the Nelder--Mead method, starting from 16 points given by $\alpha_0\in\{0.1,0.3,0.6,1.0\}$ and $\beta_0\in\{0.05,0.15,0.3,0.6\}$.
At each step, we search $e_\infty$ over a grid of 55 values in $[0.001,0.999]$.
For each candidate $e_\infty$, the nonnegative amplitudes $A_n$ and $B_r$ have a closed-form solution, which we find by checking the four combinations of amplitudes held at zero or left free.
A soft quadratic penalty keeps both exponents within $[0.05,10]$.
Nelder--Mead uses a step of $0.4$, at most $300$ iterations, and tolerances $\mathrm{ftol}=10^{-8}$ and $\mathrm{xtol}=10^{-7}$.
This objective weights all configurations equally, and its point estimates serve as the baseline for the uncertainty and sensitivity analyses.}

\subsection{Per-Group Fits and Diagnostics}
\label{app:val-fit-diagnostics}
\label{app:per-group-fits}

\paragraph{Per-group parameters.}
{Table~\ref{tab:per-group-fits} reports the fitted parameters, SSE, and $R^2$ for all 20 groups.
Pooled over the 650 cells, $R^2$ is $0.898$ and SSE is $1.088$, and the median group has an SSE of $0.024$ and an $R^2$ of $0.892$.
These numbers describe the point fits, and Table~\ref{tab:parameter-bootstrap} gives bootstrap intervals for $\alpha$, $\beta$, and $e_\infty$.
A dagger marks the 9 groups whose $\beta$ reaches the upper bound of the search range, all from InternVL, and a double dagger marks the 3 QwenVL groups whose $\alpha$ reaches that bound.
Overall, the law provides a strong fit across the 20 groups.}

\begin{table}[htbp]
\appTableStyle
\centering
\small
\caption{{Per-group Separable-law point estimates and fit diagnostics. Amplitudes $A_n$ and $B_r$ are evaluated at the common reference point $\widetilde N=6.60$B and $\widetilde R=1{,}135$. $^{\dagger}$ and $^{\ddagger}$ mark $\beta$ and $\alpha$, respectively, at the upper bound of $10$; $n$ is the cell count.}}
\label{tab:validation-per-group-fits}
\label{tab:per-group-fits}
\renewcommand{\arraystretch}{1.06}
\setlength{\tabcolsep}{4.5pt}
\begin{tabular}{@{}llrrrrrrrr@{}}
\toprule
\rowcolor{appHeader}
\multicolumn{2}{c}{\appHeadFont Group} & \multicolumn{5}{c}{\appHeadFont Separable-law parameters} & \multicolumn{3}{c}{\appHeadFont Fit diagnostics} \\
\cmidrule(lr){1-2}
\cmidrule(lr){3-7}
\cmidrule(lr){8-10}
\rowcolor{appHeader}
\appHeadFont Series & \appHeadFont Benchmark & \appHeadFont $e_\infty$ & \appHeadFont {$A_n$} & \appHeadFont $\alpha$ & \appHeadFont {$B_r$} & \appHeadFont $\beta$ & \appHeadFont SSE & \appHeadFont $n$ & \appHeadFont $R^2$ \\
\midrule
InternVL2.5 & HR-Bench & 0.001 & 0.308 & 0.149 & 1.24 & 4.66 & 0.031 & 42 & 0.92 \\
InternVL2.5 & MME-RealWorld & 0.001 & 0.527 & 0.058 & 11.41 & $10.0^{\dagger}$ & 0.043 & 36 & 0.77 \\
InternVL2.5 & TreeBench & 0.001 & 0.582 & 0.057 & 9.65 & $10.0^{\dagger}$ & 0.030 & 36 & 0.73 \\
InternVL2.5 & {V\textsuperscript{*}\,Bench} & 0.001 & 0.089 & 0.381 & 56.31 & $10.0^{\dagger}$ & 0.203 & 36 & 0.59 \\
\addlinespace[1.5pt]
\rowcolor{appStripe}
InternVL3 & HR-Bench & 0.001 & 0.262 & 0.183 & 1.05 & 3.87 & 0.025 & 42 & 0.94 \\
\rowcolor{appStripe}
InternVL3 & MME-RealWorld & 0.36 & 0.120 & 0.194 & 12.48 & $10.0^{\dagger}$ & 0.043 & 36 & 0.76 \\
\rowcolor{appStripe}
InternVL3 & TreeBench & 0.06 & 0.506 & 0.071 & 8.57 & $10.0^{\dagger}$ & 0.013 & 36 & 0.87 \\
\rowcolor{appStripe}
InternVL3 & {V\textsuperscript{*}\,Bench} & 0.01 & 0.008 & 1.262 & 65.96 & $10.0^{\dagger}$ & 0.231 & 36 & 0.55 \\
\addlinespace[1.5pt]
InternVL3.5 & HR-Bench & 0.001 & 0.312 & 0.145 & 1.42 & 4.93 & 0.023 & 42 & 0.93 \\
InternVL3.5 & MME-RealWorld & 0.46 & 0.008 & 1.203 & 13.78 & $10.0^{\dagger}$ & 0.044 & 36 & 0.80 \\
InternVL3.5 & TreeBench & 0.50 & 0.054 & 0.418 & 11.96 & $10.0^{\dagger}$ & 0.010 & 36 & 0.83 \\
InternVL3.5 & {V\textsuperscript{*}\,Bench} & 0.001 & 0.057 & 0.465 & 62.21 & $10.0^{\dagger}$ & 0.258 & 36 & 0.51 \\
\midrule
\rowcolor{appStripe}
Qwen2.5-VL & HR-Bench & 0.001 & 0.043 & 0.462 & 0.372 & 0.123 & 0.021 & 28 & 0.93 \\
\rowcolor{appStripe}
Qwen2.5-VL & MME-RealWorld & 0.001 & $1.3{\times}10^{-6}$ & {$10.0^{\ddagger}$} & 0.599 & 0.075 & 0.010 & 24 & 0.93 \\
\rowcolor{appStripe}
Qwen2.5-VL & TreeBench & 0.24 & 0.261 & 0.053 & 0.092 & 0.187 & 0.008 & 24 & 0.81 \\
\rowcolor{appStripe}
Qwen2.5-VL & {V\textsuperscript{*}\,Bench} & 0.001 & $3.8{\times}10^{-6}$ & {$10.0^{\ddagger}$} & 0.324 & 0.205 & 0.041 & 24 & 0.91 \\
\addlinespace[1.5pt]
Qwen3-VL & HR-Bench & 0.001 & 0.036 & 0.756 & 0.343 & 0.162 & 0.023 & 28 & 0.94 \\
Qwen3-VL & MME-RealWorld & 0.08 & $1.5{\times}10^{-8}$ & {$10.0^{\ddagger}$} & 0.468 & 0.110 & 0.005 & 24 & 0.96 \\
Qwen3-VL & TreeBench & 0.40 & 0.033 & 0.800 & 0.121 & 0.184 & 0.003 & 24 & 0.95 \\
Qwen3-VL & {V\textsuperscript{*}\,Bench} & 0.001 & 0.057 & 0.589 & 0.230 & 0.323 & 0.023 & 24 & 0.96 \\
\bottomrule
\end{tabular}
\end{table}

\begin{figure}[t]
\centering
\includegraphics[width=\linewidth]{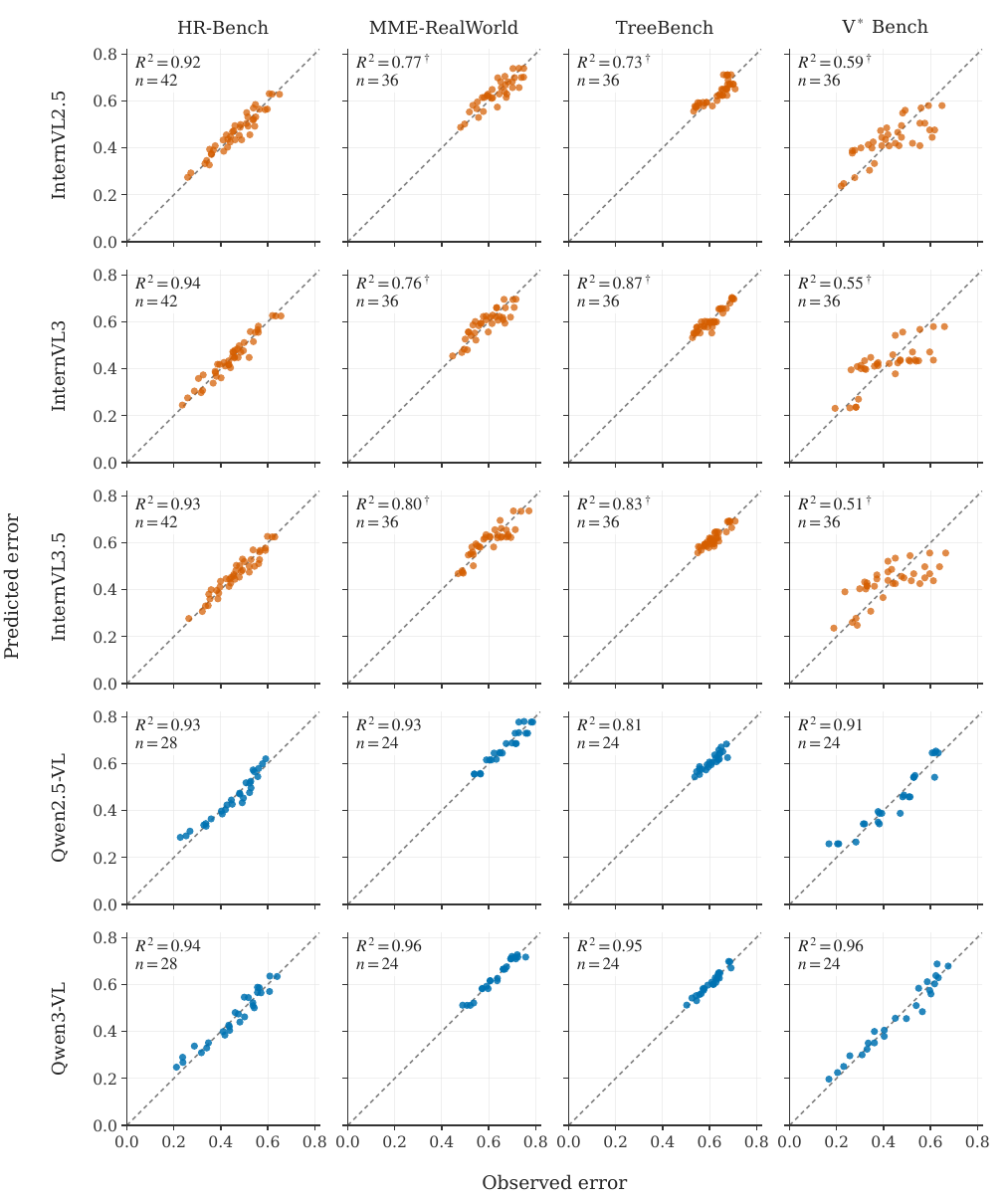}
\caption{{\textbf{Predicted versus observed error for each group.}
Rows correspond to the five model series and columns to the four benchmarks.
Each point is one $(\nlin,\rlin)$ cell, all panels share $[0,0.82]^2$ axes, and the dashed line marks perfect agreement, $y=x$.
Orange denotes InternVL and blue denotes QwenVL; each panel reports its $R^2$ and cell count $n$.
$^\dagger$ marks groups whose fitted $\beta$ reaches the search upper bound of $10$.
The three InternVL groups on V\textsuperscript{*}\,Bench have the lowest $R^2$ values, consistent with their narrow visual-token ranges.}}
\label{fig:validation-per-group-scatter}
\end{figure}

\paragraph{Predicted versus observed.}
{Figure~\ref{fig:validation-per-group-scatter} plots predicted against observed error for each group, with the same $[0,0.82]^2$ axes in every panel so that the spread can be compared directly.
Seventeen of the 20 groups reach $R^2\ge0.65$.
The other three are the V\textsuperscript{*}\,Bench groups of the three InternVL series, with $R^2$ of $0.594$, $0.548$, and $0.515$.
In these groups the measured visual token count spans the narrowest range, with the largest count only about $1.067$ times the smallest.
Thus, fit quality is high whenever the visual-token axis spans a substantive range.}

\begin{figure}[t]
\centering
\includegraphics[width=\linewidth]{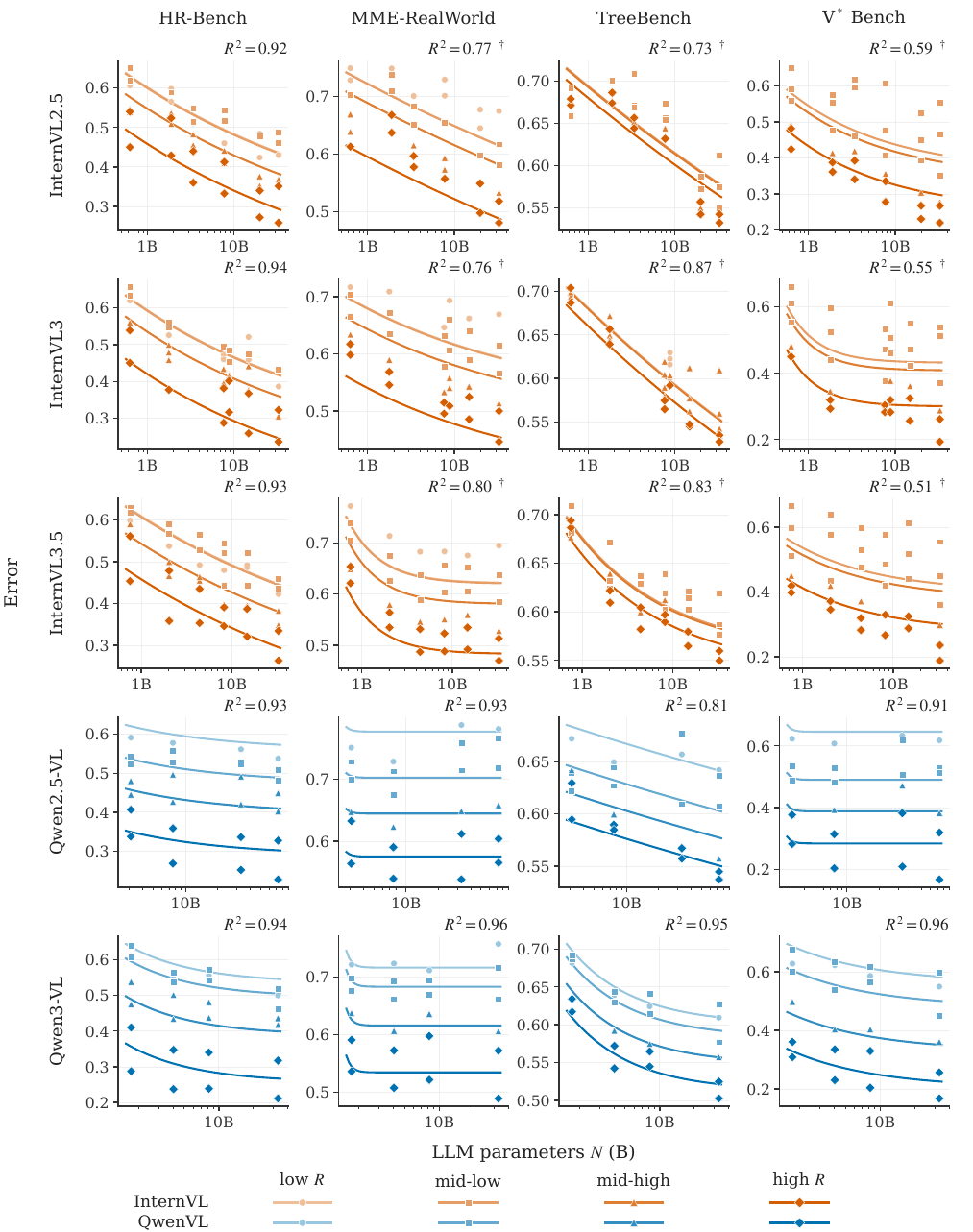}
\caption{{\textbf{Separable-law fits across model series and benchmarks.}
Rows show model series and columns show benchmarks; each panel plots observed error against $\nlin$ on a logarithmic scale.
Points are $(\nlin,\rlin)$ cells, and curves are predictions at the median $\rlin$ of each visual-token quartile; color denotes family and shade or marker denotes quartile.
$^\dagger$ marks $\beta=10$; Table~\ref{tab:per-group-fits} reports the parameter estimates.}}
\label{fig:validation-per-group-fits}
\end{figure}

\paragraph{{Fitted curves.}}
{Figure~\ref{fig:validation-per-group-fits} overlays the fitted Separable law on the observed errors of each group.
$\beta$ reaches its upper bound mainly in groups whose measured visual token count spans a narrow range.
The curves therefore capture the observed scaling trends across language backbone sizes and visual-token quartiles.}

\begin{figure}[t]
\centering
\includegraphics[width=0.85\linewidth]{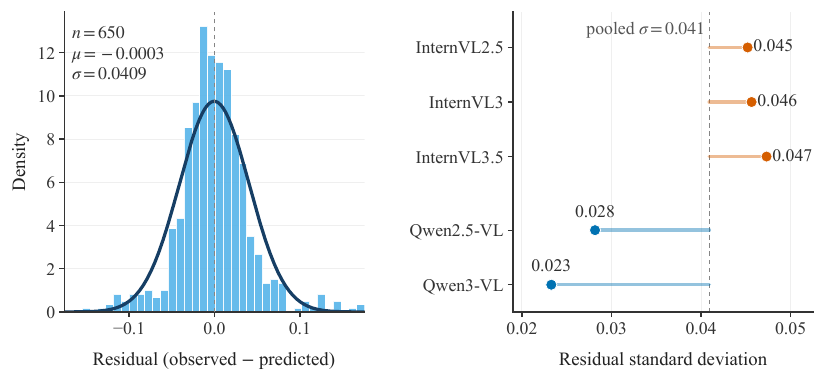}
\caption{{\textbf{Residual diagnostics for the per-group Separable-law fits.}
\textbf{Left:} Distribution of all $650$ residuals (observed minus predicted), with zero marked by the dashed line and a matched Gaussian shown for reference.
\textbf{Right:} Residual standard deviation by model series; the dashed line gives the pooled value.
Residuals are centered near zero, but InternVL shows greater dispersion than QwenVL.}}
\label{fig:validation-residual-diag}
\end{figure}

\paragraph{Residual diagnostics.}
{Figure~\ref{fig:validation-residual-diag} pools the residuals of all 650 cells.
They center near zero, with a mean of $-0.000$ and a standard deviation of $0.041$, but their tails are heavier than those of a Gaussian with the same mean and variance, with a sample excess kurtosis of $2.74$.
We show the Gaussian curve only as a visual reference, since the residuals of different cells need not be independent or identically distributed.
The residual standard deviation is $0.046$ for InternVL and $0.026$ for QwenVL, ranging from $0.045$ to $0.047$ across the three InternVL series and from $0.023$ to $0.028$ across the two QwenVL series.
Across five bins of observed error with equal width, the mean residuals are $-0.037$, $-0.024$, $+0.003$, $+0.006$, and $+0.014$.
The law therefore slightly overpredicts error in the lowest bin and underpredicts it in the highest, which is the trend noted in \S\ref{sec:fit-law}.
Overall, the residuals remain centered near zero across both families.}

\subsection{Comparison of Functional Forms}
\label{app:val-formrace}

\paragraph{Protocol.}
{To compare functional forms fairly, we fit every candidate to the same 650 cells, share its shape parameters across the 20 groups, and let the group coefficients vary.
This gives Separable 62 parameters and an SSE of $1.221$; fitting separate shape parameters to every group uses 100 parameters and reaches an SSE of $1.088$ and $R^2=0.898$.
Separable and Coupled use the same unconstrained group intercepts, nonnegative amplitudes, and exponents in $[0.05,10]$, with Coupled reducing to Separable when $C_g=0$.
We treat BIC as a descriptive comparison rather than a formal model-selection test.
This protocol isolates the effect of coupling under matched data and constraints.}

\paragraph{Candidate forms.}
The alternatives below differ in how they combine backbone capacity and visual tokens.
\begin{align}
\text{Multiplicative (Eq.\eqref{eq:li-law}):}\quad
   & e = e_{\infty,g}+K_g\,\nlin^{-\alpha}\rlin^{-\beta}
     \\[2pt]
\text{Coupled:}\quad
   & e = e_{\infty,g}+A_g\,\nlin^{-\alpha}+B_g\,\rlin^{-\beta}
     +C_g\,(\nlin\rlin)^{-\gamma}
     \\[2pt]
\text{CES:}\quad
   & e = e_{\infty,g}
     +K_g\bigl[w\,\nlin^{\delta}+(1-w)\,\rlin^{\delta}\bigr]^{-1/\delta}
     \\[2pt]
p\text{-norm:}\quad
   & e = \bigl[(A_g\,\nlin^{-\alpha})^p
              +(B_g\,\rlin^{-\beta})^p+C_g^{\,p}\bigr]^{1/p}.
\end{align}
The multiplicative form has the single product term of Eq.\eqref{eq:li-law}~\citep{li2024inference}, with nonnegative $K_g$.
Coupled adds to Separable a nonnegative interaction amplitude $C_g$ for each group and a shared exponent $\gamma$, which raises the parameter count from $62$ to $83$.
The CES form~\citep{arrow1961capital} combines the two axes with nonnegative $K_g$ and shared $w$ and $\delta$, searched over $0<w<1$ and $\delta\in[-8,4]$.
Multiplicative and CES use unconstrained group intercepts.
The $p$-norm form has nonnegative $A_g$, $B_g$, and $C_g$, with $\alpha,\beta\in[0.05,3]$ and $p\in[0.5,15]$.
Its $C_g$ is a nonnegative floor inside the norm and takes the place of the intercept.
At $p=1$ it reduces to a Separable law with a nonnegative intercept, and as $p\to\infty$ it approaches $\max\{A_g\nlin^{-\alpha},B_g\rlin^{-\beta},C_g\}$, where error is set by the scarcest resource.
Our fitter finds the group amplitudes by least squares on $e^p$ and scores the result on the original error scale.
The reported $p$-norm score is therefore the best fit this procedure finds, which may lie above the lowest SSE the $p$-norm form can reach.

\paragraph{Results.}
{Table~\ref{tab:robust} reports the parameter count, pooled SSE, and BIC of each form, and we compute every percentage difference in SSE from the unrounded values.
The multiplicative form has $53.2\%$ higher SSE and, although it uses $20$ fewer parameters, a BIC higher by $147.8$.
Its fitted exponents are $\alpha=0.05$, at the lower bound, and $\beta=0.40$.
Coupled is the only alternative with lower training SSE, reducing it from $1.221$ to $1.182$, or $3.1\%$, but its $21$ extra parameters raise BIC by $115.2$.
CES and the $p$-norm form have $73.9\%$ and $56.8\%$ higher SSE.
The fitted $p$ sits at its lower bound of $0.5$, far from the large values at which error is set by the scarcest resource.
Among the tested forms and under these constraints, the Separable law therefore gives the best balance between fit and parameter count.}

\begin{table}[htbp]
\appTableStyle
\centering
\small
\caption{{Comparison of functional forms on the same 650 cells, with exponents and shape parameters shared across groups. $k$ counts the parameters of each fitted law.}}
\label{tab:robust}
\begin{tabular}{lrrrrr}
\toprule
\rowcolor{appHeader}
\appHeadFont Form & \appHeadFont $k$ & \appHeadFont SSE & \appHeadFont BIC & \appHeadFont $\Delta$SSE & \appHeadFont $\Delta$BIC \\
\midrule
{Separable} & 62 & 1.221 & \appEmph{$-1834.3$} & --- & --- \\
\rowcolor{appStripe}
Multiplicative & 42 & 1.870 & $-1686.4$ & $+53.2\%$ & $+147.8$ \\
Coupled & 83 & \appEmph{1.182} & $-1719.0$ & $-3.1\%$ & $+115.2$ \\
\rowcolor{appStripe}
CES & 42 & 2.123 & $-1604.0$ & $+73.9\%$ & $+230.3$ \\
{$p$-norm} & 63 & 1.913 & $-1535.6$ & $+56.8\%$ & $+298.7$ \\
\bottomrule
\end{tabular}
\end{table}

\paragraph{Per-group multiplicative comparison.}
{Table~\ref{tab:multiplicative-per-group} refits the multiplicative form with exponents specific to each group, using the same exponent range, nonnegative amplitudes, and floor grid as the Separable fits.
It still fits worse than the Separable law in 19 of the 20 groups, with $15.2\%$ higher pooled SSE, $1.254$ against $1.088$, and a median group $R^2$ of $0.840$ against $0.892$.
The multiplicative form thus stays behind even when each group has its own exponents.}

\begin{table}[htbp]
\appTableStyle
\centering
\caption{{Separable and multiplicative laws with exponents specific to each group.}}
\label{tab:multiplicative-per-group}
\begin{tabular}{lrrr}
\toprule
\rowcolor{appHeader}
\appHeadFont Form & \appHeadFont Pooled SSE & \appHeadFont Median group $R^2$ & \appHeadFont {Groups with lower SSE} \\
\midrule
\rowcolor{appHighlight}
\appEmph{Separable} & \appEmph{1.088} & \appEmph{0.892} & \appEmph{19/20} \\
\rowcolor{appStripe}
Multiplicative  & 1.254 & 0.840 & 1/20 \\
\bottomrule
\end{tabular}
\end{table}

\subsection{Sampling Uncertainty and Precision Weighting}
\label{app:separability-uncertainty}

\paragraph{Paired image bootstrap.}
To test whether correlated questions change the comparison between Separable and Coupled, we resample images within each benchmark while preserving all associated questions, repeated records, configurations, and model series.
The $650$ cells are rebuilt from $536{,}781$ records covering $3{,}975$ question IDs, $3{,}175$ distinct questions, and $2{,}253$ images; matched HR-Bench questions receive the same weight in the 4K and 8K releases.
Every replicate recomputes the cell error and mean visual token count, and we run $500$ in-sample replicates and $200$ full held-out replicates with seed $20260909$.
Separable and Coupled use shared exponents, matched constraints, and verified optimization, with Coupled containing Separable at $C_g=0$.
Coupled lowers unweighted SSE by $3.15\%$, but the BIC difference is $115.23$, with a $95\%$ bootstrap interval of $[93.95,123.10]$; keeping only the last valid record per question and configuration gives the same preference.
The paired bootstrap therefore confirms that the preference for Separable is stable under correlated image sampling.

\paragraph{Precision-weighted comparison.}
To test whether unequal benchmark sizes affect the comparison, we give more precise cell estimates greater weight.
For cell $j$, let $a_{ij}$ be the number of valid records and $z_{ij}$ the number of errors contributed by image $i$, with $n_j=\sum_i a_{ij}$, $\widehat e_j=\sum_i z_{ij}/n_j$, and $m_s$ images in benchmark $s$.
We estimate the cell variance by
\begin{equation}
\label{eq:precision-weight-variance}
\widehat v_j = \frac{1}{n_j^2}\sum_s\frac{m_s}{m_s-1}
\left\{\sum_{i\in s}(z_{ij}-\widehat e_j a_{ij})^2
-\frac{\left[\sum_{i\in s}(z_{ij}-\widehat e_j a_{ij})\right]^2}{m_s}\right\}.
\end{equation}
We fit both forms by minimizing $Q=\sum_{j=1}^{650}w_j(\widehat e_j-\widehat L_j)^2$, where $w_j\propto\widehat v_j^{-1}$ is scaled to a mean of one and kept at its original value in every bootstrap refit.
The paired bootstrap retains correlation across configurations, and Table~\ref{tab:separability-bootstrap} summarizes both experiments.
Its BIC score combines the logarithm of the Coupled-to-Separable objective ratio over the $650$ cells with the penalty for Coupled's $21$ additional parameters.
Coupled lowers the weighted objective by $4.38\%$, while the BIC difference is $106.91$, with a $95\%$ bootstrap interval of $[89.65,116.90]$.
The precision-weighted experiment therefore confirms the preference for Separable when more reliable cells receive greater influence.

\begin{table}[htbp]
\appTableStyle
\centering
\small
\setlength{\tabcolsep}{7pt}
\renewcommand{\arraystretch}{1.15}

\caption{Paired image-bootstrap comparison with shared exponents ($500$ replicates). Cells give point estimates and $95\%$ percentile intervals; differences are Coupled minus Separable. $Q$ denotes unweighted or precision-weighted SSE, and positive $\Delta\mathrm{BIC}$ favors Separable.}
\label{tab:separability-bootstrap}
\begin{tabular}{@{}lccc@{}}
\toprule
\rowcolor{appHeader}
\appHeadFont Objective & \appHeadFont $\Delta Q$ & \appHeadFont Relative $\Delta Q$ (\%) & \appHeadFont $\Delta\mathrm{BIC}$ \\
\midrule
Unweighted
& \makecell[c]{$-0.0384$\\{\scriptsize $(-0.0823,-0.0277)$}}
& \makecell[c]{$-3.15$\\{\scriptsize $(-6.27,-1.97)$}}
& \cellcolor{appHighlight}\makecell[c]{$\mathbf{115.23}$\\{\scriptsize $(93.95,123.10)$}} \\
\rowcolor{appStripe}
Weighted by precision
& \makecell[c]{$-0.0435$\\{\scriptsize $(-0.0758,-0.0314)$}}
& \makecell[c]{$-4.38$\\{\scriptsize $(-6.88,-2.90)$}}
& \cellcolor{appHighlight}\makecell[c]{$\mathbf{106.91}$\\{\scriptsize $(89.65,116.90)$}} \\
\bottomrule
\end{tabular}
\end{table}

\subsection{Held-Out Prediction}
\label{app:val-cv}

\paragraph{Held-out configurations.}
To test whether the laws predict configurations not used for fitting, we leave out one backbone size ($22$ folds) or one input size ($7$ folds), refit Separable and Coupled under matched constraints, and predict the omitted cells using their measured visual-token counts.
We evaluate shared and group-specific exponents with nonnegative amplitudes, exponents in $[0.05,10]$, and continuous floors, and each of $200$ paired image bootstraps rebuilds the cells and repeats every fold without initialization from a full-data fit.
Table~\ref{tab:loo-cv} reports pooled RMSE over all $650$ cells with predictions clipped to $[0,1]$ and over the interior levels without clipping, comprising $400$ cells from $15$ backbone folds and $442$ cells from $5$ input-size folds.
With group-specific exponents, Separable has lower RMSE on the all-cell backbone holdouts and matches Coupled on the input-size holdouts; on interior configurations the Coupled-minus-Separable differences are only $-0.012$ and $-0.011$ percentage points.
Thus, at the group level used in the main analysis, the simpler Separable law predicts unseen configurations as well as or better than the Coupled law.

\begin{table}[htbp]
\appTableStyle
\centering
\small
\setlength{\tabcolsep}{6pt}
\renewcommand{\arraystretch}{1.12}

\caption{Held-out RMSE over $200$ paired image bootstraps. Differences are Coupled minus Separable, with $95\%$ percentile intervals below; positive values favor Separable. All-cell predictions are clipped to $[0,1]$, whereas interior predictions are not.}
\label{tab:loo-cv}
\begin{tabular}{@{}llccc@{}}
\toprule
\rowcolor{appHeader}
\appHeadFont Exponents & \appHeadFont Held-out axis & \appHeadFont Separable & \appHeadFont Coupled & \appHeadFont $\Delta$RMSE (pp) \\
\midrule
\multicolumn{5}{l}{\emph{All cells}} \\
\rowcolor{appStripe}
Shared & Backbone size & $0.0655$ & $0.0579$ & \makecell[c]{$-0.755$\\{\scriptsize $(-1.668,2.940)$}} \\
Shared & Input size & $0.0763$ & $0.0891$ & \makecell[c]{$1.283$\\{\scriptsize $(-0.412,2.388)$}} \\
\rowcolor{appStripe}
Per-group & Backbone size & $0.0577$ & $0.0781$ & \cellcolor{appHighlight}\makecell[c]{$\mathbf{2.040}$\\{\scriptsize $(0.488,5.111)$}} \\
Per-group & Input size & $0.0601$ & $0.0599$ & \cellcolor{appHighlight}\makecell[c]{$\mathbf{-0.028}$\\{\scriptsize $(-0.182,0.044)$}} \\
\midrule
\rowcolor{appStripe}
\multicolumn{5}{l}{\emph{Interior levels}} \\
Shared & Backbone size & $0.0608$ & $0.0512$ & \makecell[c]{$-0.962$\\{\scriptsize $(-2.004,-0.421)$}} \\
\rowcolor{appStripe}
Shared & Input size & $0.0606$ & $0.0544$ & \makecell[c]{$-0.614$\\{\scriptsize $(-1.403,0.661)$}} \\
Per-group & Backbone size & $0.0447$ & $0.0446$ & \cellcolor{appHighlight}\makecell[c]{$\mathbf{-0.012}$\\{\scriptsize $(-0.045,0.037)$}} \\
\rowcolor{appStripe}
Per-group & Input size & $0.0456$ & $0.0455$ & \cellcolor{appHighlight}\makecell[c]{$\mathbf{-0.011}$\\{\scriptsize $(-0.037,0.009)$}} \\
\bottomrule
\end{tabular}
\end{table}

\paragraph{Endpoint extrapolation.}
To distinguish interpolation from extrapolation beyond the observed grid, we inspect the unclipped predictions from folds that omit the smallest or largest level of an axis.
Predictions outside $[0,1]$ occur only in these endpoint folds, whereas every prediction from the interior folds remains within $[0,1]$.
The held-out analysis therefore gives a clean comparison on interior configurations.

\begin{figure}[t]
\centering
\includegraphics[width=\linewidth]{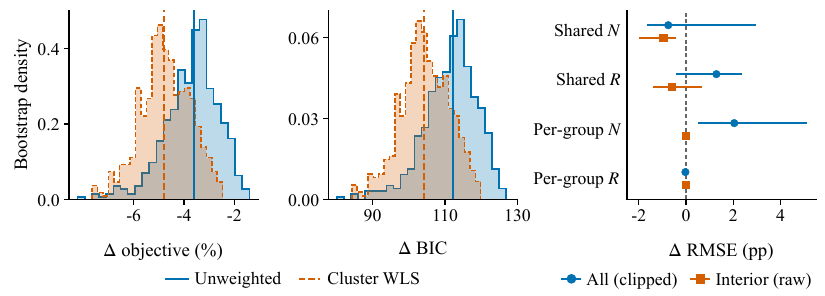}
\caption{{\textbf{Bootstrap comparison of Separable and Coupled laws.}
Left and middle show bootstrap objective and BIC differences for unweighted (blue) and precision-weighted (orange) fits; right shows held-out RMSE differences with $95\%$ intervals.
All differences are Coupled minus Separable: negative objective or RMSE differences favor Coupled, whereas positive BIC differences favor Separable.}}
\label{fig:separability-uncertainty}
\end{figure}

\subsection{A Bounded Version of the Law}
\label{app:bounded-error-law}

\paragraph{Bounded formulation.}
To test whether the main conclusions depend on unbounded error predictions, we replace the Separable output with a logistic link,
\begin{equation}
\eta_g(n,r)=c_g+A_g n^{-\alpha_g}+B_g r^{-\beta_g},
\qquad L^{\mathrm{bnd}}_{g}(n,r)=\sigma(\eta_g(n,r)),
\label{eq:bounded-logit-power}
\end{equation}
where $\sigma(x)=(1+\exp(-x))^{-1}$ and $n$ and $r$ use the main-fit normalization.
We fit the same $650$ cells with $A_g,B_g\ge0$, $\alpha_g,\beta_g\in[0.05,10]$, and $c_g\ge\operatorname{logit}(0.001)$ by least squares on the logit of observed error.
For a matched comparison, we refit the Separable law with a continuous floor, the same five parameters, and the same constraints.
The bounded formulation therefore preserves the monotone two-axis structure while guaranteeing predictions in $[0,1]$.

\paragraph{Configuration selection.}
To test whether boundedness changes practical allocation, we refit both laws when leaving out one backbone size, one input size, or both together; all $1{,}798$ full-grid and held-out optimizations converge.
For each fold we evaluate direct selection at four FLOP budgets, retain decisions with at least two feasible held-out configurations, and define regret as the measured error above the best feasible configuration.
The two laws agree on $100.0\%$ of backbone holdouts, $99.3\%$ of input-size holdouts, and $98.3\%$ of the $2{,}379$ joint-holdout decisions, with nearly identical mean regret.
Thus, enforcing bounded predictions leaves the allocation decisions essentially unchanged.

\begin{table}[htbp]
\appTableStyle
\centering
\small

\caption{{{Configuration selection is stable under bounded outputs.}
Mean regret is reported in percentage points (pp), and agreement is the share of decisions for which the two laws select the same configuration.
Each decision includes at least two feasible candidates.}}
\label{tab:bounded-allocation}
\renewcommand{\arraystretch}{1.12}
\setlength{\tabcolsep}{7pt}
\begin{tabular}{@{}lrrrr@{}}
\toprule
\rowcolor{appHeader}
\appHeadFont Held-out design & \appHeadFont Eligible decisions & \multicolumn{2}{c}{\appHeadFont \textbf{Mean regret (pp)}} & \appHeadFont Same choice (\%) \\
\cmidrule(lr){3-4}
\rowcolor{appHeader}
& & \appHeadFont Separable & \appHeadFont Bounded & \\
\midrule
Backbone holdout & 297 & 0.48 & 0.48 & \cellcolor{appHighlight}\appEmph{100.0} \\
\rowcolor{appStripe}
Input-size holdout & 411 & 0.66 & 0.67 & \cellcolor{appHighlight}\appEmph{99.3} \\
Joint holdout & 2,379 & 0.76 & 0.75 & \cellcolor{appHighlight}\appEmph{98.3} \\
\rowcolor{appStripe}
Joint holdout, two-axis tradeoff & 1,284 & 0.88 & 0.87 & \cellcolor{appHighlight}\appEmph{97.1} \\
\bottomrule
\end{tabular}
\end{table}

\paragraph{Exponent stability.}
To test whether boundedness changes the qualitative scaling relation, we fit both laws on the complete grid and compare $\alpha$ and $\beta$ across all $20$ groups.
Their ordering is identical in every group: $\beta>\alpha$ in all $12$ InternVL groups and in one of the $8$ QwenVL groups, while the rank correlations are $0.82$ for $\alpha$ and $0.96$ for $\beta$.
Although the numerical values differ across error scales, the family-level ordering of the two resource axes is unchanged.

\begin{table}[htbp]
\appTableStyle
\centering
\small

\caption{Median exponents and numbers of groups at the search bounds.}
\label{tab:bounded-exponents}
\renewcommand{\arraystretch}{1.12}
\setlength{\tabcolsep}{8pt}
\begin{tabular}{@{}llrrrrr@{}}
\toprule
\rowcolor{appHeader}
\appHeadFont Family & \appHeadFont Law & \appHeadFont Median $\alpha$ & \appHeadFont Median $\beta$ & \appHeadFont $\alpha=0.05$ & \appHeadFont $\beta=0.05$ & \appHeadFont $\beta=10$ \\
\midrule
InternVL & Separable & 0.196 & 10.000 & 0 & 0 & 9 \\
\rowcolor{appStripe}
InternVL & Bounded & \cellcolor{appHighlight}\textbf{0.113} & \cellcolor{appHighlight}\textbf{10.000} & 5 & 0 & 9 \\
\midrule
QwenVL & Separable & 0.779 & 0.174 & 1 & 0 & 0 \\
\rowcolor{appStripe}
QwenVL & Bounded & \cellcolor{appHighlight}\textbf{0.664} & \cellcolor{appHighlight}\textbf{0.074} & 2 & 3 & 0 \\
\bottomrule
\end{tabular}
\end{table}

\paragraph{Prediction stability.}
To test whether boundedness changes predictive accuracy, we compare the laws on folds that leave out interior levels and folds that leave out endpoints.
Across the three designs, their interior RMSEs differ by at most $0.12$ percentage points, while the bounded law keeps every prediction in $[0,1]$.
The bounded formulation therefore preserves interior predictive accuracy while guaranteeing valid predictions at the edges of the grid.

\begin{table}[htbp]
\appTableStyle
\centering
\small

\caption{{\textbf{Predictive stability under bounded outputs.}
Held-out RMSE is reported in percentage points (pp), together with the number of predictions outside $[0,1]$.
Separable predictions are clipped only when computing RMSE; neither law clips predictions during configuration selection.
Highlighted cells mark the lower RMSE or the bounded law's removal of invalid predictions.}}
\label{tab:bounded-prediction}
\renewcommand{\arraystretch}{1.12}
\setlength{\tabcolsep}{6pt}
\begin{tabular}{@{}llrrrr@{}}
\toprule
\rowcolor{appHeader}
& & \multicolumn{2}{c}{\appHeadFont \textbf{Held-out RMSE (pp)}} & \multicolumn{2}{c}{\appHeadFont \textbf{Predictions outside $[0,1]$}} \\
\cmidrule(lr){3-4}
\cmidrule(lr){5-6}
\rowcolor{appHeader}
\appHeadFont Design & \appHeadFont Held-out level & \appHeadFont Separable & \appHeadFont Bounded & \appHeadFont Separable & \appHeadFont Bounded \\
\midrule
Backbone size & Interior & 4.47 & \cellcolor{appHighlight}\appEmph{4.43} & 0 & 0 \\
\rowcolor{appStripe}
Backbone size & Endpoint & 7.38 & \cellcolor{appHighlight}\appEmph{7.18} & 6 & \cellcolor{appHighlight}\appEmph{0} \\
\midrule
Input size & Interior & 4.56 & \cellcolor{appHighlight}\appEmph{4.45} & 0 & 0 \\
\rowcolor{appStripe}
Input size & Endpoint & \cellcolor{appHighlight}\appEmph{8.29} & 11.79 & 0 & 0 \\
\midrule
Joint holdout & Interior & 4.59 & \cellcolor{appHighlight}\appEmph{4.54} & 0 & 0 \\
\rowcolor{appStripe}
Joint holdout & Endpoint & \cellcolor{appHighlight}\appEmph{6.97} & 8.26 & 36 & \cellcolor{appHighlight}\appEmph{0} \\
\bottomrule
\end{tabular}
\end{table}

\subsection{Parameter Uncertainty and Search Box}
\label{app:val-boxsens}

\paragraph{Parameter uncertainty.}
To measure how precisely the scaling parameters are determined, we refit all $20$ Separable groups in $500$ paired image bootstraps using the original floor grid and exponent range $[0.05,10]$.
The refits reproduce every original-sample SSE to within $1.1\times10^{-11}$, and Table~\ref{tab:parameter-bootstrap} reports the $95\%$ intervals together with the rate at which $\beta$ reaches a search bound.
QwenVL keeps compact $\beta$ intervals below $0.51$ and rarely reaches a bound, whereas the steep InternVL fits frequently reach the upper limit.
The bootstrap therefore makes the frontend contrast explicit. QwenVL has a well-resolved visual-token exponent, while InternVL is consistently much steeper.

\begin{table}[htbp]
\appTableStyle
\centering
\footnotesize
\setlength{\tabcolsep}{4pt}
\renewcommand{\arraystretch}{1.08}

\caption{{\textbf{Bootstrap uncertainty of the Separable-law parameters.}
Entries report $95\%$ percentile intervals from $500$ paired image bootstraps and the share of refits in which $\beta$ reaches either search bound.
Intervals are conditional on the original exponent box $[0.05,10]$; rates of at least $90\%$ are highlighted.}}
\label{tab:parameter-bootstrap}
\begin{tabular}{@{}llcccr@{}}
\toprule
\rowcolor{appHeader}
\appHeadFont Series & \appHeadFont Benchmark & \multicolumn{3}{c}{\appHeadFont \textbf{$95\%$ bootstrap interval}} & \appHeadFont \makecell[r]{$\beta$ at search\\bound (\%)} \\
\cmidrule(lr){3-5}
\rowcolor{appHeader}
& & \appHeadFont $\alpha$ & \appHeadFont $\beta$ & \appHeadFont $e_\infty$ & \\
\midrule
\multirow{4}{*}{InternVL2.5} & HR-Bench & $[0.106, 0.332]$ & $[1.654, 10.000]$ & $[0.001, 0.140]$ & 3.2 \\
& MME-RealWorld & $[0.050, 0.068]$ & $[10.000, 10.000]$ & $[0.001, 0.020]$ & \cellcolor{appHighlight}\appEmph{100.0} \\
& TreeBench & $[0.050, 0.077]$ & $[6.695, 10.000]$ & $[0.001, 0.140]$ & \cellcolor{appHighlight}\appEmph{95.4} \\
& V\textsuperscript{*}\,Bench & $[0.201, 0.651]$ & $[10.000, 10.000]$ & $[0.001, 0.030]$ & \cellcolor{appHighlight}\appEmph{99.0} \\
\midrule
\multirow{4}{*}{InternVL3} & HR-Bench & $[0.135, 0.601]$ & $[1.310, 9.660]$ & $[0.001, 0.200]$ & 1.8 \\
& MME-RealWorld & $[0.050, 0.530]$ & $[10.000, 10.000]$ & $[0.001, 0.440]$ & \cellcolor{appHighlight}\appEmph{100.0} \\
& TreeBench & $[0.050, 0.588]$ & $[1.597, 10.000]$ & $[0.001, 0.520]$ & 88.8 \\
& V\textsuperscript{*}\,Bench & $[0.379, 10.000]$ & $[8.041, 10.000]$ & $[0.001, 0.140]$ & \cellcolor{appHighlight}\appEmph{90.6} \\
\midrule
\multirow{4}{*}{InternVL3.5} & HR-Bench & $[0.112, 0.585]$ & $[1.219, 10.000]$ & $[0.001, 0.260]$ & 8.0 \\
& MME-RealWorld & $[0.753, 1.861]$ & $[10.000, 10.000]$ & $[0.440, 0.480]$ & \cellcolor{appHighlight}\appEmph{100.0} \\
& TreeBench & $[0.050, 2.398]$ & $[0.622, 10.000]$ & $[0.001, 0.560]$ & 61.2 \\
& V\textsuperscript{*}\,Bench & $[0.189, 0.905]$ & $[10.000, 10.000]$ & $[0.001, 0.040]$ & \cellcolor{appHighlight}\appEmph{97.8} \\
\midrule
\multirow{4}{*}{Qwen2.5-VL} & HR-Bench & $[0.247, 10.000]$ & $[0.099, 0.152]$ & $[0.001, 0.001]$ & 0.0 \\
& MME-RealWorld & $[0.100, 10.000]$ & $[0.070, 0.083]$ & $[0.001, 0.060]$ & 0.0 \\
& TreeBench & $[0.050, 10.000]$ & $[0.050, 0.454]$ & $[0.001, 0.540]$ & 6.0 \\
& V\textsuperscript{*}\,Bench & $[0.100, 10.000]$ & $[0.169, 0.257]$ & $[0.001, 0.001]$ & 0.0 \\
\midrule
\multirow{4}{*}{Qwen3-VL} & HR-Bench & $[0.406, 4.459]$ & $[0.128, 0.201]$ & $[0.001, 0.001]$ & 0.0 \\
& MME-RealWorld & $[0.051, 10.000]$ & $[0.091, 0.228]$ & $[0.001, 0.340]$ & 0.0 \\
& TreeBench & $[0.056, 10.000]$ & $[0.050, 0.501]$ & $[0.001, 0.520]$ & 2.6 \\
& V\textsuperscript{*}\,Bench & $[0.343, 1.317]$ & $[0.257, 0.417]$ & $[0.001, 0.001]$ & 0.0 \\
\bottomrule
\end{tabular}
\end{table}

\paragraph{Search-box sensitivity.}
To test whether the frontend contrast depends on the exponent search range, we refit all $20$ groups with $\alpha,\beta\in[0.05,3]$.
All $12$ InternVL groups move to the upper limit $\beta=3$, whereas all $8$ QwenVL groups remain below $0.5$ under both ranges; the narrower box raises pooled SSE from $1.088$ to $1.400$.
Four InternVL group-level floors move by more than $0.02$, indicating some sensitivity to the narrower box.
The search-box experiment nevertheless preserves the family-level exponent contrast while showing that the exact large-$\beta$ and group-level floor estimates depend on the search range.

\begin{table}[htbp]
\appTableStyle
\centering
\small
\setlength{\tabcolsep}{7pt}
\caption{Fits under the narrower exponent box $[0.05,3]$. $^\dagger$ and $^\ddagger$ mark upper-bound $\beta$ and $\alpha$; pooled SSE is $1.400$.}
\label{tab:per-group-fit-narrow}
\begin{tabular}{@{}llrrrr@{}}
\toprule
\rowcolor{appHeader}
\appHeadFont Series & \appHeadFont Benchmark & \appHeadFont $\alpha$ & \appHeadFont $\beta$ & \appHeadFont $e_\infty$ & \appHeadFont $R^2$ \\
\midrule
InternVL2.5 & HR-Bench & 0.17 & \appEmph{$3.00^\dagger$} & 0.001 & 0.91 \\
\rowcolor{appStripe}
InternVL2.5 & MME-RealWorld & 0.07 & \appEmph{$3.00^\dagger$} & 0.001 & 0.69 \\
InternVL2.5 & TreeBench & 0.06 & \appEmph{$3.00^\dagger$} & 0.001 & 0.72 \\
\rowcolor{appStripe}
InternVL2.5 & V\textsuperscript{*}\,Bench & 0.49 & \appEmph{$3.00^\dagger$} & 0.001 & 0.44 \\
InternVL3 & HR-Bench & 0.21 & \appEmph{$3.00^\dagger$} & 0.001 & 0.94 \\
\rowcolor{appStripe}
InternVL3 & MME-RealWorld & 0.22 & \appEmph{$3.00^\dagger$} & 0.300 & 0.65 \\
InternVL3 & TreeBench & 0.07 & \appEmph{$3.00^\dagger$} & 0.001 & 0.87 \\
\rowcolor{appStripe}
InternVL3 & V\textsuperscript{*}\,Bench & 1.58 & \appEmph{$3.00^\dagger$} & 0.001 & 0.39 \\
InternVL3.5 & HR-Bench & 0.17 & \appEmph{$3.00^\dagger$} & 0.001 & 0.92 \\
\rowcolor{appStripe}
InternVL3.5 & MME-RealWorld & 1.20 & \appEmph{$3.00^\dagger$} & 0.380 & 0.69 \\
InternVL3.5 & TreeBench & 0.45 & \appEmph{$3.00^\dagger$} & 0.420 & 0.82 \\
\rowcolor{appStripe}
InternVL3.5 & V\textsuperscript{*}\,Bench & 0.62 & \appEmph{$3.00^\dagger$} & 0.001 & 0.34 \\
\midrule
Qwen2.5-VL & HR-Bench & 0.46 & 0.12 & 0.001 & 0.93 \\
\rowcolor{appStripe}
Qwen2.5-VL & MME-RealWorld & 3.00$^\ddagger$ & 0.08 & 0.001 & 0.93 \\
Qwen2.5-VL & TreeBench & 0.05 & 0.19 & 0.240 & 0.81 \\
\rowcolor{appStripe}
Qwen2.5-VL & V\textsuperscript{*}\,Bench & 3.00$^\ddagger$ & 0.21 & 0.001 & 0.91 \\
Qwen3-VL & HR-Bench & 0.76 & 0.16 & 0.001 & 0.94 \\
\rowcolor{appStripe}
Qwen3-VL & MME-RealWorld & 3.00$^\ddagger$ & 0.11 & 0.060 & 0.96 \\
Qwen3-VL & TreeBench & 0.80 & 0.18 & 0.400 & 0.95 \\
\rowcolor{appStripe}
Qwen3-VL & V\textsuperscript{*}\,Bench & 0.59 & 0.32 & 0.001 & 0.96 \\
\bottomrule
\end{tabular}
\end{table}

\subsection{Sensitivity to the Largest Input Size and to Visual-Token Aggregation}
\label{app:val-without-hr4096}

\paragraph{Removing the HR-Bench 4096 px level.}
To test whether the release-specific $4096$\,px setting drives the conclusions, we remove its $26$ cells and refit the remaining $624$ with the original normalization and constraints.
The $\alpha$--$\beta$ ordering and allocation-path direction remain unchanged in all $20$ groups, pooled RMSE stays nearly constant at $4.102$ percentage points versus $4.092$ on the full grid, and the reduced-grid fits reproduce $79$ of $80$ decisions on the common candidates.
The exponent and allocation conclusions therefore do not depend on the additional HR-Bench input size.

\paragraph{Question-level token aggregation.}
To test whether the cell-mean approximation drives the results, we replace $B(\overline{\rlin})^{-\beta}$ with $B\,\overline{\rlin^{-\beta}}$, computed from the exact visual-token histogram of the scored question records while keeping the same $650$ cells, errors, weights, and constraints.
The alternative preserves the $\alpha$--$\beta$ ordering in all $20$ groups and $79$ of $80$ configuration choices, lowers pooled RMSE from $4.092$ to $3.077$ percentage points, and removes all upper-bound $\beta$ solutions.
Exact question-level aggregation therefore strengthens the fit while preserving the qualitative scaling and allocation conclusions.

\begin{table}[htbp]
\appTableStyle
\centering
\small
\caption{Effect of visual-token aggregation on fit quality and exponent ordering.}
\label{tab:question-token-aggregation}
\begin{tabular}{@{}lrrr@{}}
\toprule
\rowcolor{appHeader}
\appHeadFont Parameterization & \appHeadFont Pooled RMSE (pp) & \appHeadFont $\beta$ at upper bound & \appHeadFont $\alpha>\beta$ \\
\midrule
Cell mean & $4.092$ & $9$ & $7$ \\
\rowcolor{appHighlight}
\appEmph{Question moment} & \appEmph{$3.077$} & \appEmph{$0$} & $7$ \\
\bottomrule
\end{tabular}
\end{table}

\section{Finding 1: The Cognitive Ceiling}
\label{app:ceiling}

{This section reports the task-level fits by series (Appendix~\ref{app:per-family-einf}), tests the regime assignments under alternative cutoffs, held-out configurations, and high-compute evaluation (Appendices~\ref{app:regime-cutoffs} and~\ref{app:ceiling-response-robustness}), and examines domain--skill effects and their stability (Appendices~\ref{app:taxonomy} and~\ref{app:ceiling-skill-fits}).}

\subsection{Task-Level Floors}
\label{app:per-family-einf}

\paragraph{{Reasoning has a higher fitted floor in every series.}}
{To test whether the pooled difference of \S\ref{sec:einf-task} holds across architectures, we break the 40 task-level fits down by model series and task type in Table~\ref{tab:einf-per-family}.
The mean floor is higher for reasoning than for perception in all five series, with pooled means of $0.155$ and $0.073$.
The consistent direction across series shows that the higher reasoning floor is not driven by a single model family.}

\begin{table}[htbp]
\appTableStyle
\centering
\caption{{Mean fitted $\einf$ by model series and by the perception and reasoning labels, pooling the fits of \S\ref{sec:einf-task}. The gap between reasoning and perception is positive in all five series. Gaps are computed from the unrounded means and rounded on their own.}}
\label{tab:einf-per-family}
\footnotesize
\renewcommand{\arraystretch}{1.12}
\setlength{\tabcolsep}{6pt}
\begin{tabular}{lccc}
\toprule
\rowcolor{appHeader}
\appHeadFont \textbf{Series}
& \appHeadFont \textbf{Perception}
& \appHeadFont \textbf{Reasoning}
& \appHeadFont \textbf{Gap} \\
\midrule
InternVL2.5 & 0.001 & 0.067 & $+0.066$ \\
\rowcolor{appStripe}
InternVL3   & 0.097 & 0.187 & $+0.090$ \\
InternVL3.5 & 0.180 & 0.194 & $+0.013$ \\
\rowcolor{appStripe}
Qwen2.5-VL  & 0.001 & 0.187 & $+0.186$ \\
Qwen3-VL    & 0.085 & 0.141 & $+0.056$ \\
\midrule
\rowcolor{appHighlight}
\appEmph{Pooled} & \appEmph{0.073} & \appEmph{0.155} & \appEmph{$+0.082$} \\
\bottomrule
\end{tabular}
\end{table}

\paragraph{{High floors are concentrated in a few pairings of benchmark and task.}}
{To locate the task subsets behind the pooled result, we inspect the fits that meet the Saturation threshold, $\einf \ge 0.20$.
Ten of the 40 fits saturate, namely five of 25 perception fits and five of 15 reasoning fits, corresponding to $20\%$ and $33\%$ in Table~\ref{tab:einf-task}. Membership is determined after rounding to two decimals because two estimates lie numerically at the threshold.
Six cases cluster in TreeBench reasoning and MME-RealWorld, while the remaining four occur in individual HR-Bench or TreeBench subsets.
By contrast, $18$ perception fits and $8$ reasoning fits reach the Floor criterion, $\einf \le 0.01$.
Thus, most task-level fits approach a near-zero floor, while the high floors are localized rather than universal.}

\paragraph{{The comparison uses matched task-specific fits.}}
{To construct the comparison, we fit every combination of model series, benchmark, and task subset, yielding the 40 estimates in Table~\ref{tab:per-task-fits}.
HR-Bench is divided into Cross and Single, MME-RealWorld and TreeBench into Perception and Reasoning, and V\textsuperscript{*}\,Bench into Direct Attributes and Relative Position.
For MME-RealWorld, we retain the native subtasks and weight every pair of subtask and configuration equally. Each InternVL series therefore contributes $180$ points to its perception fit and $144$ to its reasoning fit.
This construction preserves the benchmark-specific task structure while supporting a consistent perception--reasoning comparison.}

\begin{table}[htbp]
\appTableStyle
\centering
\footnotesize
\renewcommand{\arraystretch}{0.98}
\caption{{\textbf{Task-level Separable-law exponents.}
The table covers all $40$ series--benchmark--task fits; $^\dagger$ and $^\ddagger$ mark the upper and lower exponent bounds, $10$ and $0.05$.
Interior values are point estimates and need not be statistically identified.}}
\label{tab:per-task-fits}
\setlength{\tabcolsep}{5pt}
\begin{tabular}{lllrr}
\toprule
\rowcolor{appHeader}
\appHeadFont Series & \appHeadFont Benchmark & \appHeadFont Task & \multicolumn{2}{c}{\appHeadFont \textbf{Exponent}} \\
\cmidrule(lr){4-5}
\rowcolor{appHeader}
& & & \appHeadFont $\alpha$ & \appHeadFont $\beta$ \\
\midrule
\multirow{8}{*}{InternVL2.5}
& \multirow{2}{*}{HR-Bench} & Cross & 0.099 & 5.18 \\
& & Single & 0.221 & 9.50 \\
& \multirow{2}{*}{MME-RealWorld} & Perception & 0.068 & \appEmph{$10.0^{\dagger}$} \\
& & Reasoning & 0.065 & 7.05 \\
& \multirow{2}{*}{TreeBench} & Perception & 0.115 & \appEmph{$10.0^{\dagger}$} \\
& & Reasoning & \appEmph{$0.05^{\ddagger}$} & 5.11 \\
& \multirow{2}{*}{{V\textsuperscript{*}\,Bench}} & Direct Attributes & 0.645 & \appEmph{$10.0^{\dagger}$} \\
& & Relative Position & 0.234 & \appEmph{$10.0^{\dagger}$} \\
\midrule
\multirow{8}{*}{InternVL3}
& \multirow{2}{*}{HR-Bench} & Cross & 0.136 & 3.28 \\
& & Single & 0.240 & 9.15 \\
& \multirow{2}{*}{MME-RealWorld} & Perception & 0.465 & 9.08 \\
& & Reasoning & 0.069 & 5.48 \\
& \multirow{2}{*}{TreeBench} & Perception & 0.134 & \appEmph{$10.0^{\dagger}$} \\
& & Reasoning & 0.152 & \appEmph{$10.0^{\dagger}$} \\
& \multirow{2}{*}{{V\textsuperscript{*}\,Bench}} & Direct Attributes & 2.90 & \appEmph{$10.0^{\dagger}$} \\
& & Relative Position & 0.470 & \appEmph{$10.0^{\dagger}$} \\
\midrule
\multirow{8}{*}{InternVL3.5}
& \multirow{2}{*}{HR-Bench} & Cross & 0.129 & 3.27 \\
& & Single & 0.193 & \appEmph{$10.0^{\dagger}$} \\
& \multirow{2}{*}{MME-RealWorld} & Perception & 0.999 & 6.36 \\
& & Reasoning & 0.569 & 4.57 \\
& \multirow{2}{*}{TreeBench} & Perception & 0.701 & \appEmph{$10.0^{\dagger}$} \\
& & Reasoning & 0.213 & 0.180 \\
& \multirow{2}{*}{{V\textsuperscript{*}\,Bench}} & Direct Attributes & 0.491 & \appEmph{$10.0^{\dagger}$} \\
& & Relative Position & 0.396 & \appEmph{$10.0^{\dagger}$} \\
\midrule
\multirow{8}{*}{Qwen2.5-VL}
& \multirow{2}{*}{HR-Bench} & Cross & 0.051 & 0.289 \\
& & Single & 1.10 & 0.175 \\
& \multirow{2}{*}{MME-RealWorld} & Perception & \appEmph{$10.0^{\dagger}$} & 0.119 \\
& & Reasoning & \appEmph{$10.0^{\dagger}$} & 0.088 \\
& \multirow{2}{*}{TreeBench} & Perception & 0.074 & 0.247 \\
& & Reasoning & 1.23 & 0.095 \\
& \multirow{2}{*}{{V\textsuperscript{*}\,Bench}} & Direct Attributes$^{a}$ & 0.116 & 0.230 \\
& & Relative Position & \appEmph{$10.0^{\dagger}$} & 0.152 \\
\midrule
\multirow{8}{*}{Qwen3-VL}
& \multirow{2}{*}{HR-Bench} & Cross & 0.502 & 0.197 \\
& & Single & 1.09 & 0.222 \\
& \multirow{2}{*}{MME-RealWorld} & Perception & 2.13 & 0.161 \\
& & Reasoning & 4.34 & 0.118 \\
& \multirow{2}{*}{TreeBench} & Perception & 0.681 & 0.281 \\
& & Reasoning & 0.840 & 0.053 \\
& \multirow{2}{*}{{V\textsuperscript{*}\,Bench}} & Direct Attributes & 0.835 & 0.338 \\
& & Relative Position & 0.367 & 0.307 \\
\bottomrule
\end{tabular}
\end{table}

\subsection{Sensitivity of the Regime Cutoffs}
\label{app:regime-cutoffs}

\paragraph{{The low-accuracy stratum persists across cutoff choices.}}
{To test whether the three regimes of \S\ref{sec:regime} depend on their thresholds, we sweep $\tau_{\text{easy}}$ over $\{0.85,0.90,0.95\}$ and $\tau_{\text{cog}}$ over $\{0.15,0.20,0.25\}$ for all $19{,}890$ trajectories.
Table~\ref{tab:regime-cutoffs} shows that the \textsc{Ceiling Bound} share remains between $25.2\%$ and $33.2\%$, with the default cutoffs giving $30.7\%$.
Thus, the substantial low-accuracy stratum is not an artifact of one cutoff choice.}

\begin{table}[htbp]
\appTableStyle
\centering
\caption{{Share of trajectories in each regime (\%) under the cutoff sweep, written as \textsc{Easy} / \textsc{Scaling Bound} / \textsc{Ceiling Bound}. The default cutoffs, $\tau_{\text{easy}}=0.90$ and $\tau_{\text{cog}}=0.20$, are in bold.}}
\label{tab:regime-cutoffs}
\footnotesize
\renewcommand{\arraystretch}{1.12}
\setlength{\tabcolsep}{6pt}
\begin{tabular}{lccc}
\toprule
\rowcolor{appHeader}
& \appHeadFont $\tau_{\text{easy}}=0.85$
& \appHeadFont $\tau_{\text{easy}}=0.90$
& \appHeadFont $\tau_{\text{easy}}=0.95$ \\
\midrule
$\tau_{\text{cog}}=0.15$ & 22.3 / 52.5 / 25.2 & 20.0 / 54.8 / 25.2 & 17.4 / 57.4 / 25.2 \\
\rowcolor{appStripe}
$\tau_{\text{cog}}=0.20$ & 22.3 / 47.0 / 30.7 & \appEmph{20.0 / 49.4 / 30.7} & 17.4 / 51.9 / 30.7 \\
$\tau_{\text{cog}}=0.25$ & 22.3 / 44.4 / 33.2 & 20.0 / 46.8 / 33.2 & 17.4 / 49.3 / 33.2 \\
\bottomrule
\end{tabular}
\end{table}

\paragraph{{The regimes generalize to held-out configurations.}}
{To separate regime definition from evaluation, we assign each trajectory using half of its configurations and measure accuracy and scaling response on the other half.
Across all five series, held-out accuracy is $7.08\%$--$8.85\%$ for \textsc{Ceiling Bound} and $50.53\%$--$54.79\%$ for \textsc{Scaling Bound}, while their slopes are respectively $0.13$--$1.24$ and $3.13$--$4.55$ percentage points per doubling of $\nlin\rlin$ (Table~\ref{tab:regime-split-sample}).
Among questions shared by all series, $17.95\%$ are \textsc{Ceiling Bound} in every series.
The regime labels therefore capture reproducible differences in both accuracy and scaling response.}

\begin{table}[htbp]
\appTableStyle
\centering
\small
\caption{{Held-out performance after defining the regimes on the other half of the configurations. Accuracy is in percent; slopes are percentage points per doubling of $\nlin\rlin$ on the held-out half. Light-blue cells mark the held-out \textsc{Ceiling Bound} results.}}
\label{tab:regime-split-sample}
\setlength{\tabcolsep}{4pt}
\begin{tabular}{lrrrrrr}
\toprule
\rowcolor{appHeader}
& \multicolumn{2}{c}{\appHeadFont \textsc{Ceiling Bound}}
& \multicolumn{2}{c}{\appHeadFont \textsc{Scaling Bound}}
& \multicolumn{2}{c}{\appHeadFont \textsc{Easy}} \\
\cmidrule(lr){2-3}\cmidrule(lr){4-5}\cmidrule(lr){6-7}
\rowcolor{appHeader}
\appHeadFont Series
& \appHeadFont Accuracy & \appHeadFont Slope
& \appHeadFont Accuracy & \appHeadFont Slope
& \appHeadFont Accuracy & \appHeadFont Slope \\
\midrule
InternVL2.5 & \appEmph{$7.08$} & \appEmph{$0.30$} & $53.40$ & $4.55$ & $96.56$ & $1.01$ \\
\rowcolor{appStripe}
InternVL3   & \appEmph{$7.62$} & \appEmph{$0.13$} & $54.79$ & $4.36$ & $96.48$ & $0.80$ \\
InternVL3.5 & \appEmph{$8.21$} & \appEmph{$0.60$} & $52.82$ & $3.82$ & $95.35$ & $1.47$ \\
\rowcolor{appStripe}
Qwen2.5-VL  & \appEmph{$7.50$} & \appEmph{$0.58$} & $50.93$ & $3.13$ & $93.28$ & $1.22$ \\
Qwen3-VL    & \appEmph{$8.85$} & \appEmph{$1.24$} & $50.53$ & $4.49$ & $94.74$ & $0.62$ \\
\bottomrule
\end{tabular}
\end{table}

\subsection{Persistence at High Compute}
\label{app:ceiling-response-robustness}

\paragraph{{\textsc{Ceiling Bound} errors persist at the highest observed compute.}}
{To test whether \textsc{Ceiling Bound} questions remain difficult at high compute, we retain trajectories with sufficient configurations, backbone sizes, valid probabilities, and consistent option sets, yielding $15{,}830$ trajectories from $3{,}166$ questions evaluated by all five series.
Of these, $5{,}392$, or $34.1\%$, belong to $\mathcal{C}_{\rm mean}$, the \textsc{Ceiling Bound} stratum in this cohort.
Within each trajectory, we order configurations by $u_{ij}=\nlin_jR^{\rm q}_{ij}$ and form tied compute quartiles $Q_1$ to $Q_4$.
To compare questions with two to eleven options, we verify the prompts and labels, exclude 12 inconsistent questions, and normalize correctness $a_{ij}$ and correct-option probability $p_{ij}$ relative to the uniform-guessing rate $c_i=1/K_i$:}
\begin{equation}
c_i=\frac{1}{K_i},\qquad
z^*_{ij}=\frac{z_{ij}-c_i}{1-c_i},\qquad z\in\{a,p\},
\label{eq:ceiling-response-chance}
\end{equation}
{keeping negative values unchanged.
In the highest-compute quartile, $4{,}602$ of the $5{,}392$ trajectories, or $85.3\%$, still have accuracy below $0.20$; their mean accuracy is $8.6\%$, and only $6.6\%$ reach at least $0.50$.
Meanwhile, $32.1\%$ gain more than $0.05$ in correct-option probability from $Q_1$ to $Q_4$, while $27.8\%$ lose more than $0.05$.
The matched within-trajectory experiment therefore confirms that low accuracy persists at the highest observed compute even though individual scaling responses vary.}

\paragraph{{A stricter response-based core lies inside \textsc{Ceiling Bound}.}}
{To distinguish low performance at high compute from weak response to scaling, for $z\in\{a,p,a^*,p^*\}$ we write $z_q$ for the mean over quartile $q$, $\Delta z=z_4-z_1$, $\Delta_{\rm late}z=z_4-z_3$, and $b_z$ for the slope against $\log_2u$ over all configurations, and compare}
\begin{equation}
\begin{aligned}
H_z &: z_4<\tau,\\
E_z &: H_z\land |\Delta z|\le\delta,\qquad
P_z : H_z\land |\Delta_{\rm late}z|\le\delta,\\
B_z &: H_z\land |b_z|\le\eta,\qquad
J_z : E_z\cap P_z\cap B_z.
\end{aligned}
\label{eq:ceiling-response-criteria}
\end{equation}
{$H_z$ requires low performance at high compute, while $J_z$ additionally requires small endpoint change, small late change, and a small fitted slope.
The default margins are $\tau=0.20$ for the raw measures, $\tau=0.10$ for the normalized ones, $\delta=0.05$, and $\eta=0.01$ per doubling of compute.
Applying the joint rule selects $1{,}502$ trajectories, or $9.5\%$, using raw probability and $1{,}296$, or $8.2\%$, after chance normalization (Table~\ref{tab:ceiling-response-definitions}).
Of these, $98.9\%$ and $97.1\%$, respectively, lie inside $\mathcal{C}_{\rm mean}$.
The response-based experiment therefore isolates a smaller persistent-error core while confirming that \textsc{Ceiling Bound} captures nearly all of it.}

\begin{table}[htbp]
\appTableStyle
\centering
\caption{{Share of the same $15{,}830$ trajectories selected by each criterion (\%). Every row requires low $Q_4$ performance. The raw columns use $\tau=0.20$ and the normalized columns $\tau=0.10$, with $\delta=0.05$ and $\eta=0.01$ per doubling. The \textsc{Ceiling Bound} rule selects $34.1\%$ of this cohort, and the $30.7\%$ of \S\ref{sec:regime} uses the full cohort instead.}}
\label{tab:ceiling-response-definitions}
\footnotesize
\renewcommand{\arraystretch}{1.12}
\setlength{\tabcolsep}{5pt}
\begin{tabular}{@{}lrrrr@{}}
\toprule
\rowcolor{appHeader}
\appHeadFont \textbf{Criterion} & \appHeadFont $a$ & \appHeadFont $p$ & \appHeadFont $a^*$ & \appHeadFont $p^*$ \\
\midrule
{Low $Q_4$ performance} $H_z$ & 37.5 & 30.6 & 39.1 & 40.0 \\
\rowcolor{appStripe}
Small endpoint change $E_z$ & 20.8 & 13.1 & 20.8 & 13.0 \\
Small late change $P_z$ & 23.5 & 18.3 & 23.4 & 18.1 \\
\rowcolor{appStripe}
Small fitted slope $B_z$ & 19.0 & 14.1 & 18.3 & 13.9 \\
\rowcolor{appHighlight}
\appEmph{All three together} $J_z$ & \appEmph{16.4} & \appEmph{9.5} & \appEmph{16.3} & \appEmph{8.2} \\
\bottomrule
\end{tabular}
\end{table}

\paragraph{{Persistent-error rates are highest on MME-RealWorld, TreeBench, and reasoning tasks.}}
{To test whether the group-level pattern depends on the definition, Table~\ref{tab:ceiling-response-groups} compares \textsc{Ceiling Bound} with the two joint response criteria on the same observations.
Under both criteria, MME-RealWorld and TreeBench keep higher rates than HR-Bench and V\textsuperscript{*}\,Bench, and reasoning keeps a higher rate than perception, $11.1\%$ against $6.7\%$ after chance normalization.
The broad benchmark and task contrasts therefore persist even when the definition of persistent error becomes stricter.}

\begin{table}[htbp]
\appTableStyle
\centering
\caption{{Benchmarks and tasks compared on the same observations (\%). The benchmark rows and the task rows are two separate partitions of the same $15{,}830$ trajectories. HR-Bench covers only the six shared input sizes, since the trajectories that exist only at $4096$\,px are excluded. The \textsc{Ceiling Bound} column uses mean accuracy below $0.20$, and $J_p$ and $J_{p^*}$ follow Eq.\eqref{eq:ceiling-response-criteria}.}}
\label{tab:ceiling-response-groups}
\footnotesize
\renewcommand{\arraystretch}{1.12}
\setlength{\tabcolsep}{6pt}
\begin{tabular}{@{}lrccc@{}}
\toprule
\rowcolor{appHeader}
\appHeadFont \textbf{Group} & \appHeadFont \textbf{Trajectories} & \appHeadFont {\textsc{Ceiling Bound}} & \appHeadFont $J_p$ & \appHeadFont $J_{p^*}$ \\
\midrule
{HR-Bench} & 4{,}000 & 23.0 & 4.5 & 3.9 \\
\rowcolor{appStripe}
MME-RealWorld & 8{,}900 & 39.3 & 12.4 & 10.8 \\
TreeBench & 1{,}975 & 40.8 & 9.7 & 8.3 \\
\rowcolor{appStripe}
V\textsuperscript{*}\,Bench & 955 & 17.6 & 2.4 & 1.6 \\
\midrule
Perception & 10{,}525 & 30.0 & 7.9 & 6.7 \\
\rowcolor{appStripe}
Reasoning & 5{,}305 & 42.1 & 12.7 & 11.1 \\
\bottomrule
\end{tabular}
\end{table}

\paragraph{{A small-response core remains across thresholds and token aggregation.}}
{To test sensitivity to the response definition, we sweep the raw and normalized performance cutoffs, the change margin, and the slope margin.
The joint rule then selects between $4.2\%$ and $18.4\%$ of trajectories on raw probability and between $3.5\%$ and $15.5\%$ after normalization, so a group with small response remains throughout the sweep while its size depends on the margins.
Using the mean visual token count of each configuration instead of the count logged for each question gives $9.7\%$ and $8.4\%$, close to the defaults of $9.5\%$ and $8.2\%$.
Requiring the four quartile means to span at most $0.05$ still leaves $970$ candidates, or $6.1\%$.
Thus, a nontrivial small-response core remains across reasonable thresholds and visual-token aggregation choices.}

\paragraph{{High-compute difficulty replicates across configuration halves.}}
{To check internal stability, we split the configurations in every quartile into two outcome-independent halves using a fixed hash ordering with seed $20260909$.
Selecting $J_{p^*}$ on one half gives $1{,}268$ candidates, of which $824$, or $65.0\%$, also satisfy the rule on the other half, and reversing the halves gives $1{,}180$ candidates with $69.8\%$ replication.
Difficulty at high compute itself carries over to the other half for $98.2\%$ and $96.8\%$ of the selected candidates.
The main high-compute difficulty signal is therefore highly stable across disjoint configurations, and together the experiments establish persistent error over the evaluated grid without extrapolating to unseen architectures or to configurations beyond our grid.}

\subsection{Taxonomy of Domain and Skill}
\label{app:taxonomy}

\paragraph{{Why we relabel.}}
{The four benchmarks label their questions in ways that do not match.
HR-Bench separates perception of a single instance from perception across instances, MME-RealWorld gives subtasks and categories, TreeBench names subtypes of reasoning, and V\textsuperscript{*}\,Bench separates direct attributes from relative position.
These labels work within a benchmark but give no shared vocabulary across benchmarks.
To locate the cognitive ceiling, we therefore give every question two labels: the domain $D$ is the visual scene of the image, and the skill $S$ is the operation the question requires.
Keeping the two labels separate is what lets us ask whether the floor $\einf$ follows the domain, the skill, or a particular domain and skill pairing.
Table~\ref{tab:taxonomy} gives the $6\times6$ grid, with domains D1 Documents/Charts/Slides, D2 Aerial/Satellite, D3 Vehicles/Driving, D4 Indoor, D5 Outdoor, and D6 People/Surveillance, and skills S1 Attribute Recognition, S2 Text Reading, S3 Counting, S4 Spatial Relation, S5 Object Identification, and S6 Scene Reasoning.}

\paragraph{{Labeling and coverage.}}
{GPT-4o-mini assigns the labels at temperature $0$, reading the image at low detail together with the full question text and its answer options, and returning exactly one domain and one skill.
We join the labels back to the evaluation records by benchmark and question ID.
The labeling run covered the $800$ HR-Bench questions evaluated at the six shared input sizes.
The records at $4096$\,px come from the other HR-Bench release and carry no labels, so the fits below use the six shared input sizes.
This gives one pair of labels for each of the $3{,}178$ unique questions, namely $800$ from HR-Bench, $1{,}782$ from MME-RealWorld, $405$ from TreeBench, and $191$ from V\textsuperscript{*}\,Bench.}

\paragraph{{Checking the labels.}}
{We inspect the assigned labels by hand against the taxonomy and the decision rules.
As an automatic check, we compare them with a fixed mapping of the native metadata of MME-RealWorld.
The MME-RealWorld metadata does not correspond one to one with our taxonomy, so the agreement rate measures how consistent the two label sets are rather than how accurate ours is.}

\paragraph{{Labeling prompt.}}
{The box below abridges the system prompt.
The full prompt also gives ordered rules for choosing a domain, rules for skills that overlap, and instructions on the output format.
Afterwards we only parse the returned labels and map them to their canonical names, and we never change an assignment.}
\begin{quote}\small
You are labeling an image for a vision-language scaling-law study.
Look at the image and read the question text, then output TWO labels.

DOMAIN -- pick exactly one: D1 Documents/Charts/Slides, D2 Aerial/Satellite, D3 Vehicles/Driving, D4 Indoor, D5 Outdoor-General, or D6 People/Surveillance.

SKILL -- pick exactly one: S1 Attribute Recognition, S2 Text Reading, S3 Counting, S4 Spatial Relation, S5 Identification, or S6 Reasoning over Scene.

Output a single JSON line, no other text:
\verb|{"domain": "D?", "skill": "S?"}|.
\end{quote}
{The prompt writes D5 as Outdoor-General, S5 as Identification, and S6 as Reasoning over Scene, and the paper writes them as Outdoor, Object Identification, and Scene Reasoning.
Labels are stored as the codes D1 to D6 and S1 to S6, so the wording changes nothing in the analysis.}

\subsection{Fits for Each Domain--Skill Pairing}
\label{app:ceiling-skill-fits}

\paragraph{{The 33 supported pairings provide a common basis for comparing fitted floors.}}
{To compare persistent error across domains and skills, we fit Eq.\eqref{eq:sep-law} separately to every supported domain--skill pairing at the six shared input sizes.
Each fit pools the five model series and all contributing benchmarks, uses the same error definition and parameterization as the main analysis, and weights eligible benchmark--configuration--domain--skill subcells equally.
Retaining subcells with at least five questions yields $10{,}135$ subcells across $624$ configurations and $33$ supported pairings.
This common protocol makes the fitted floors directly comparable across the domain--skill grid.}

\paragraph{{High floors are concentrated in specific domain--skill pairings.}}
{To locate the strongest ceiling effects, we compare the 33 fitted floors and their domain and skill marginals.
Table~\ref{tab:DS-ceiling-pockets} lists ten pairings with a fitted floor of at least $0.30$, spanning several visual domains.
The highest is Aerial/Satellite with Scene Reasoning, at $\einf=0.780$.
Table~\ref{tab:DS-marginals} averages the floors over the supported domain--skill pairings to give one marginal mean for each skill and one for each domain.
The skill means run from $0.001$ for Text Reading to $0.432$ for Attribute Recognition, and the domain means from $0.041$ for Documents/Charts/Slides to $0.272$ for People/Surveillance.
Text Reading and Object Identification remain near the lower floor across almost every supported domain.
The largest ceilings are therefore localized to particular combinations of scene and required skill rather than entire benchmarks or domains.}

\begin{table}[htbp]
\appTableStyle
\centering
\caption{Marginal fitted floors by skill and domain. Each mean equally weights the supported domain--skill pairings; Pairings gives their number.}
\label{tab:DS-marginals}
\footnotesize
\renewcommand{\arraystretch}{1.12}
\setlength{\tabcolsep}{4pt}
\begin{tabular}{@{}lrr@{\hskip 2em}lrr@{}}
\toprule
\rowcolor{appHeader}
\multicolumn{3}{c}{\appHeadFont Skill marginals}
& \multicolumn{3}{c}{\appHeadFont Domain marginals} \\
\cmidrule(lr){1-3}\cmidrule(lr){4-6}
\rowcolor{appHeader}
\appHeadFont Skill & \appHeadFont Mean $\einf$ & \appHeadFont Pairings
& \appHeadFont Domain & \appHeadFont Mean $\einf$ & \appHeadFont Pairings \\
\midrule
S1 Attribute Recognition & \appEmph{0.432} & 5 & D1 Documents/Charts/Slides & \appEmph{0.041} & 4 \\
\rowcolor{appStripe}
S2 Text Reading & \appEmph{0.001} & 6 & D2 Aerial/Satellite & 0.227 & 6 \\
S3 Counting & 0.250 & 6 & D3 Vehicles/Driving & 0.204 & 6 \\
\rowcolor{appStripe}
S4 Spatial Relation & 0.213 & 5 & D4 Indoor & 0.124 & 6 \\
S5 Object Identification & 0.044 & 6 & D5 Outdoor & 0.214 & 6 \\
\rowcolor{appStripe}
S6 Scene Reasoning & 0.228 & 5 & D6 People/Surveillance & \appEmph{0.272} & 5 \\
\bottomrule
\end{tabular}
\end{table}

\paragraph{{Skill accounts for more of the spread than domain.}}
{To separate the two sources of variation, we fit least-squares models to the 33 floors using domain alone, skill alone, and both factors together.
All three weight the 33 domain--skill pairings equally and leave out the three unsupported pairings, rather than filling them in.
They explain $9.26\%$, $38.09\%$, and $44.51\%$ of the variance of the floors.
Either factor can enter the model with both terms first or second, so we average its share over the two orders,}
\begin{equation}
V_S=\tfrac12\bigl[R_S^2+(R_{D+S}^2-R_D^2)\bigr],
\qquad
V_D=\tfrac12\bigl[R_D^2+(R_{D+S}^2-R_S^2)\bigr].
\label{eq:domain-skill-variance-decomposition}
\end{equation}
{This assigns $36.67\%$ of the variance to skill and $7.84\%$ to domain, a ratio of $4.68$.
Skill therefore captures substantially more of the broad variation in fitted floors, while specific pairings account for the remaining structure.}

\begin{table}[htbp]
\appTableStyle
\centering
\caption{{The ten domain--skill pairings with a fitted floor of at least $0.30$, under the taxonomy of Table~\ref{tab:taxonomy} and the protocol above.}}
\label{tab:DS-ceiling-pockets}
\footnotesize
\renewcommand{\arraystretch}{1.12}
\setlength{\tabcolsep}{4pt}
\begin{tabular}{llc}
\toprule
\rowcolor{appHeader}
\appHeadFont \textbf{Pairing} & \appHeadFont \textbf{Domain $\times$ Skill} & \appHeadFont \textbf{$\einf$} \\
\midrule
\rowcolor{appHighlight}
\appEmph{D2$\times$S6} & \appEmph{Aerial/Satellite $\times$ Scene Reasoning} & \appEmph{0.780} \\
\rowcolor{appStripe}
D6$\times$S4 & People/Surveillance $\times$ Spatial Relation & 0.580 \\
D3$\times$S3 & Vehicles/Driving $\times$ Counting & 0.540 \\
\rowcolor{appStripe}
D5$\times$S1 & Outdoor $\times$ Attribute Recognition & 0.540 \\
D2$\times$S1 & Aerial/Satellite $\times$ Attribute Recognition & 0.520 \\
\rowcolor{appStripe}
D5$\times$S4 & Outdoor $\times$ Spatial Relation & 0.480 \\
D6$\times$S3 & People/Surveillance $\times$ Counting & 0.480 \\
\rowcolor{appStripe}
D3$\times$S1 & Vehicles/Driving $\times$ Attribute Recognition & 0.400 \\
D4$\times$S1 & Indoor $\times$ Attribute Recognition & 0.400 \\
\rowcolor{appStripe}
D6$\times$S1 & People/Surveillance $\times$ Attribute Recognition & 0.300 \\
\bottomrule
\end{tabular}
\end{table}

\paragraph{{Cluster resampling preserves the domain--skill comparison.}}
{To assess whether the ordering between $V_S$ and $V_D$ depends on particular images, we repeat the full analysis under two image-cluster resampling schemes.
Both schemes retain the original fitting choices, namely the $55$-point floor grid, $\alpha,\beta\in[0.05,10]$, nonnegative amplitudes, equal subcell weights, and the original scoring rule.
Within each benchmark, a cluster contains one image together with all of its questions, models, and input sizes.
Each replicate reweights or resamples these clusters, recomputes the subcell errors and measured visual token counts, refits every domain--skill floor that remains defined, and repeats the equal-weight two-factor decomposition.
The ordinary bootstrap yields $397$ complete-support replicates out of $500$, while a positive-weight bootstrap preserves all 33 pairings in all $500$ replicates.
The skill share exceeds the domain share in $82.6\%$ of ordinary-bootstrap replicates and $81.6\%$ of positive-weight replicates, with closely aligned percentile ranges (Table~\ref{tab:DS-bootstrap-decomposition}).
Although the ratio intervals cross one and do not pin down the point-estimate ratio of $4.68$, both schemes support the same skill-dominant ordering.
Cluster resampling therefore preserves the main domain--skill comparison under complementary checks with and without support loss in rare pairings.}

\begin{table}[htbp]
\appTableStyle
\centering
\footnotesize
\caption{Image-cluster bootstrap stability of the domain--skill variance decomposition. Intervals are the $2.5$th--$97.5$th percentiles; ordinary-bootstrap intervals use the 397 replicates retaining all 33 pairings.}
\label{tab:DS-bootstrap-decomposition}
\setlength{\tabcolsep}{3.5pt}
\begin{tabular}{@{}lccccc@{}}
\toprule
\rowcolor{appHeader}
\multirow{2}{*}{\appHeadFont Resampling}
& \multirow{2}{*}{\appHeadFont Complete}
& \multicolumn{3}{c}{\appHeadFont $95\%$ percentile interval}
& \multirow{2}{*}{\makecell{\appHeadFont $V_S>V_D$\\\appHeadFont (\% replicates)}} \\
\cmidrule(lr){3-5}
& & \appHeadFont $V_S$ (\%) & \appHeadFont $V_D$ (\%) & \appHeadFont $V_S/V_D$ & \\
\midrule
Ordinary cluster & $397/500$ & $10.60$--$40.89$ & $6.25$--$29.07$ & $0.60$--$4.27$ & \appEmph{$82.6$} \\
\rowcolor{appStripe}
Positive-weight cluster & $500/500$ & $11.94$--$39.13$ & $6.28$--$28.78$ & $0.57$--$4.65$ & \appEmph{$81.6$} \\
\bottomrule
\end{tabular}
\end{table}

\paragraph{{Most domain--skill fits have interior exponents.}}
{To check whether the pairing-level fits are driven by the exponent search limits, we inspect the optimizer solutions across all 33 pairings.
$\beta$ lies away from the bounds in 29 fits, and both exponents are interior in 28.
D2$\times$S1 has $\alpha$ at the upper bound of $10$, and D2$\times$S3, D2$\times$S6, D3$\times$S6, and D5$\times$S3 have $\beta$ at the lower bound of $0.05$.
For D2$\times$S4 and D2$\times$S6 the fitted amplitude $A_n$ is zero, so the matching $\alpha$ has no effect on the fitted curve at any observed configuration and is not determined, even when the optimizer returns a value away from the bounds.}
{Thus, most pairing-level fits are not pinned to the exponent limits, although interior estimates alone do not establish statistical identification.}

\section{Finding 2: Architectural Divergence}
\label{app:arch}

{We first report the two fitted exponents by family and series and compare their ratio on the task-level fits (Appendix~\ref{app:arch-alpha-beta}).
Because $\beta$ reaches the upper bound in most InternVL groups, we then measure how far the visual token count of each frontend actually moves, test effective pixels as an alternative visual measure, and vary source-image quality at fixed token sequences (Appendix~\ref{app:arch-token-ranges}).}

\subsection{Exponents by Family, Series, and Task}
\label{app:arch-alpha-beta}

\paragraph{{The backbone capacity exponent is interior in all InternVL groups and most QwenVL groups.}}
{To compare backbone capacity scaling across the two families, we inspect $\alpha$ in all 20 series--benchmark fits and summarize the estimates that lie inside the search bounds.
All $12$ InternVL fits are interior, with a median of $0.19$, while $5$ of the $8$ QwenVL fits are interior, with a median of $0.59$; including the three boundary estimates raises the QwenVL median to $0.78$.
The boundary cases are Qwen2.5-VL on MME-RealWorld and V\textsuperscript{*}\,Bench and Qwen3-VL on MME-RealWorld.
In all three, the fitted amplitude at the reference point is $A_n\le4\times10^{-6}$.
A small amplitude there does not make the backbone capacity term negligible across the observed range, since that term is $A_n(\nlin/\widetilde N)^{-\alpha}$ and $\nlin$ falls well below $\widetilde N$ for the smaller models.
Backbone capacity scaling is therefore represented by an interior exponent in most groups, with greater boundary sensitivity in QwenVL.}

\paragraph{{The visual-token exponent sharply separates the two families.}}
{To compare visual-token scaling, we examine $\beta$ under the same group fits.
Nine of the $12$ InternVL fits reach the upper bound of $10$, and the three that do not are all on HR-Bench, at $3.87$, $4.66$, and $4.93$.
All $8$ QwenVL fits stay away from the bounds, with a median of $0.17$ ($0.173$ before rounding).
Although individual interior estimates can still be imprecise, the group fits consistently distinguish the steep InternVL response from the gradual QwenVL response.}

\paragraph{{The $\beta$ contrast repeats in every model series.}}
{To test whether the family difference is driven by one release, Table~\ref{tab:arch-alpha-beta-series} summarizes the four benchmark fits of each series.
Every InternVL series has exactly one estimate away from the bounds, always its HR-Bench fit, and that value lies between $3.87$ and $4.93$ in all three series.
Both QwenVL series have all four estimates away from the bounds, with medians of $0.155$ and $0.173$.
The backbone capacity exponent varies more across releases, but the visual-token exponent preserves the same family ordering in every series.
The architectural divergence is therefore repeated across releases rather than produced by a single model generation.}

\begin{table}[htbp]
\appTableStyle
\centering
\footnotesize
\caption{Group-level exponents by series, with four benchmark fits each. Interior means strictly inside $[0.05,10]$. For $\beta$, InternVL reports its sole interior HR-Bench estimate and QwenVL reports the median of four interior estimates.}
\label{tab:arch-alpha-beta-series}
\setlength{\tabcolsep}{6pt}
\begin{tabular}{@{}lcccc@{}}
\toprule
\rowcolor{appHeader}
\multirow{2}{*}{\appHeadFont Series}
& \multicolumn{3}{c}{\appHeadFont Backbone capacity exponent $\alpha$}
& \multirow{2}{*}{\appHeadFont Interior $\beta$} \\
\cmidrule(lr){2-4}
\rowcolor{appHeader}
& \appHeadFont Median & \appHeadFont Range & \appHeadFont At bound & \\
\midrule
\multicolumn{5}{l}{\appHeadFont InternVL} \\
InternVL2.5 & 0.10 & {$0.057$--$0.381$} & 0 & \appEmph{4.66} \\
\rowcolor{appStripe}
InternVL3   & 0.19 & {$0.071$--$1.262$} & 0 & \appEmph{3.87} \\
InternVL3.5 & 0.44 & {$0.145$--$1.203$} & 0 & \appEmph{4.93} \\
\midrule
\multicolumn{5}{l}{\appHeadFont QwenVL} \\
\rowcolor{appStripe}
Qwen2.5-VL & 0.26 & {$0.053$--$0.462$} & 2 & \appEmph{0.155} \\
Qwen3-VL   & 0.76 & {$0.589$--$0.800$} & 1 & \appEmph{0.173} \\
\bottomrule
\end{tabular}
\end{table}

\paragraph{{The exponent ratio reverses between QwenVL and InternVL.}}
\label{app:arch-ab-ratio}
{To compare the relative response to backbone capacity and visual tokens, we compute $\alpha/\beta$ on the 40 task-level fits.
Both exponents lie away from the bounds in 13 of the 16 QwenVL fits and in 11 of the 24 InternVL fits, and 12 of the 13 InternVL estimates at a bound sit at the upper bound on $\beta$.
Among the fits with both exponents away from the bounds, the median ratio is $2.55$ for QwenVL, with $\alpha>\beta$ in 10 of 13, and $0.0395$ for InternVL, with $\beta>\alpha$ in 10 of 11.
The one InternVL exception is InternVL3.5 on TreeBench Reasoning.
One QwenVL fit has $A_n=0$, so its $\alpha$ is not determined; dropping it leaves 12 fits with a median ratio of $3.72$.
On reasoning tasks alone, the QwenVL and InternVL medians are $14.47$ and $0.0685$.
The task-level ratios therefore reinforce the same family-specific balance between the two scaling axes.}

\subsection{Token Ranges of the Two Frontends}
\label{app:arch-token-ranges}

\paragraph{{InternVL tiling keeps the visual-token range narrow.}}
{To determine how strongly input size changes the resource seen by the language backbone, we compare InternVL3.5-8B and Qwen3-VL-8B on $3{,}166$ equally weighted questions for which both models record valid token counts at all six shared input sizes and receive identical input pixels.
InternVL selects at most 12 tiles by matching the input aspect ratio, uses pixel count only to break ties, and converts each $448\times448$ tile into $256$ visual tokens after pixel unshuffle~\citep{chen2024far}.
We find that $2{,}223$ questions keep the same tile count at the two endpoints and $2{,}214$, or $69.9\%$, keep it at all six sizes.
Across all input sizes within each group, including $4096$\,px on HR-Bench, the maximum token count is only $1.07$ to $1.55$ times the minimum, with a median span of $1.22$.
Consistent with this limited intervention range, $9$ of the $12$ InternVL fits place $\beta$ at its upper bound, while the three HR-Bench estimates are $3.87$, $4.66$, and $4.93$.
The tiling frontend therefore exposes a consistently narrow visual-token range over the evaluated input-size grid, as visualized in Figure~\ref{fig:sec7-combined}a.}

\paragraph{{QwenVL continuous resizing exposes a broad visual-token range.}}
{To compare the two frontends on the same resource measure, we measure how QwenVL token counts change across the evaluated input sizes.
QwenVL resizes the whole image and compresses each $2\times2$ block of patches into one visual token, so we compute the mean token count in every cell and its span within each group.
The count changes at the much finer granularity of the patch grid, although it need not rise at every step. For Qwen3-VL 2B on HR-Bench, the mean decreases from $74.875$ at $224$\,px to $73.480$ at $336$\,px.
Across the full grids, the maximum count is $34$ to $360$ times the minimum and the cells cover roughly $37$ to $15{,}000$ tokens; over the six shared input sizes alone, the spans remain $34.09$ to $87.79$.
All $8$ QwenVL fits consequently place $\beta$ away from the bounds, with a median of $0.17$.
Continuous resizing therefore supplies broad, fine-grained token variation from which a gradual visual-token response can be estimated, as visualized in Figure~\ref{fig:sec7-combined}a.}

\begin{figure}[t]
    \centering
    \includegraphics[width=\linewidth]{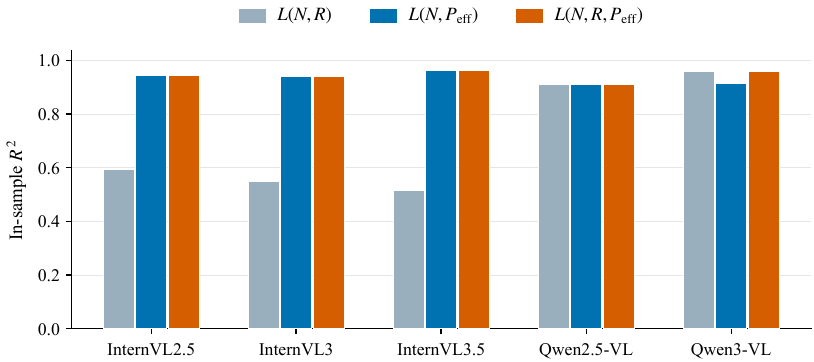}
    \caption{{\textbf{Effective pixels as an alternative visual scaling measure on V\textsuperscript{*}\,Bench.}
    In-sample $R^2$ for fits using $(\nlin,\rlin)$, $(\nlin,P_{\mathrm{eff}})$, and $(\nlin,\rlin,P_{\mathrm{eff}})$ under the same constraints.
    Effective pixels substantially improve the InternVL fits but not the Qwen3-VL fit.
    For InternVL, adding $\rlin$ after $P_{\mathrm{eff}}$ yields almost no further increase in $R^2$; together with boundary estimates or zero amplitudes, this means that the three-variable fits do not statistically separate their contributions.}}
    \label{fig:effective-resolution}
\end{figure}

\paragraph{{Effective pixels explain InternVL variation that token count alone misses.}}
\label{app:arch-effective-resolution}
{To test whether token count alone captures the image detail preserved by each frontend, we refit V\textsuperscript{*}\,Bench after replacing $L(\nlin,\rlin)$ with $L(\nlin,P_{\mathrm{eff}})$.
For every question, we define $P_{\mathrm{eff}}=\min\!\left(P_{\mathrm{input}},P_{\mathrm{grid}}\right)$, where $P_{\mathrm{input}}$ is the resized input area and $P_{\mathrm{grid}}$ is the spatial area represented by the frontend grid; for InternVL, the repeated thumbnail is excluded.
We compute this quantity before cell aggregation, retain the original fitting constraints, and compare in-sample fit, BIC, and whole-level prediction under $\rlin$, $P_{\mathrm{eff}}$, and both coordinates together.
For InternVL2.5, InternVL3, and InternVL3.5, replacing $\rlin$ by $P_{\mathrm{eff}}$ raises $R^2$ from $0.594$, $0.548$, and $0.515$ to $0.943$, $0.940$, and $0.963$, respectively, and none of the effective-pixel exponents reaches a search bound.
For InternVL3.5, holding out one backbone size at a time reduces RMSE from $8.95$ to $3.32$ percentage points, while holding out one input-size level at a time reduces it from $10.50$ to $3.16$ points.
BIC likewise favors $P_{\mathrm{eff}}$ for all three InternVL series, whereas Qwen2.5-VL is essentially unchanged and Qwen3-VL decreases from $R^2=0.959$ to $0.914$.
Adding $\rlin$ after $P_{\mathrm{eff}}$ produces almost no further increase in $R^2$ for InternVL and introduces boundary estimates or zero amplitudes, so the current grid cannot separately identify their conditional effects.
Figure~\ref{fig:effective-resolution} summarizes these comparisons.
The experiment therefore identifies effective pixels as a substantially more informative and complementary visual measure for InternVL, especially when token counts are fixed or change only slightly.}

\begin{figure}[t]
    \centering
    \includegraphics[width=\linewidth]{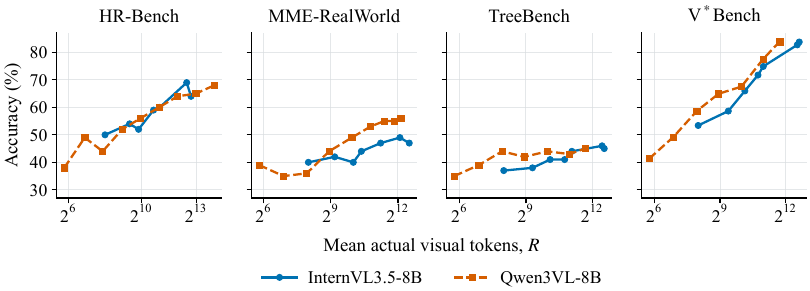}
    \caption{{\textbf{Visual-token sweeps on native source images.}
    Accuracy against the mean actual visual-token count for InternVL3.5-8B and Qwen3-VL-8B on the four benchmarks.
    The two frontends expose different token ranges, so differences between their endpoint gains do not by themselves isolate an architectural effect.}}
    \label{fig:native-token-sweeps}
\end{figure}

\paragraph{{Raising the visual token count on native images improves accuracy on several benchmarks.}}
\label{app:quality-budget-factorial}
{To test whether visual tokens are a genuine intervention rather than only a fitted quantity, we hold model size fixed at 8B, retain the native source images, and vary only the number of visual tokens.
For InternVL3.5-8B, we sweep \texttt{max\_num} over $\{1,2,4,6,12,24,36\}$. For Qwen3-VL-8B, pixel caps set token counts from $64$ to $16{,}384$.
We compare the largest and smallest token counts with paired accuracy differences, pointwise $95\%$ percentile intervals from $20{,}000$ image bootstraps, and Holm correction over the $16$ prespecified primary tests.
Table~\ref{tab:native-token-sweeps} reports the endpoint tests, while Figure~\ref{fig:native-token-sweeps} shows the full accuracy curves against the measured token count.
The endpoint gain is significant for InternVL on V\textsuperscript{*}\,Bench at $30.37$ points, and for QwenVL on HR-Bench, MME-RealWorld, and V\textsuperscript{*}\,Bench at $30.00$, $17.00$, and $42.41$ points.
These controlled sweeps confirm that raising the visual token count can produce substantial accuracy gains for both frontends.}

\begin{table}[htbp]
\appTableStyle
\centering
\small
\caption{{Endpoint gains in the native-source visual-token sweeps. Gains and intervals are in percentage points. Light-blue rows mark gains with Holm-adjusted $p<0.05$.}}
\label{tab:native-token-sweeps}
\setlength{\tabcolsep}{5pt}
\begin{tabular}{llrrrr}
\toprule
\rowcolor{appHeader}
\appHeadFont Series & \appHeadFont Benchmark & \appHeadFont Questions
& \appHeadFont Gain & \appHeadFont $95\%$ interval & \appHeadFont Holm $p$ \\
\midrule
InternVL3.5-8B & HR-Bench & $100$ & $14.00$ & $[4.00,24.00]$ & $0.0692$ \\
\rowcolor{appStripe}
InternVL3.5-8B & MME-RealWorld & $100$ & $7.00$ & $[-1.00,15.00]$ & $0.225$ \\
InternVL3.5-8B & TreeBench & $100$ & $8.00$ & $[1.00,16.00]$ & $0.225$ \\
\rowcolor{appHighlight}
InternVL3.5-8B & V\textsuperscript{*}\,Bench & $191$ & \appEmph{$30.37$} & $[23.56,37.17]$ & \appEmph{$2.62{\times}10^{-14}$} \\
\rowcolor{appHighlight}
Qwen3-VL-8B & HR-Bench & $100$ & \appEmph{$30.00$} & $[19.00,41.00]$ & \appEmph{$3.21{\times}10^{-6}$} \\
\rowcolor{appHighlight}
Qwen3-VL-8B & MME-RealWorld & $100$ & \appEmph{$17.00$} & $[8.00,26.00]$ & \appEmph{$0.00554$} \\
Qwen3-VL-8B & TreeBench & $100$ & $10.00$ & $[0.00,20.00]$ & $0.225$ \\
\rowcolor{appHighlight}
Qwen3-VL-8B & V\textsuperscript{*}\,Bench & $191$ & \appEmph{$42.41$} & $[34.55,50.26]$ & \appEmph{$4.15{\times}10^{-20}$} \\
\bottomrule
\end{tabular}
\end{table}

\begin{figure}[t]
    \centering
    \includegraphics[width=\linewidth]{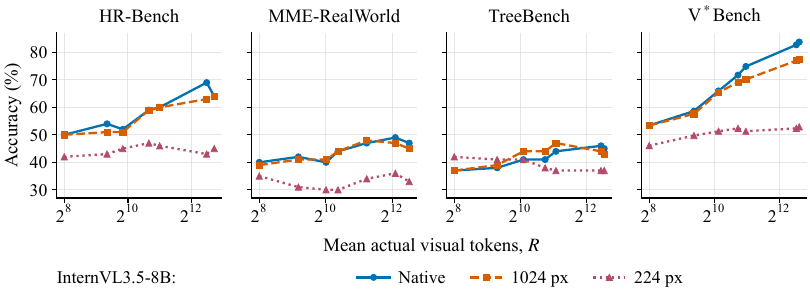}
    \caption{{\textbf{Source-image information quality modulates the return to visual tokens.}
    InternVL3.5-8B accuracy under native, $1024$-pixel, and $224$-pixel source-quality conditions crossed with seven tile limits.
    Actual visual-token counts and input-sequence hashes match across quality conditions for every combination of question and cap, so the curves vary pixel information while holding the input sequence structure fixed.}}
    \label{fig:quality-budget-factorial}
\end{figure}

\paragraph{{Source-image quality amplifies the return to more visual tokens.}}
{To separate source information from input-sequence length, we cross the seven InternVL3.5-8B tile limits with native images and versions whose long side is reduced to $1024$ or $224$ pixels and then restored to the native canvas.
Across all $3{,}437$ combinations of question and cap, the three quality conditions have identical actual token counts and identical input-token-sequence hashes, so the factorial changes available pixel information while preserving the model input structure.
On V\textsuperscript{*}\,Bench, the native condition gains $30.37$ points between the smallest and largest token counts, from $53.40\%$ to $83.77\%$, whereas the $224$-pixel condition gains only $6.81$ points, from $46.07\%$ to $52.88\%$.
Source quality and token count therefore interact, by $23.56$ points ($95\%$ interval $[14.66,32.46]$, Holm-adjusted $p=4.32{\times}10^{-6}$), and at the largest token count the native condition leads the $224$-pixel condition by $30.89$ points ($[24.08,38.22]$, Holm-adjusted $p=4.01{\times}10^{-14}$).
The corresponding quality effects at the largest token count are also significant at $19$ points on HR-Bench and $14$ on MME-RealWorld, while the $8$-point TreeBench effect is not.
Figure~\ref{fig:quality-budget-factorial} shows the complete factorial curves for all four benchmarks.
The factorial therefore confirms that high-quality source information enables models to make substantially better use of additional visual tokens.}

\paragraph{{Controlled sweeps support a frontend-dependent, grid-specific visual response.}}
{To identify what the fitted frontend contrast represents, we combine the effective-pixel comparison in Figure~\ref{fig:effective-resolution}, the native-source intervention in Table~\ref{tab:native-token-sweeps} and Figure~\ref{fig:native-token-sweeps}, and the source-quality factorial in Figure~\ref{fig:quality-budget-factorial}.
The sweeps establish $\rlin$ as a genuine resource axis, while the factorial shows that its return depends on the usable image information summarized by $P_{\mathrm{eff}}$.
Together, these results provide stronger evidence for a frontend-dependent visual response than the fitted $\beta$ values alone.
They do not identify an intrinsic architectural exponent reversal, because the two families traverse different token and image-information ranges and the fixed-8B sweeps contain no variation in $\nlin$ from which to recover the ordering of $\alpha$ and $\beta$ across model sizes.
The VPE allocation rates $\beta/(p\beta+q\alpha)$ and $\alpha/(p\beta+q\alpha)$ should therefore be read as conditional on the evaluated grids.
The two frontends therefore induce different fitted resource responses on the grids we evaluate, rather than fixed architecture-wide constants.}

\section{Finding 3: From Laws to Practice}
\label{app:vpe}

{We first define the cost measure and test sensitivity to the FLOPs of the vision encoder and projector (Appendix~\ref{app:cost}), then derive the continuous VPE allocation in a machine-checkable form (Appendix~\ref{app:vpe-lean}).
We next evaluate discrete choices on held-out configurations and in sample by series (Appendices~\ref{app:vpe-heldout-allocation} and~\ref{app:vpe-regret}), before conditioning the benchmark profiles on model family (Appendix~\ref{app:vpe-diag-byfamily}).}

\subsection{Cost Measures and Cost-Law Fits}
\label{app:cost}
\label{app:visual-cost-sensitivity}

\paragraph{{Cost measure.}}
{We measure allocation cost in terms of the FLOPs the language backbone spends on the prompt,}
\begin{equation}
C \;=\; 2\,P_{\mathrm{LLM}}\,T_{\mathrm{prompt}},
\end{equation}
{where $P_{\mathrm{LLM}}$ is the exact parameter count of the language backbone and $T_{\mathrm{prompt}}$ is the logged prompt length, which includes the question text, the visual tokens, and the chat and special tokens.
This is the standard $2NT$ measure, and it gives one accounting basis for both visual frontends.
A variant that adds the short generation phase and the attention FLOPs, $C'=2P_{\mathrm{LLM}}(T_{\mathrm{prompt}}+T_{\mathrm{gen}})+C_{\mathrm{attn}}$, is available for the same $650$ cells and leads to the same conclusions.
The original FLOP logs cover the vision encoder and projector for only $257$ and $150$ of the $650$ cells, under conventions that differ across series, so the main budget leaves them out.
The visual-side sensitivity analysis below reconstructs these FLOPs for all $650$ cells and tests their effect on the fitted allocation paths.
The main allocation remains a language-side budget, and the expanded measure is a sensitivity analysis rather than an end-to-end system cost.
The question-level robustness checks instead order configurations by $u=\nlin R^{\rm q}$, which ranks the configurations of one trajectory and leaves out the question and template tokens that enter the budget here.}

\paragraph{{Fitted cost law.}}
\label{app:arch-cost-law}
{We fit the cost law $C=\kappa \nlin^p\rlin^q$ of \S\ref{sec:vpe-decision} to all 650 cells pooled together and to each series separately, using the FLOPs above and the measured wall-clock latency as two candidate cost measures.
The allocations in this paper use the pooled fit of the FLOPs measure together with the performance fit of each group, and the fits for each series are diagnostic rather than a source of budgets.
Table~\ref{tab:arch-cost-law} shows that the FLOPs measure follows this law in all five series, with $R^2$ in log space from $0.986$ to $1.000$.
The fitted $p$ is $1.00$ in every series, as expected for the size of a transformer backbone.
The fitted $q$ is $0.96$ in all three InternVL series and $0.75$ to $0.78$ in the two QwenVL series.
The prompt also carries the question and the template, so cost grows less than proportionally with $\rlin$ over the wide token range of QwenVL, while the narrow range of InternVL keeps $q$ near one.
Wall-clock latency is not a stable cost axis across the two families.
Its $R^2$ runs from $0.019$ to $0.764$ across series, which reflects scheduler noise, transfers between CPU and GPU, differences in batching, and implementation overheads that $\nlin$ and $\rlin$ do not capture.
We therefore use the FLOPs measure as the common cost axis and treat latency as a property of the implementation rather than of the architecture.}

\begin{table}[htbp]
\appTableStyle
\centering
\small
\caption{{Cost-law fits $C=\kappa \nlin^p \rlin^q$, with $\nlin$ in billions of language backbone parameters and $\rlin$ in visual tokens. The law is fitted to the FLOPs measure and, separately, to wall-clock latency, and $R^2$ is computed in log space. Only the pooled fit of the FLOPs measure sets budgets, and the fits for each series and for latency are diagnostic.}}
\label{tab:arch-cost-law}
\setlength{\tabcolsep}{5pt}
\begin{tabular}{lrrrrrrrr}
\toprule
\rowcolor{appHeader}
& & \multicolumn{4}{c}{\appHeadFont {FLOPs of the language backbone}} & \multicolumn{3}{c}{\appHeadFont Latency} \\
\cmidrule(lr){3-6}\cmidrule(lr){7-9}
\rowcolor{appHeader}
\appHeadFont Scope & \appHeadFont $n$ & \appHeadFont $\kappa$ & \appHeadFont $p$ & \appHeadFont $q$ & \appHeadFont $R^2$ & \appHeadFont $p$ & \appHeadFont $q$ & \appHeadFont $R^2$ \\
\midrule
\rowcolor{appHighlight}
\appEmph{Pooled} & 650 & $1.24{\times}10^{10}$ & \appEmph{1.00} & \appEmph{0.77} & \appEmph{0.995} & 0.07  & 0.54  & 0.366 \\
\rowcolor{appStripe}
InternVL2.5 & 150 & $2.95{\times}10^{9}$  & 1.00 & 0.96 & 1.000 & -0.04 & -0.15 & 0.019 \\
InternVL3   & 150 & $2.96{\times}10^{9}$  & 1.00 & 0.96 & 1.000 & 0.14  & -0.53 & 0.029 \\
\rowcolor{appStripe}
InternVL3.5 & 150 & $2.93{\times}10^{9}$  & 1.00 & 0.96 & 1.000 & 0.24  & 1.04  & 0.150 \\
Qwen2.5-VL  & 100 & $1.38{\times}10^{10}$ & 1.00 & 0.75 & 0.986 & 0.03  & 0.43  & 0.764 \\
\rowcolor{appStripe}
Qwen3-VL    & 100 & $1.14{\times}10^{10}$ & 1.00 & 0.78 & 0.991 & -0.14 & 0.45  & 0.543 \\
\bottomrule
\end{tabular}
\end{table}

\paragraph{{Adding visual-side cost preserves the family-level allocation direction.}}
{To test whether omitting vision computation changes the allocation result, we extend the prompt cost to
\begin{equation}
C_{\lambda}
=C_{\mathrm{LLM,prompt}}
+\lambda\bigl(C_{\mathrm{vision}}+C_{\mathrm{projector}}\bigr),
\qquad \lambda\in\{0,0.5,1,2\},
\end{equation}
over all $650$ configurations, $26$ models, and five series.
For Qwen2.5-VL, the vision cost includes both window and global attention, and for Qwen3-VL it includes global attention and the three DeepStack mergers.
For each question, we use the actual temporal, height, and width of its visual patch grid to determine the tensor shapes, compute the visual cost, and then average within a cell.
For InternVL, we retain the original parameter-times-token estimator, which reproduces the historical component accounting on all $150$ InternVL3 cells.
We then keep the fitted performance surfaces fixed and recompute their finite-candidate allocation paths under each $C_{\lambda}$.
On the same $101$ log-spaced budgets supported by all four values of $\lambda$, every one of the $12$ InternVL groups has a larger growth slope in $\log\nlin$ than in $\log\rlin$, while all eight QwenVL groups show the reverse.
An oracle path constructed directly from the observed errors agrees in $19$ of the $20$ groups, with only Qwen3-VL on TreeBench reversing direction.
At the original four absolute budgets and $\lambda=1$, $10$ of the $80$ decisions become infeasible and $39$ of the $70$ commonly feasible decisions select the same configuration as the language-only cost.
Thus the family-level allocation direction is stable after adding the visual-side cost, while absolute budgets and individual choices must be recalibrated under the new cost definition.
This in-sample check excludes language-model attention and decoding, preprocessing, communication, and measured latency, and therefore does not claim an end-to-end system cost.
Because fixed $\alpha$ and $\beta$ already determine the ordering of the continuous elasticities, the comparison here is made on the finite candidate paths.}

\subsection{Derivation of the VPE Allocation}
\label{app:vpe-lean}

\paragraph{{What this derivation covers.}}
{The fitting procedure estimates the parameters of the Separable Law and of the cost law from data.
Given those two laws and positive coefficients and exponents, the derivation below produces the continuous VPE solution and its budget elasticities.
It therefore says nothing about how well the two laws fit, or about what happens outside the observed grid.}

\paragraph{{From the fitted amplitudes to physical coordinates.}}
{The fits store the amplitudes at the common reference point $(\widetilde N,\widetilde R)$,}
\begin{equation}
  L(\nlin,\rlin)
  =
  e_\infty
  +
  A_n\left(\frac{\nlin}{\widetilde N}\right)^{-\alpha}
  +
  B_r\left(\frac{\rlin}{\widetilde R}\right)^{-\beta},
  \label{eq:lean-normalized-law}
\end{equation}
{while the cost law is written in the resource variables themselves.
Absorbing the reference scales into the amplitudes, $A=A_n\widetilde N^{\alpha}$ and $B=B_r\widetilde R^{\beta}$, leaves every predicted error unchanged and returns Eq.\eqref{eq:sep-law} in physical coordinates.
The derivation below uses $A$ and $B$, and $\nlin$ carries the same unit as in the pooled cost fit, namely billions of parameters.
Changing that unit to $\nlin'=s\nlin$ takes $\widetilde N$ to $s\widetilde N$ and $\kappa$ to $\kappa/s^p$, which leaves the allocation itself unchanged.
This conversion leaves the allocation invariant to the unit chosen for backbone size.}

\paragraph{{The problem.}}
{We take positive resources, a positive budget, and strictly positive coefficients and exponents on both the error side and the cost side,}
\begin{equation}
  A,B,\alpha,\beta,p,q,\kappa,C_0>0.
\end{equation}
{The continuous allocation problem then minimizes error over the positive resource pairs that the budget allows,}
\begin{equation}
  \min_{\nlin,\rlin>0}
  \left[
    e_\infty+A \nlin^{-\alpha}
    +B \rlin^{-\beta}
  \right]
  \quad
  \text{subject to}
  \quad
  \kappa \nlin^p\rlin^q\le C_0.
  \label{eq:lean-vpe-program}
\end{equation}
{Error falls in both $\nlin$ and $\rlin$ while cost rises in both, so any optimum spends the whole budget and the inequality becomes an equality.
The search therefore runs along the boundary of fixed cost.}

{For the closed form, write the rescaled budget and two positive constants,}
\begin{equation}
  \overline C=\frac{C_0}{\kappa},
  \qquad
  D=p\beta+q\alpha,
  \qquad
  K=
  \frac{\alpha A q}
       {\beta B p}.
  \label{eq:lean-vpe-defs}
\end{equation}
{The assumptions give $\overline C>0$, $D>0$, and $K>0$, so the candidate below is well defined and assigns positive resources to both axes,}
\begin{equation}
  \nlin^*
  =
  \overline C^{\beta/D}K^{q/D},
  \qquad
  \rlin^*
  =
  \overline C^{\alpha/D}K^{-p/D}.
  \label{eq:lean-vpe-closed}
\end{equation}

\paragraph{{The closed form and its uniqueness.}}
{In log coordinates $x=\log \nlin$ and $y=\log \rlin$, the budget constraint becomes affine, and the positive $(\nlin,\rlin)$ domain maps to all of $\mathbb{R}^2$.
On the boundary of fixed cost,}
\begin{equation}
  px+qy=\log\overline C.
  \label{eq:lean-log-budget}
\end{equation}
{At a stationary allocation, the error removed per proportional unit of spending is equal on the two axes,}
\begin{equation}
  \frac{\alpha A}{p}\nlin^{-\alpha}
  =
  \frac{\beta B}{q}\rlin^{-\beta},
  \label{eq:lean-log-foc}
\end{equation}
{and taking logarithms turns this balance into a second linear relation,}
\begin{equation}
  \alpha x-\beta y=\log K.
  \label{eq:lean-log-balance}
\end{equation}
{Since $D=p\beta+q\alpha>0$, the system of
Eqs.\eqref{eq:lean-log-budget} and \eqref{eq:lean-log-balance} has the single solution}
\begin{equation}
  x^*
  =
  \frac{\beta}{D}\log\overline C
  +
  \frac{q}{D}\log K,
  \qquad
  y^*
  =
  \frac{\alpha}{D}\log\overline C
  -
  \frac{p}{D}\log K,
  \label{eq:lean-log-solution}
\end{equation}
{and exponentiating it recovers Eq.\eqref{eq:lean-vpe-closed}.
This gives one stationary candidate on the boundary, and global optimality still needs an argument.}

{To supply it, eliminate $y$ through the constraint, $y=(\log\overline C-px)/q$, and write the objective on the boundary as a function of $x$ alone,}
\begin{equation}
  G(x)
  =
  e_\infty
  +
  A e^{-\alpha x}
  +
  B\,
  \overline C^{-\beta/q}
  e^{(\beta p/q)x}.
  \label{eq:lean-one-dimensional-objective}
\end{equation}
{The floor does not depend on the allocation, and differentiating the other two terms twice gives}
\begin{equation}
  G''(x)
  =
  \alpha^2 A e^{-\alpha x}
  +
  \left(\frac{\beta p}{q}\right)^2
  B\,
  \overline C^{-\beta/q}
  e^{(\beta p/q)x}
  >0,
  \label{eq:lean-second-derivative}
\end{equation}
{so $G$ is strictly convex along the whole boundary.
It also diverges at both ends, since $G(x)\to+\infty$ as $x\to-\infty$ and as $x\to+\infty$, so a minimizer exists and strict convexity makes it unique.
Its stationarity condition is Eq.\eqref{eq:lean-log-foc}, so the minimizer is exactly the pair in Eq.\eqref{eq:lean-vpe-closed}.}

\paragraph{{Budget elasticities.}}
{Holding the fitted parameters fixed and taking logarithms of Eq.\eqref{eq:lean-vpe-closed} gives}
\begin{align}
  \log \nlin^*
  &=
  \frac{\beta}{D}
  \bigl(\log C_0-\log\kappa\bigr)
  +
  \frac{q}{D}\log K,
  \\
  \log \rlin^*
  &=
  \frac{\alpha}{D}
  \bigl(\log C_0-\log\kappa\bigr)
  -
  \frac{p}{D}\log K.
\end{align}
{Both are affine in the log budget, and their slopes are the proportional response of each optimal resource to a proportional increase in budget,}
\begin{equation}
  \frac{\partial\log \nlin^*}{\partial\log C_0}
  =
  \frac{\beta}{p\beta+q\alpha},
  \qquad
  \frac{\partial\log \rlin^*}{\partial\log C_0}
  =
  \frac{\alpha}{p\beta+q\alpha}.
  \label{eq:lean-vpe-elasticities}
\end{equation}
{The level of the allocation depends on the coefficients as well, while these two rates depend only on $\alpha$, $\beta$, $p$, and $q$.
They describe continuous growth rather than a choice among available configurations.}

\paragraph{{Lean 4 certifies the algebraic core of the allocation.}}
{To verify the symbolic derivation independently of the fitting code, we formalize its log-coordinate core as a $2\times2$ linear system.
The Lean~4 source below records the positivity assumptions, checks the budget and balance identities, proves that their intersection is unique, and verifies the finite-difference form of the two budget elasticities.
It compiles with Lean~4.34.0 and the matching Mathlib release.}

\begin{leancertificate}
import Mathlib

noncomputable section
open Real

namespace VPE

structure Params where
  A      : Real
  B      : Real
  alpha  : Real
  beta   : Real
  p      : Real
  q      : Real
  kappa  : Real
  C0     : Real
  hA     : 0 < A
  hB     : 0 < B
  ha     : 0 < alpha
  hb     : 0 < beta
  hp     : 0 < p
  hq     : 0 < q
  hk     : 0 < kappa
  hC0    : 0 < C0

def D (t : Params) : Real :=
  t.p * t.beta + t.q * t.alpha

def Cbar (t : Params) : Real :=
  t.C0 / t.kappa

def K (t : Params) : Real :=
  t.alpha * t.A * t.q /
    (t.beta * t.B * t.p)

lemma D_pos (t : Params) : 0 < D t := by
  unfold D
  exact add_pos
    (mul_pos t.hp t.hb)
    (mul_pos t.hq t.ha)

lemma Cbar_pos (t : Params) : 0 < Cbar t := by
  unfold Cbar
  exact div_pos t.hC0 t.hk

lemma K_pos (t : Params) : 0 < K t := by
  unfold K
  exact div_pos
    (mul_pos (mul_pos t.ha t.hA) t.hq)
    (mul_pos (mul_pos t.hb t.hB) t.hp)

def xStar (t : Params) : Real :=
  t.beta / D t * Real.log (Cbar t) +
  t.q / D t * Real.log (K t)

def yStar (t : Params) : Real :=
  t.alpha / D t * Real.log (Cbar t) -
  t.p / D t * Real.log (K t)

theorem log_budget (t : Params) :
    t.p * xStar t + t.q * yStar t =
      Real.log (Cbar t) := by
  unfold xStar yStar
  have hD : D t ≠ 0 := ne_of_gt (D_pos t)
  field_simp [hD]
  unfold D
  ring

theorem log_balance (t : Params) :
    t.alpha * xStar t - t.beta * yStar t =
      Real.log (K t) := by
  unfold xStar yStar
  have hD : D t ≠ 0 := ne_of_gt (D_pos t)
  field_simp [hD]
  unfold D
  ring

theorem unique_log_solution
    (t : Params) {x y : Real}
    (hb :
      t.p * x + t.q * y = Real.log (Cbar t))
    (he :
      t.alpha * x - t.beta * y = Real.log (K t)) :
    x = xStar t /\ y = yStar t := by

  have hD : D t ≠ 0 := ne_of_gt (D_pos t)

  have hx :
      D t * x =
        t.beta * Real.log (Cbar t) +
        t.q * Real.log (K t) := by
    unfold D
    linear_combination t.beta * hb + t.q * he

  have hy :
      D t * y =
        t.alpha * Real.log (Cbar t) -
        t.p * Real.log (K t) := by
    unfold D
    linear_combination t.alpha * hb - t.p * he

  constructor
  · unfold xStar
    field_simp [hD]
    nlinarith [hx]

  · unfold yStar
    field_simp [hD]
    nlinarith [hy]

def logNAtBudget (t : Params) (s : Real) : Real :=
  t.beta / D t * s +
  t.q / D t * Real.log (K t)

def logRAtBudget (t : Params) (s : Real) : Real :=
  t.alpha / D t * s -
  t.p / D t * Real.log (K t)

theorem logN_budget_increment
    (t : Params) (s h : Real) :
    logNAtBudget t (s + h) - logNAtBudget t s =
      (t.beta / D t) * h := by
  unfold logNAtBudget
  ring

theorem logR_budget_increment
    (t : Params) (s h : Real) :
    logRAtBudget t (s + h) - logRAtBudget t s =
      (t.alpha / D t) * h := by
  unfold logRAtBudget
  ring

end VPE
\end{leancertificate}

{The listing certifies the algebraic core in log coordinates, namely that the closed form satisfies the budget equation and the balance equation, that their intersection is unique, and that the two log-budget responses have the stated slopes.
The convexity and the limits that give global optimality come from the argument above rather than from the listing.
The values of $A_n$, $B_r$, $\alpha$, $\beta$, $p$, $q$, and $\kappa$ remain outputs of the fitting pipeline, which the listing says nothing about.
A continuous optimum also does not name the best configuration in a finite set, so discrete selection must be evaluated separately.
The certificate therefore provides an independent, machine-checked validation of the allocation algebra while keeping its empirical and discrete-selection claims clearly separated.}

\subsection{Held-Out Allocation}
\label{app:vpe-heldout-allocation}

\paragraph{{Joint holdouts test whether the fitted law can allocate unseen configurations.}}
{This is the evaluation behind \S\ref{sec:vpe-validation}.
For each of the 20 groups we jointly hold out one backbone size and one input size, removing the full row and column they span, and refit the performance law using only the remaining cells.
The resulting $650$ folds provide withheld candidate configurations, which we evaluate at budgets equal to the $25$th, $50$th, $75$th, and $90$th percentiles of each group's language-side cell costs.
Regret is the measured error of the selected configuration minus the lowest measured error among the feasible candidates, in percentage points.
After excluding $44$ decisions with no feasible candidate and $177$ with only one, the evaluation retains $2{,}379$ of the $2{,}600$ planned decisions.
We compare two rules that use the fitted law with three static baselines.
Projection clips the continuous optimum of Eq.\eqref{eq:lean-vpe-closed} to the observed coordinate ranges and selects the nearest feasible candidate in log coordinates, whereas direct selection evaluates Eq.\eqref{eq:sep-law} over every feasible candidate and takes the lowest predicted error.
The static rules select the largest feasible backbone at the smallest visual token count, the largest feasible visual token count at the smallest backbone, or the candidate nearest the log-center of the feasible ranges; ties are broken by cell order without measured errors.
Because the closed form degenerates when either fitted amplitude approaches zero, projection falls back to direct selection when $A_n<10^{-8}$ or $B_r<10^{-8}$, which occurs in $133$ decisions.
The same folds and budgets also support the comparison between the Separable and bounded laws.}

\paragraph{{The fitted law substantially reduces held-out allocation regret.}}
{Table~\ref{tab:vpe-heldout-regret} gives the regret of the five rules over the $2{,}379$ decisions.
Direct selection carries a mean regret of $0.76$ pp and a $90$th percentile of $2.74$ pp, and projection carries $2.33$ pp and $7.71$ pp.
The three static rules run from $4.09$ to $7.56$ pp in the mean and from $12.23$ to $20.14$ pp at the $90$th percentile, so both rules that read the fitted law stay below all three on those two measures.
The worst-case result differs: projection reaches $34.55$ pp, compared with $27.00$ pp for the rule that takes the largest visual token count, while direct selection stays at $19.37$ pp.
Both law-based rules read the same fitted law over the same candidates, so the gap between them comes from the projection step rather than from the law.
Under the bounded-law robustness check, direct selection has nearly identical mean and $90$th-percentile regrets of $0.75$ pp and $2.74$ pp.}

\begin{table}[htbp]
\appTableStyle
\centering
\small

\caption{Allocation regret in percentage points over the $2{,}379$ joint-holdout decisions. All five rules see the same candidates under the same budgets, the $90$th percentile uses linear interpolation, and Worst is the largest regret of a single decision. The two rules above the horizontal line read the fitted Separable law, and the three below it do not.}
\label{tab:vpe-heldout-regret}
\renewcommand{\arraystretch}{1.12}
\setlength{\tabcolsep}{6pt}
\begin{tabular}{lrrr}
\toprule
\rowcolor{appHeader}
\appHeadFont \textbf{Rule} & \appHeadFont \textbf{Mean} & \appHeadFont \textbf{90th pct.} & \appHeadFont \textbf{Worst} \\
\midrule
\rowcolor{appHighlight}
Direct selection & \appEmph{0.76} & \appEmph{2.74} & \appEmph{19.37} \\
\rowcolor{appStripe}
Projection & 2.33 & 7.71 & 34.55 \\
\midrule
Largest token count & 4.09 & 12.23 & 27.00 \\
\rowcolor{appStripe}
Largest backbone & 7.21 & 20.14 & 42.93 \\
Log-center & 7.56 & 16.22 & 38.74 \\
\bottomrule
\end{tabular}
\end{table}

\subsection{In-Sample Allocation and Breakdown by Series}
\label{app:vpe-regret}

\paragraph{{Projection outperforms static allocation rules on the complete grid.}}
{To diagnose allocation behavior by series and identify the largest-regret cases, we retain an earlier in-sample evaluation in which the fit and evaluation use the same cells; it is descriptive and does not support the held-out claim in \S\ref{sec:vpe-validation}.
For each of the 20 groups, four budgets $C_0$ sit at the $25$th, $50$th, $75$th, and $90$th percentiles of the cell costs of that group, which gives 80 decisions.
The cost is the language-side FLOP measure computed from the logged prompt lengths.
All decisions use the pooled cost fit together with the performance fit of their group.
For every budget we clip the two coordinates of the continuous optimum separately to the ranges observed in the group, and then pick the observed cell nearest in Euclidean distance in $(\log \nlin,\log \rlin)$ among the cells with $C_{\mathrm{cell}}\le C_0$.
Clipping need not keep the fitted cost equality, and picking the nearest cell need not minimize the fitted error over the finite grid.
The comparator is the cell with the lowest measured error in the same feasible set.
Regret is the error of the chosen cell minus the lowest feasible error, multiplied by 100 to give percentage points, and all 80 regrets are nonnegative.
The three static rules choose the largest feasible backbone, the largest feasible visual token count, or the log-center of the feasible ranges.
Ties are broken by dataset order, quantiles use linear interpolation, and all summaries use unrounded regrets.
Projection has lower mean, median, $90$th-percentile, and worst-case regret than every static rule and a higher exact-hit rate (Table~\ref{tab:vpe_regret_full}).
Its mean regret is $1.98$ pp, its median is $0.00$ pp, its $90$th percentile is $6.34$ pp, and its worst case is $11.52$ pp, compared with the best static mean of $6.51$ pp from the largest-token-count rule.
Although the static-rule ordering changes by benchmark, projection has the lowest median regret on all four benchmarks (Table~\ref{tab:vpe_regret_bench}).
Thus the fitted allocation gives substantially closer choices than fixed resource heuristics on the complete grid.}

\begin{table}[htbp]
\appTableStyle
\centering
\caption{{In-sample allocation regret in percentage points over the same 80 decisions, for projection and the three static rules. Exact hit counts the decisions that select the designated best cell, with ties broken by dataset order.}}
\label{tab:vpe_regret_full}
\footnotesize
\renewcommand{\arraystretch}{1.12}
\setlength{\tabcolsep}{4pt}
\begin{tabular}{lrrrrr}
\toprule
\rowcolor{appHeader}
\appHeadFont \textbf{Rule} & \appHeadFont \textbf{Mean} & \appHeadFont \textbf{Median} & \appHeadFont \textbf{90th pct.} & \appHeadFont \textbf{Worst} & \appHeadFont \textbf{Exact hit} \\
\midrule
\rowcolor{appHighlight}
{Projection} & \appEmph{1.98} & \appEmph{0.00} & \appEmph{6.34} & \appEmph{11.52} & \appEmph{51.25\%} \\
\rowcolor{appStripe}
{Largest token count} & 6.51 & 6.10 & 13.45 & 19.37 & 27.50\% \\
{Log-center} & 8.05 & 7.70 & 16.24 & 20.00 & 1.25\% \\
\rowcolor{appStripe}
{Largest backbone} & 15.40 & 14.23 & 28.32 & 41.36 & 0.00\% \\
\bottomrule
\end{tabular}
\end{table}

\begin{table}[htbp]
\appTableStyle
\centering
\caption{{Median allocation regret in percentage points by benchmark, 20 decisions each, from the same evaluation. Each entry pools four budgets across five model series.}}
\label{tab:vpe_regret_bench}
\footnotesize
\renewcommand{\arraystretch}{1.12}
\setlength{\tabcolsep}{4pt}
\begin{tabular}{lrrrr}
\toprule
\rowcolor{appHeader}
\appHeadFont \textbf{Benchmark} & \appHeadFont {\textbf{Projection}} & \appHeadFont \makecell{{\textbf{Largest}}\\{\textbf{token count}}} & \appHeadFont {\textbf{Log-center}} & \appHeadFont \makecell{{\textbf{Largest}}\\{\textbf{backbone}}} \\
\midrule
HR-Bench & \appEmph{0.00} & 5.63 & 11.06 & 14.23 \\
\rowcolor{appStripe}
{MME-RealWorld} & \appEmph{0.07} & 3.54 & 6.52 & 17.77 \\
{TreeBench} & \appEmph{2.49} & 6.09 & 4.10 & 4.85 \\
\rowcolor{appStripe}
{V\textsuperscript{*}\,Bench} & \appEmph{0.00} & 7.85 & 9.16 & 27.23 \\
\bottomrule
\end{tabular}
\end{table}

\paragraph{{Projection remains close to the optimum across series and families.}}
\label{app:vpe-byfamily}
{To determine whether the aggregate gain depends on a narrow subset of groups, we measure exact recovery and regret tolerance and break the same 80 decisions down by series, benchmark, and family.
Projection selects the designated best cell in $41$ decisions, or $51.25\%$, while $42$, or $52.50\%$, reach the lowest feasible error; the difference is one tied decision for InternVL3 on V\textsuperscript{*}\,Bench at the $75$th-percentile budget, where 9B and 8B at $2048$\,px have the same error.
Model sizes here follow the model names, and the number after the size is the input size.
Table~\ref{tab:vpe_tolerance} shows that $52$ decisions, or $65.00\%$, lie within $2$ pp of the optimum and $77$, or $96.25\%$, lie within $8$ pp; all 32 QwenVL decisions and $45$ of the 48 InternVL decisions meet the $8$ pp threshold.
Table~\ref{tab:vpe_per_series} reports the four decisions in each of the 20 groups, and Eq.\eqref{eq:vpe-pooled-regret} shows how their unrounded group means produce the pooled mean of $1.98$ pp in Table~\ref{tab:vpe_regret_full}; the held-out means in \S\ref{sec:vpe-validation} come from a separate evaluation, and pooled medians and percentiles are computed directly over the 80 regrets.
InternVL has a mean regret of $2.16$ pp, a median of $0.00$ pp, a $90$th percentile of $7.06$ pp, and a worst case of $11.52$ pp over 48 decisions, while QwenVL has $1.71$ pp, $0.00$ pp, $4.70$ pp, and $7.85$ pp over 32 decisions.
For family $f$, the empirical distribution function is $\widehat F_f(t)=n_f^{-1}\sum_{i\in f}\mathbf{1}\{\mathrm{regret}_i\le t\}$, with $n_{\mathrm{InternVL}}=48$ and $n_{\mathrm{QwenVL}}=32$; the two curves cross, so the family ordering depends on the threshold, but projection remains within $8$ pp of the optimum for nearly every decision in both families.}

\begin{equation}
\overline{\mathrm{regret}}_{\mathrm{pooled}}
=\frac{1}{20}\sum_{g=1}^{20}\left(\frac{1}{4}\sum_{b=1}^{4}\mathrm{regret}_{g,b}\right)
=1.979870\ \mathrm{pp}\simeq1.98\ \mathrm{pp}.
\label{eq:vpe-pooled-regret}
\end{equation}
\begin{table}[htbp]
\appTableStyle
\centering
\caption{{Share of projection decisions with regret at or below each threshold. The denominators are 80 decisions pooled, 48 for InternVL, and 32 for QwenVL. The zero-regret row counts every tied minimum, whichever cell was chosen.}}
\label{tab:vpe_tolerance}
\footnotesize
\renewcommand{\arraystretch}{1.12}
\setlength{\tabcolsep}{4pt}
\begin{tabular}{lrrr}
\toprule
\rowcolor{appHeader}
\appHeadFont \textbf{Threshold} & \appHeadFont \textbf{Pooled} & \appHeadFont \textbf{InternVL} & \appHeadFont \textbf{QwenVL} \\
\midrule
$\le 0$ pp & 52.50\% & 52.08\% & 53.13\% \\
\rowcolor{appStripe}
$\le 1$ pp & 58.75\% & 58.33\% & 59.38\% \\
$\le 2$ pp & 65.00\% & 64.58\% & 65.63\% \\
\rowcolor{appStripe}
$\le 5$ pp & 85.00\% & 81.25\% & 90.63\% \\
$\le 8$ pp & 96.25\% & 93.75\% & 100.00\% \\
\rowcolor{appStripe}
$\le 10$ pp & 98.75\% & 97.92\% & 100.00\% \\
\bottomrule
\end{tabular}
\end{table}

\begin{table}[htbp]
\appTableStyle
\centering
\caption{{Projection regret in percentage points by series and benchmark, from the same 80 decisions as Table~\ref{tab:vpe_regret_full}. Each row summarizes four decisions, computed from unrounded regrets, and the $90$th percentile uses linear interpolation. Hits count the selections of the designated best cell, with ties broken by dataset order. For InternVL3 on V\textsuperscript{*}\,Bench, all four decisions reach the lowest feasible error and one of them selects a different cell tied with the designated best.}}
\label{tab:vpe_per_series}
\footnotesize
\renewcommand{\arraystretch}{1.12}
\setlength{\tabcolsep}{4pt}
\begin{tabular}{llrrrrc}
\toprule
\rowcolor{appHeader}
\appHeadFont \textbf{Series} & \appHeadFont \textbf{Benchmark} & \appHeadFont \textbf{Mean} & \appHeadFont \textbf{Median} & \appHeadFont \textbf{90th pct.} & \appHeadFont \textbf{Worst} & \appHeadFont \textbf{Hits} \\
\midrule
InternVL2.5 & HR-Bench & 2.16 & 0.00 & 6.04 & 8.63 & 3/4 \\
\rowcolor{appStripe}
InternVL2.5 & MME-RealWorld & 3.44 & 2.02 & 8.03 & 9.74 & 2/4 \\
InternVL2.5 & TreeBench & 3.11 & 2.74 & 6.07 & 6.97 & 1/4 \\
\rowcolor{appStripe}
InternVL2.5 & V\textsuperscript{*}\,Bench & 4.19 & 2.62 & 9.63 & 11.52 & 2/4 \\
\midrule
InternVL3 & HR-Bench & 0.44 & 0.00 & 1.23 & 1.76 & 3/4 \\
\rowcolor{appStripe}
InternVL3 & MME-RealWorld & 2.16 & 0.68 & 5.51 & 7.30 & 2/4 \\
InternVL3 & TreeBench & 3.98 & 3.61 & 6.89 & 7.71 & 0/4 \\
\rowcolor{appStripe}
InternVL3 & V\textsuperscript{*}\,Bench & 0.00 & 0.00 & 0.00 & 0.00 & 3/4 \\
\midrule
InternVL3.5 & HR-Bench & 0.00 & 0.00 & 0.00 & 0.00 & 4/4 \\
\rowcolor{appStripe}
InternVL3.5 & MME-RealWorld & 1.51 & 0.32 & 3.94 & 5.42 & 1/4 \\
InternVL3.5 & TreeBench & 3.52 & 4.31 & 5.25 & 5.47 & 1/4 \\
\rowcolor{appStripe}
InternVL3.5 & V\textsuperscript{*}\,Bench & 1.44 & 1.05 & 3.19 & 3.66 & 2/4 \\
\midrule
Qwen2.5-VL & HR-Bench & 3.38 & 3.31 & 6.24 & 6.88 & 1/4 \\
\rowcolor{appStripe}
Qwen2.5-VL & MME-RealWorld & 1.66 & 1.20 & 3.68 & 4.22 & 2/4 \\
Qwen2.5-VL & TreeBench & 3.11 & 3.23 & 3.91 & 3.98 & 0/4 \\
\rowcolor{appStripe}
Qwen2.5-VL & V\textsuperscript{*}\,Bench & 3.53 & 3.14 & 7.38 & 7.85 & 2/4 \\
\midrule
Qwen3-VL & HR-Bench & 1.06 & 0.06 & 2.93 & 4.13 & 2/4 \\
\rowcolor{appStripe}
Qwen3-VL & MME-RealWorld & 0.72 & 0.00 & 2.01 & 2.87 & 3/4 \\
Qwen3-VL & TreeBench & 0.19 & 0.00 & 0.52 & 0.75 & 3/4 \\
\rowcolor{appStripe}
Qwen3-VL & V\textsuperscript{*}\,Bench & 0.00 & 0.00 & 0.00 & 0.00 & 4/4 \\
\bottomrule
\end{tabular}
\end{table}

\paragraph{{Large regrets are rare and localized to three InternVL2.5 decisions.}}
{Exactly three decisions exceed $8$ pp, and all three come from InternVL2.5.
On V\textsuperscript{*}\,Bench at the $75$th-percentile budget, the rule selects 26B at $512$\,px against a best feasible 8B at $2048$\,px, with a regret of $11.52$ pp and $\beta=10.00$.
On MME-RealWorld at the $25$th-percentile budget, it selects 2B at $512$\,px against 1B at $2048$\,px, with $9.74$ pp and $\beta=10.00$.
On HR-Bench at the $25$th-percentile budget, it selects 2B at $768$\,px against 1B at $4096$\,px, with $8.63$ pp and $\beta=4.6623$, under the same budget for both cells.
The HR-Bench case has a $\beta$ away from the bounds, so these tails are not confined to the fits that reach the upper bound.
The in-sample gains are therefore broad rather than hiding many large failures, although these three cases do not identify an architectural cause because both fitted-surface error and the clipping and projection step can produce regret on a finite grid.}

\subsection{Benchmark Profiles by Family}
\label{app:vpe-diag-byfamily}

\paragraph{{Family-specific profiles reveal different dominant resource axes.}}
{To test whether pooled benchmark profiles conceal frontend-specific scaling behavior, we split the $19{,}890$ trajectories by benchmark and model family, report the three regime shares in Table~\ref{tab:diag}, and within the \textsc{Scaling Bound} subset compare the range of mean accuracy along the backbone and visual-token axes in Table~\ref{tab:vpe_family_profile}.
We call a trajectory $\nlin$-dominated when its accuracy range is larger along the backbone axis and $\rlin$-dominated when it is larger along the visual-token axis. This measures the magnitude of movement rather than whether adding the resource improves accuracy. Of the $193$ $\rlin$-dominated QwenVL trajectories on V\textsuperscript{*}\,Bench, for instance, $9$ are less accurate and $11$ are equally accurate at the largest input size as at the smallest.
HR-Bench requires an additional check because $1{,}423$ \textsc{Scaling Bound} trajectories exist only at $4096$\,px, $1{,}007$ from InternVL and $416$ from QwenVL, so their visual-token range is zero and they are classified as $\nlin$-dominated by construction. After retaining only trajectories with at least two input sizes, the $\nlin$-dominated, $\rlin$-dominated, and tie shares are $82.2\%$, $11.6\%$, and $6.2\%$ for InternVL and $46.3\%$, $45.8\%$, and $7.9\%$ for QwenVL, so InternVL is more backbone-dominated under either definition.
Across the accuracy cutoffs, MME-RealWorld and TreeBench have the largest \textsc{Ceiling Bound} shares in both families, from $38.1\%$ to $42.0\%$, and the benchmark contrast also holds under the response-based criteria even though the exact shares depend on the criterion and cohort.
The family difference is clearest in the dominant axis. On every benchmark the $\rlin$-dominated share of QwenVL exceeds that of InternVL, from $31.6\%$ versus $6.7\%$ on HR-Bench to $72.6\%$ versus $38.5\%$ on V\textsuperscript{*}\,Bench, and on MME-RealWorld QwenVL is majority $\rlin$-dominated at $53.5\%$ even though the pooled profile is $\nlin$-dominated.
Family-level analysis therefore gives a consistent result. QwenVL trajectories are more sensitive to the visual-token axis, whereas InternVL trajectories are more often separated by backbone size, so benchmark scaling profiles should be interpreted together with the visual frontend.}

\begin{table}[htbp]
\appTableStyle
\centering
\caption{{Regime shares for each benchmark over the $19{,}890$ trajectories, under the accuracy cutoffs of \S\ref{sec:regime}. A separate response-based analysis uses a smaller matched cohort and stricter criteria.}}
\label{tab:diag}
\footnotesize
\renewcommand{\arraystretch}{1.10}
\begin{tabular}{lccc}
\toprule
\rowcolor{appHeader}
\appHeadFont \textbf{Benchmark} & \appHeadFont \textsc{Easy} & \appHeadFont {\textsc{Scaling Bound}} & \appHeadFont {\textsc{Ceiling Bound}} \\
\midrule
HR-Bench & 33.5\% & 46.5\% & \appEmph{20.0\%} \\
\rowcolor{appStripe}
{MME-RealWorld} & 10.3\% & 50.3\% & \appEmph{39.4\%} \\
{TreeBench} & 10.4\% & 48.6\% & \appEmph{41.0\%} \\
\rowcolor{appStripe}
{V\textsuperscript{*}\,Bench} & 16.2\% & \appEmph{66.2\%} & 17.6\% \\
\bottomrule
\end{tabular}
\end{table}

\begin{table}[htbp]
\appTableStyle
\centering
\caption{{Regime shares and dominant axis for each benchmark and family, over the same $19{,}890$ trajectories. Regime shares are fractions of all trajectories of that family and benchmark, and the dominant axis is computed over the \textsc{Scaling Bound} trajectories only. The HR-Bench rows include the trajectories that exist only at $4096$\,px, which the text discusses.}}
\label{tab:vpe_family_profile}
\footnotesize
\setlength{\tabcolsep}{4pt}
\resizebox{\linewidth}{!}{%
\begin{tabular}{llrrrrrr}
\toprule
\rowcolor{appHeader}
& & \multicolumn{3}{c}{\appHeadFont \textbf{Regime share}} & \multicolumn{3}{c}{\appHeadFont {\textbf{Dominant axis within \textsc{Scaling Bound}}}} \\
\cmidrule(lr){3-5}
\cmidrule(lr){6-8}
\rowcolor{appHeader}
\appHeadFont \textbf{Family} & \appHeadFont \textbf{Benchmark} & \appHeadFont \textsc{Easy} & \appHeadFont {\textsc{Scaling}} & \appHeadFont {\textsc{Ceiling}} & \appHeadFont {$\nlin$-dominated} & \appHeadFont {$\rlin$-dominated} & \appHeadFont \textbf{Tie} \\
\midrule
InternVL & HR-Bench      & 29.0\% & 49.6\% & 21.4\% & 89.7\% & 6.7\%  & 3.6\% \\
\rowcolor{appStripe}
InternVL & {MME-RealWorld} & 10.1\% & 51.8\% & 38.1\% & 70.5\% & 20.7\% & 8.8\% \\
InternVL & {TreeBench}     & 8.6\%  & 49.4\% & 42.0\% & 88.5\% & 7.0\%  & 4.5\% \\
\rowcolor{appStripe}
InternVL & {V\textsuperscript{*}\,Bench}  & 16.8\% & 63.9\% & 19.4\% & 47.8\% & 38.5\% & 13.7\% \\
\midrule
QwenVL   & HR-Bench      & 40.3\% & 41.8\% & 17.9\% & 63.0\% & 31.6\% & 5.5\% \\
\rowcolor{appStripe}
QwenVL   & {MME-RealWorld} & 10.7\% & 48.0\% & 41.2\% & 40.4\% & 53.5\% & 6.1\% \\
QwenVL   & {TreeBench}     & 13.1\% & 47.4\% & 39.5\% & 58.6\% & 35.9\% & 5.5\% \\
\rowcolor{appStripe}
QwenVL   & {V\textsuperscript{*}\,Bench}  & 15.4\% & 69.6\% & 14.9\% & 22.6\% & 72.6\% & 4.9\% \\
\bottomrule
\end{tabular}
}
\end{table}

\section{Case Studies}
\label{app:case-studies}

{We close with two sets of examples, one that holds the model fixed and raises the input size, and one that holds the input size fixed and compares model sizes.
These are individual questions chosen to show what changes along each axis rather than averages over the full grid.
The outputs alone also leave open why an answer changes.}

\paragraph{{Sensitivity to input size across model sizes.}}
{Figure~\ref{fig:casestudy_multipage} shows selected outputs of the 2B, 4B, 8B, and 32B models of Qwen3-VL at input sizes from $224$ to $2048$\,px.
The two 32B examples become correct at $768$ and $512$\,px and stay correct at the larger input sizes.
The 8B example asks the model to identify a face, and it is answered incorrectly from $224$ to $768$\,px and correctly at $1024$ and $2048$\,px.
Of the two 4B examples, one becomes correct at $512$\,px and the remote-sensing example at $768$\,px; both stay correct above.
In these five examples a larger input helps on questions that turn on a small detail.
The 2B example moves in both directions instead, since it is answered correctly at $224$ and $336$\,px, incorrectly from $512$ to $1024$\,px, and correctly again at $2048$\,px.
Raising the input size therefore need not help every individual question.}

\paragraph{{Sensitivity to model size across input sizes.}}
{Figure~\ref{fig:casestudy_multipage2} compares the four models on selected questions, with the input size held fixed within each example.
At $224$\,px the question asks for the color of a shirt, and the 8B and 32B models answer it correctly while the 2B and 4B models do not.
The same pattern holds for the text on a billboard at $512$\,px, the location on a map at $768$\,px, the fax number at $1024$\,px, and the color of a bicycle basket at $2048$\,px.
At $336$\,px the question asks for the color of a flag, and there the 4B model answers correctly as well.
These examples show errors that persist in the smaller models even at a large input size.
Each panel fixes the input size and different panels show different questions, so they say nothing about how the gap between model sizes changes with the input size.}

% ========= Page 1: the 2B and 4B models =========
\begin{figure}[t]
    \centering

    % --- 2B ---
    \includegraphics[width=\linewidth]{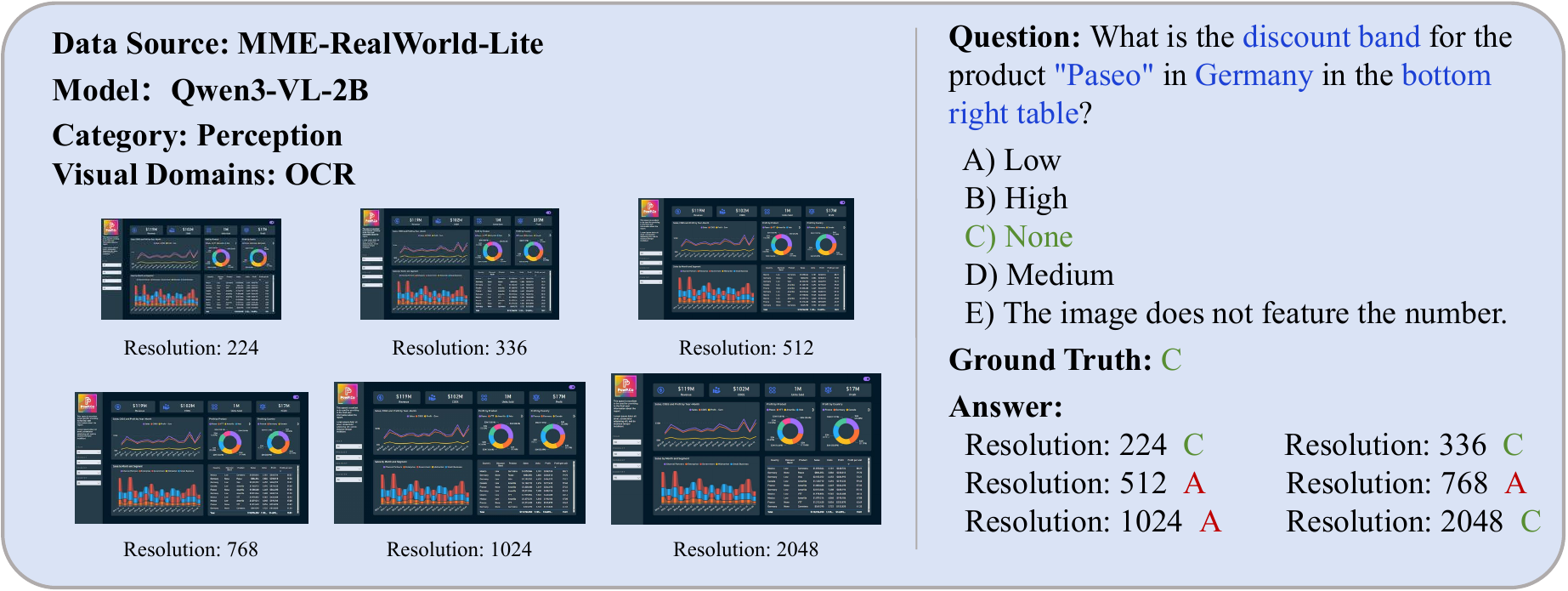}
    \par\vspace{14pt}

    % --- 4B, first example ---
    \includegraphics[width=\linewidth]{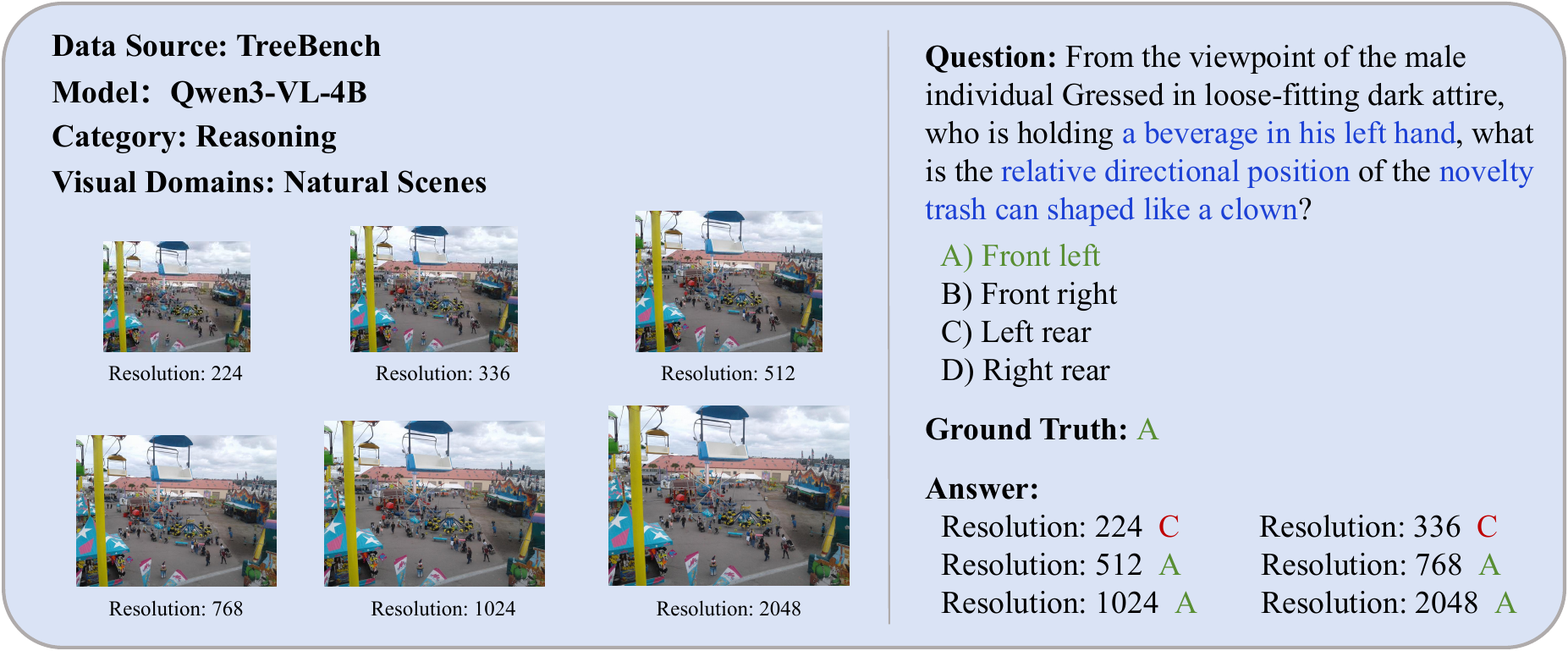}
    \par\vspace{14pt}

    % --- 4B, second example ---
    \includegraphics[width=\linewidth]{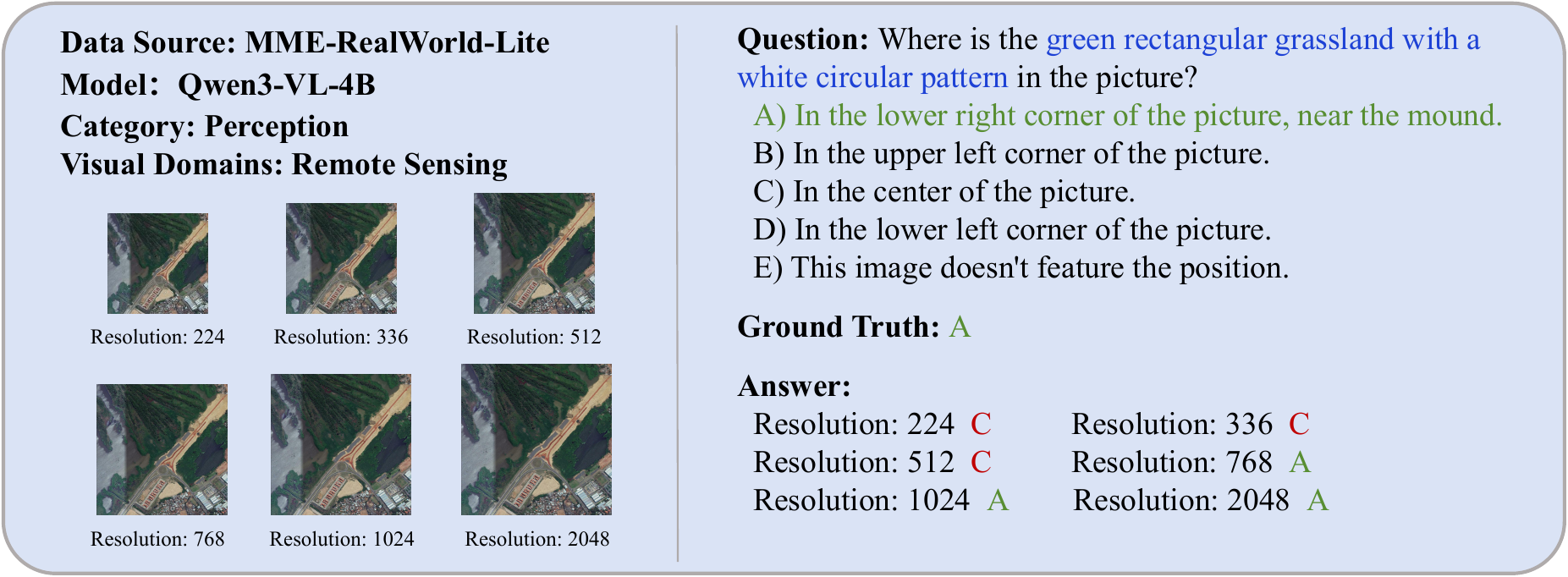}

    % --- Caption, part 1 ---
    \caption{{\textbf{Sensitivity to input size across model sizes.}
    Selected outputs of Qwen3-VL at 2B, 4B, 8B, and 32B and at input sizes from $224$ to $2048$\,px.
    Green marks the correct option and red marks an incorrect answer.
    This page shows the 2B and 4B models.
    The 2B example moves in both directions; one 4B example becomes correct at $512$\,px and the remote-sensing example at $768$\,px.}}
    \label{fig:casestudy_multipage}
\end{figure}

\clearpage

% ========= Page 2: the 8B and 32B models =========
\begin{figure}[t]
    \ContinuedFloat % keeps the figure number of the previous page
    \captionsetup{labelformat=continued}
    \centering

    % --- 8B ---
    \includegraphics[width=\linewidth]{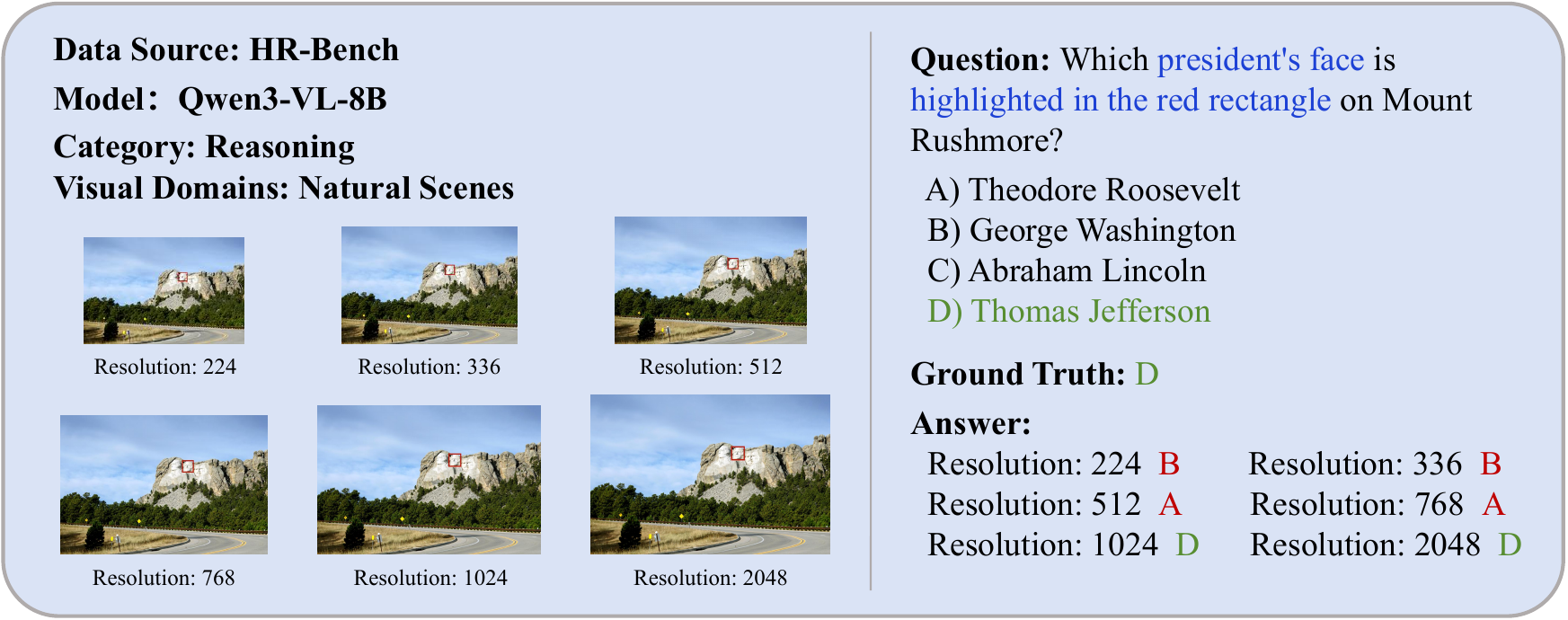}
    \par\vspace{14pt}

    % --- 32B, first example ---
    \includegraphics[width=\linewidth]{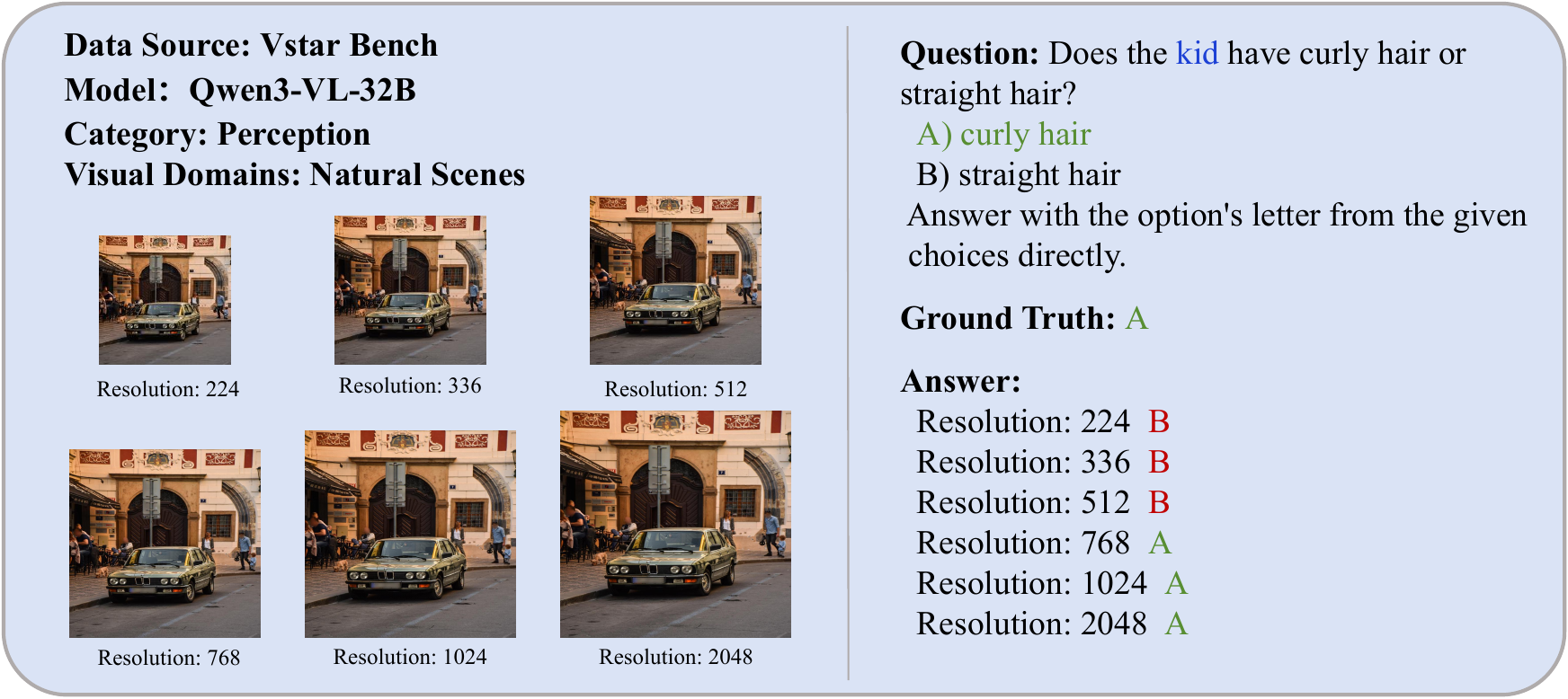}
    \par\vspace{14pt}

    % --- 32B, second example ---
    \includegraphics[width=\linewidth]{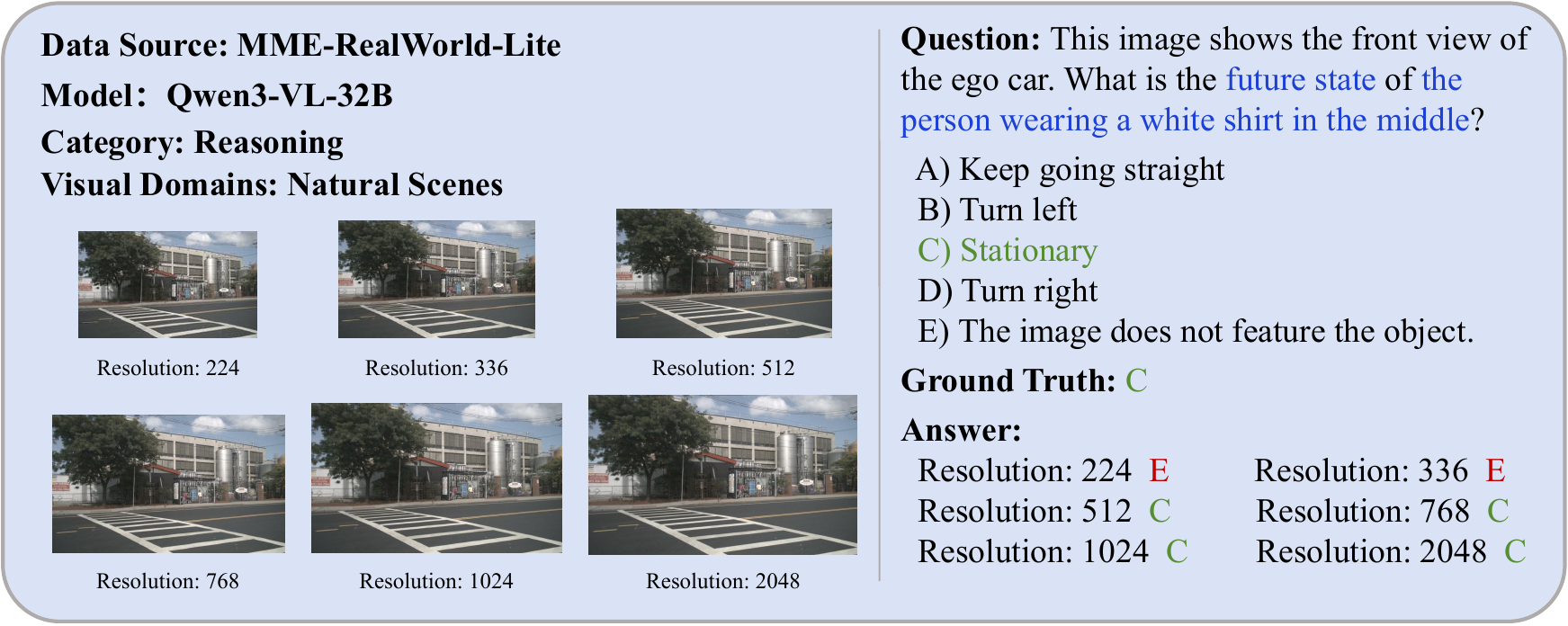}

    % --- Caption, part 2 ---
    \caption[]{{Sensitivity to input size for the 8B and 32B models.
    The 8B answer becomes correct at $1024$\,px, and the two 32B answers become correct at $768$ and $512$\,px.}}
    \label{fig:casestudy_multipage_1}
\end{figure}

\clearpage

% ========= Page 3: one question per panel, at 224, 336 and 512 px =========
\begin{figure}[t]
    \centering

    % --- 224 px ---
    \includegraphics[width=\linewidth]{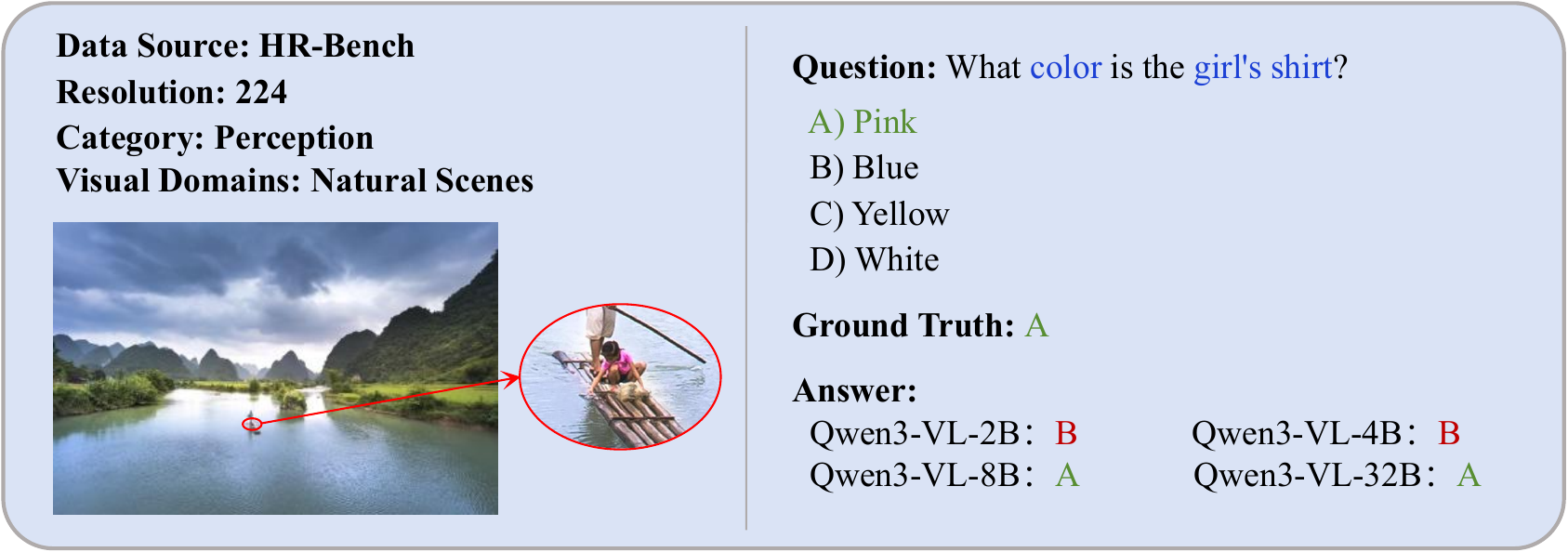}

    \par\vspace{14pt}

    % --- 336 px ---
    \includegraphics[width=\linewidth]{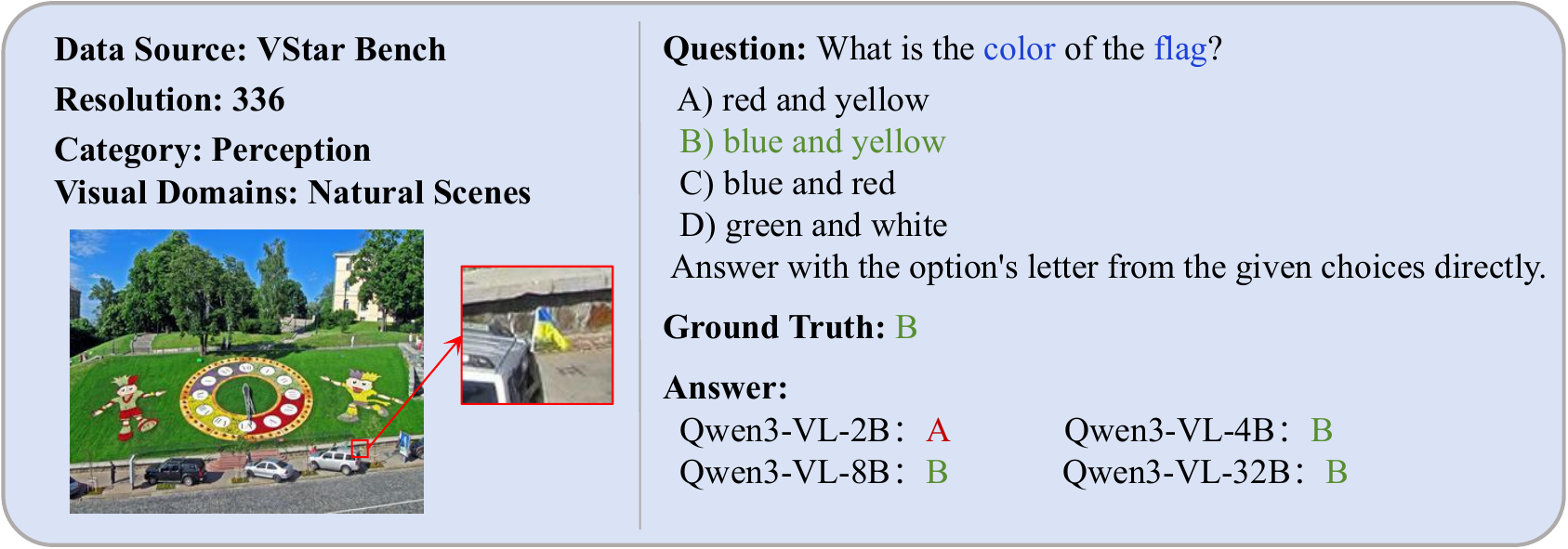}

     \par\vspace{14pt}

    % --- 512 px ---
    \includegraphics[width=\linewidth]{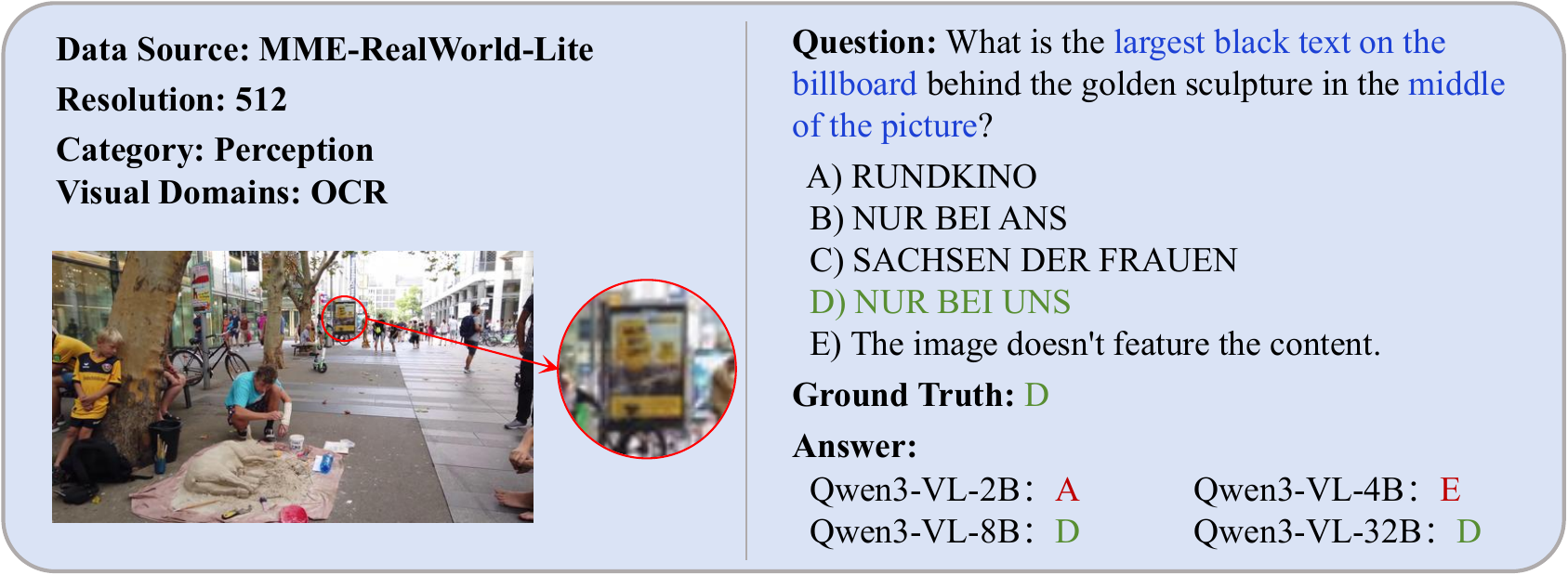}

    % --- Caption, part 1 ---
    \caption{{\textbf{Sensitivity to model size across input sizes.}
    Each panel compares the Qwen3-VL models on one question at a fixed input size.
    These three examples use $224$, $336$, and $512$\,px, and the question differs from panel to panel.}}
    \label{fig:casestudy_multipage2}
\end{figure}

\clearpage

% ========= Page 4: one question per panel, at 768, 1024 and 2048 px =========
\begin{figure}[t]
    \ContinuedFloat % keeps the figure number of the previous page
    \captionsetup{labelformat=continued}
    \centering

    % --- 768 px ---
    \includegraphics[width=\linewidth]{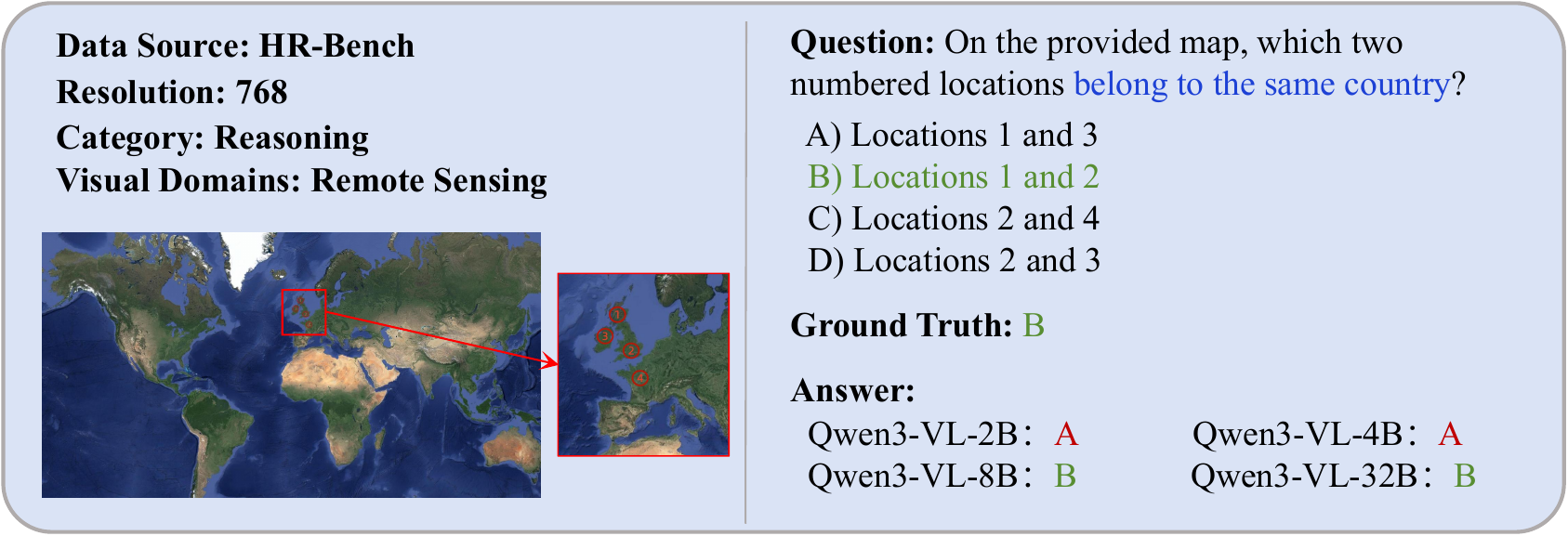}

    \par\vspace{14pt}

    % --- 1024 px ---
    \includegraphics[width=\linewidth]{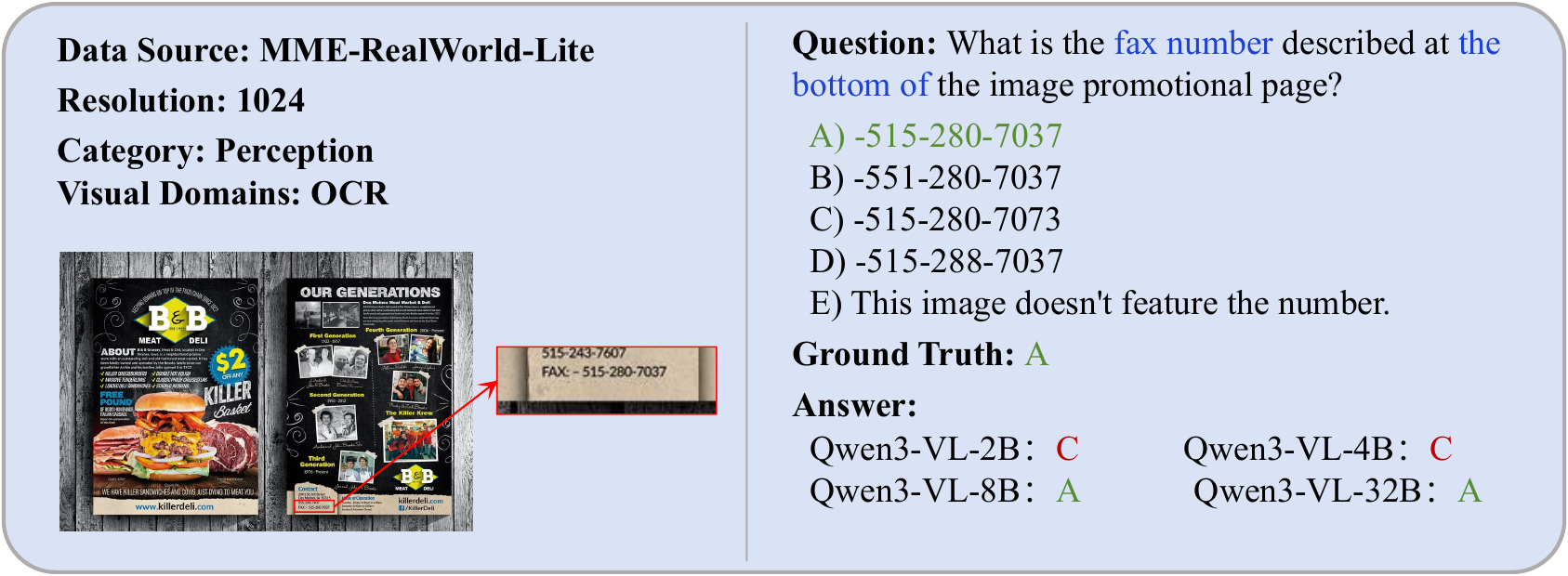}

     \par\vspace{14pt}

    % --- 2048 px ---
    \includegraphics[width=\linewidth]{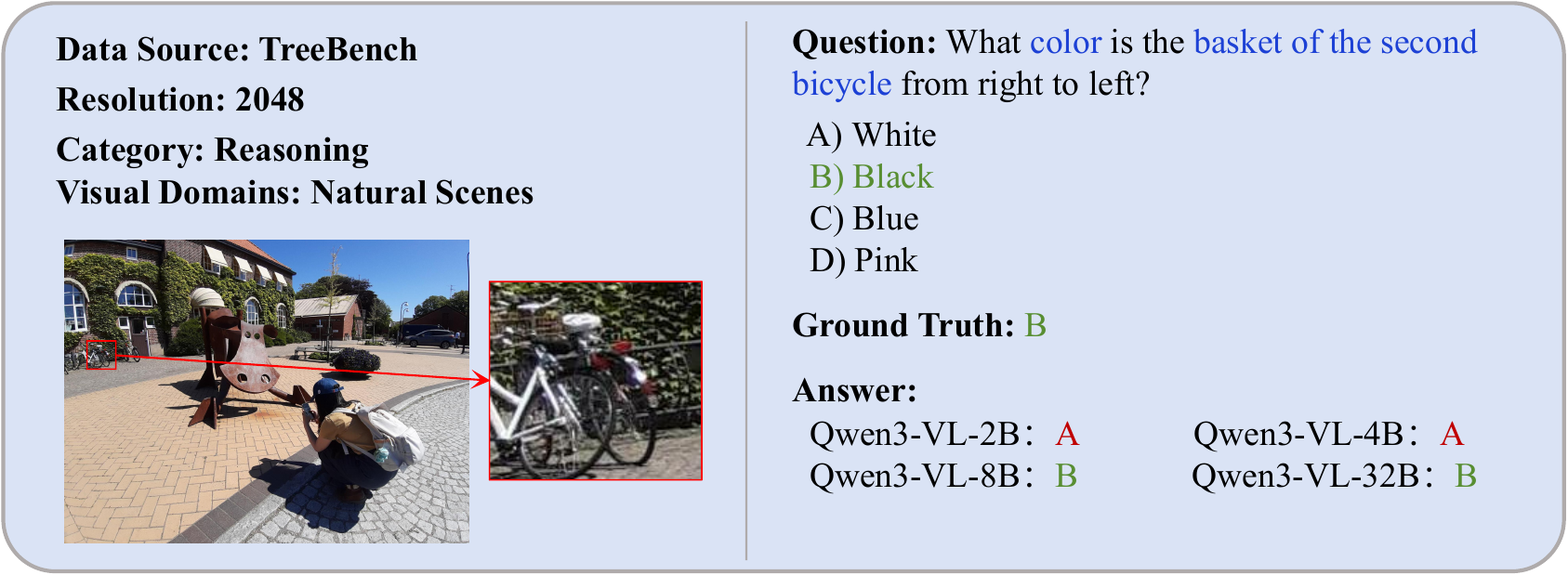}

    % --- Caption, part 2 ---
    \caption[]{{Model comparisons at $768$, $1024$, and $2048$\,px.
    In these three examples the 8B and 32B models answer correctly while the 2B and 4B models do not.}}
    \label{fig:casestudy_multipage2_2}
\end{figure}
\end{document}